\documentclass{article} 
\usepackage{iclr2022_conference,times}

\usepackage{hyperref}
\usepackage{url}

\usepackage[utf8]{inputenc} 
\usepackage[T1]{fontenc}    
\usepackage{adjustbox}
\usepackage{url}            
\usepackage{booktabs}       
\usepackage{amsfonts}       
\usepackage{nicefrac}       
\usepackage{microtype}      
\usepackage{xcolor}         
\usepackage{algorithm}
\usepackage{algpseudocode}
\usepackage{amsmath}
\usepackage{amssymb}
\usepackage{array}
\usepackage{booktabs}
\usepackage{caption}
\usepackage{cite}
\usepackage{color}
\usepackage{dirtytalk}
\usepackage{epsfig}
\usepackage{graphicx}
\usepackage{gensymb}
\usepackage{lipsum}
\usepackage{makecell}
\usepackage{mathtools}
\usepackage{multirow}
\usepackage{stmaryrd}
\usepackage{times}
\usepackage{xcolor}
\usepackage{enumitem}
\usepackage{pgfplotstable}
\usepackage{subcaption}
\usepackage{wrapfig}
\usepackage{xspace}
\usepackage{stackengine}
\usepackage{tikz}

\usepackage[toc,page,header]{appendix}
\usepackage{minitoc}

\DeclareUnicodeCharacter{2212}{-}

\renewcommand{\cite}{\citep}

\newcommand{\ra}[1]{\renewcommand{\arraystretch}{#1}}
\def\m#1{\ensuremath{\mathtt{#1}}}

\def\v#1{\ensuremath{\mathbf{#1}}}

\def\parr#1{\left(#1\right)}
\def\curl#1{\left\{#1\right\}}
\def\croch#1{\left[#1\right]}

\def\I{\m I}

\def\H{\m H}

\def\R{\m R}

\def\t{\v t}
\def\d{\v d}

\def\S{\m S}
\def\D{\m D}
\def\C{\m C}

\def\K{\m K}
\def\p{\v p}

\def\u{\v u}

\def\x{\v x}
\def\T{\m T}

\def\H{\m H}

\def\Real{\mathbb{R}}
\def\SO{\mathrm{SO(3)}}

\newcommand{\neurhal}{\rm{NeurHal}\xspace}

\newcommand{\aka}{\textit{a.k.a. }}
\newcommand{\wrt}{\textit{w.r.t. }}
\newcommand{\eg}{\textit{e.g. }}
\newcommand{\ie}{\textit{i.e. }}
\newcommand{\defeq}{\vcentcolon=}

\definecolor{orange}{rgb}{1.0,0.5,0}

\definecolor{DarkGreen}{rgb}{0,0.5,0}

\definecolor{color_SPSG}{rgb}{0.12,0.47,0.71}
\definecolor{color_LoFTR}{rgb}{1.00,0.50,0.05}
\definecolor{color_S2DLoc}{rgb}{0.17,0.63,0.17}
\definecolor{color_R2D2}{rgb}{0.84,0.15,0.16}
\definecolor{color_Random}{rgb}{0.58,0.40,0.74}
\definecolor{color_Identity}{rgb}{0.55,0.34,0.29}
\definecolor{color_Mixed6e}{rgb}{0.89,0.47,0.76}
\definecolor{color_DRCNet}{rgb}{0.50,0.50,0.50}
\definecolor{color_S2DLoc_RE}{rgb}{0.74,0.74,0.13}

\definecolor{color_VCH}{rgb}{0.74,0.81,0.71}
\definecolor{color_VCH_I}{rgb}{0.25,0.62,0.51}
\definecolor{color_VCH_O}{rgb}{0.10,0.37,0.42}
\definecolor{color_VCH_IO}{rgb}{0.10,0.20,0.31}
\definecolor{color_VCH_IO_covis}{rgb}{1.00,0.78,0.77}
\definecolor{color_VCH_IO_noncovis}{rgb}{0.68,0.27,0.42}
\definecolor{color_VCH_IO_RE}{rgb}{0.68,0.27,0.42}

\definecolor{color_gamma_50}{rgb}{0.10,0.20,0.31}
\definecolor{color_gamma_25}{rgb}{0.07,0.45,0.46}
\definecolor{color_gamma_10}{rgb}{0.43,0.68,0.54}
\definecolor{color_gamma_00}{rgb}{0.74,0.81,0.71}
\definecolor{color_covis_only}{rgb}{0.96,0.64,0.66}
\definecolor{color_non_covis_only}{rgb}{0.80,0.38,0.49}
\definecolor{color_matchable_only}{rgb}{0.53,0.17,0.64}
\definecolor{color_identified_only}{rgb}{0.12,0.47,0.71}
\definecolor{color_RE}{rgb}{0.80,0.38,0.49}
\definecolor{color_scannet_with_megadepth}{rgb}{0.96,0.64,0.66}
\definecolor{color_scannet_with_scannet}{rgb}{0.68,0.27,0.42}
\definecolor{color_NeurHal_NRE}{rgb}{0.10,0.20,0.31}
\definecolor{color_NeurHal_RE}{rgb}{0.96,0.64,0.66}
\definecolor{color_NRE_NRE}{rgb}{0.68,0.27,0.42}
\definecolor{color_NRE_RE}{rgb}{1.00,0.78,0.77}
\definecolor{color_scannet}{rgb}{0.86, 0.35, 0.31}
\definecolor{color_megadepth}{rgb}{0.12, 0.47, 0.71}
\definecolor{color_inpainting}{rgb}{0.31, 0.64, 0.87}
\definecolor{color_inpainting_loftr}{rgb}{0.96, 0.28, 0.29}
\definecolor{color_inpainting_s2d}{rgb}{0.86, 0.52, 0.78}
\definecolor{color_inpainting_drcnet}{rgb}{0.55, 0.56, 0.56}

\definecolor{homography_neural}{HTML}{009ADE}
\definecolor{homography_ls}{HTML}{d1b840}
\definecolor{homography_oracle}{HTML}{4711d9}
\definecolor{homography_ransac}{HTML}{a5333c}
\definecolor{homography_ransac_gt}{HTML}{7896ab}

\definecolor{homography_ransac_1}{HTML}{fa6e78}
\definecolor{homography_ransac_2}{HTML}{da5761}
\definecolor{homography_ransac_3}{HTML}{a5333c}
\definecolor{homography_ransac_4}{HTML}{871c28}
\definecolor{homography_ransac_5}{HTML}{730a1b}
\definecolor{homography_ransac_6}{HTML}{5c2500}
\definecolor{homography_ransac_7}{HTML}{4d3400}

\title{Visual Correspondence Hallucination}

\author{Hugo Germain$^1$, Vincent Lepetit$^1$ and Guillaume Bourmaud$^2$\\
  $^1$LIGM, \'Ecole des Ponts, Univ Gustave Eiffel, CNRS, Marne-la-Vall\'ee, France\\
  $^2$IMS, University of Bordeaux, Bordeaux INP, CNRS, Bordeaux, France\\
  \texttt{\{firstname.lastname\}@enpc.fr}, \texttt{guillaume.bourmaud@u-bordeaux.fr}\\
}

\iclrfinalcopy 

\begin{document}

\maketitle

\doparttoc 
\part{} 
\vspace{-2cm}

\begin{abstract}

Given a pair of partially overlapping source and target images and a keypoint in
the source image, the keypoint's correspondent in the target image can be either
visible, occluded or outside the field of view. Local feature matching methods
are only able to identify the correspondent's location when it is visible, while
humans can also hallucinate (\ie predict) its location when it is
occluded or outside the field of view through geometric reasoning.  In this
paper, we bridge this gap by training a network to output a peaked probability
distribution over the correspondent's location, regardless of this correspondent
being visible, occluded, or outside the field of view.  We experimentally
demonstrate that this network is indeed able to hallucinate correspondences on
pairs of images captured in scenes that were not seen at training-time.  We also
apply this network to an absolute camera pose estimation problem and find it is
significantly more robust than state-of-the-art local feature matching-based
competitors.

\end{abstract}

\section{Introduction}


Establishing correspondences between two partially overlapping images is a
fundamental computer vision problem with many applications.  For example,
state-of-the-art methods for visual localization from an input image rely on
keypoint matches between the input image and a reference
image~\cite{6DOFBenchmark, Sarlin2019FromCT, SuperGlue, R2D2}. However, these
local feature matching methods will still fail when few keypoints are
\emph{covisible}, \ie when many image locations in one image are outside
the field of view or become occluded in the second image.
These failures are to be expected since these methods are pure pattern
recognition approaches that seek to \emph{identify} correspondences, \ie to find
correspondences in covisible regions, and consider the non-covisible regions as
noise. By contrast, humans explain the presence of these non-covisible regions
through geometric reasoning and consequently are able to
\emph{hallucinate} (\ie predict) correspondences at those locations.
%
%
%
%
Geometric reasoning has already been used in computer vision for image matching,
but usually as an \emph{a posteriori} processing~\cite{Fischler1981RandomSC,
Luong1996, barath2018graph, Chum2003LocallyOR, DEGENSAC, MAGSAC, MAGSACpp}.
These methods seek to remove outliers from the set of correspondences produced
by a local feature matching approach using only limited geometric models such as
epipolar geometry or planar assumptions.



\paragraph{Contributions.} In this paper we tackle the problem of correspondence
hallucination. In doing so we seek to answer two questions: $(i)$ can we derive
a network architecture able to learn to hallucinate correspondences? and $(ii)$
is correspondence hallucination beneficial for absolute pose estimation?
%
%
The answer to these questions is the main novelty of this paper.
More precisely, we consider a network that takes as input a pair of partially
overlapping source/target images and keypoints in the source image, and outputs
for each keypoint a probability distribution over its correspondent's location
in the target image plane. We propose to train this network to both identify and
hallucinate the keypoints' correspondents.  We call the resulting method
NeurHal, for Neural Hallucinations.  To the best of our knowledge, learning to
hallucinate correspondences is a virgin territory, thus we first provide an
analysis of the specific features of that novel learning task. 
This analysis guides us towards employing an appropriate loss function and
designing the architecture of the network. After training the network, we
experimentally demonstrate that it is indeed able to hallucinate correspondences
on unseen pairs of images captured in novel scenes. We also apply this
network to a camera pose estimation problem and find it is significantly more
robust than state-of-the-art local feature matching-based competitors.

\begin{figure}[t]
 \begin{center}
    \centering
    \includegraphics[width=\textwidth]{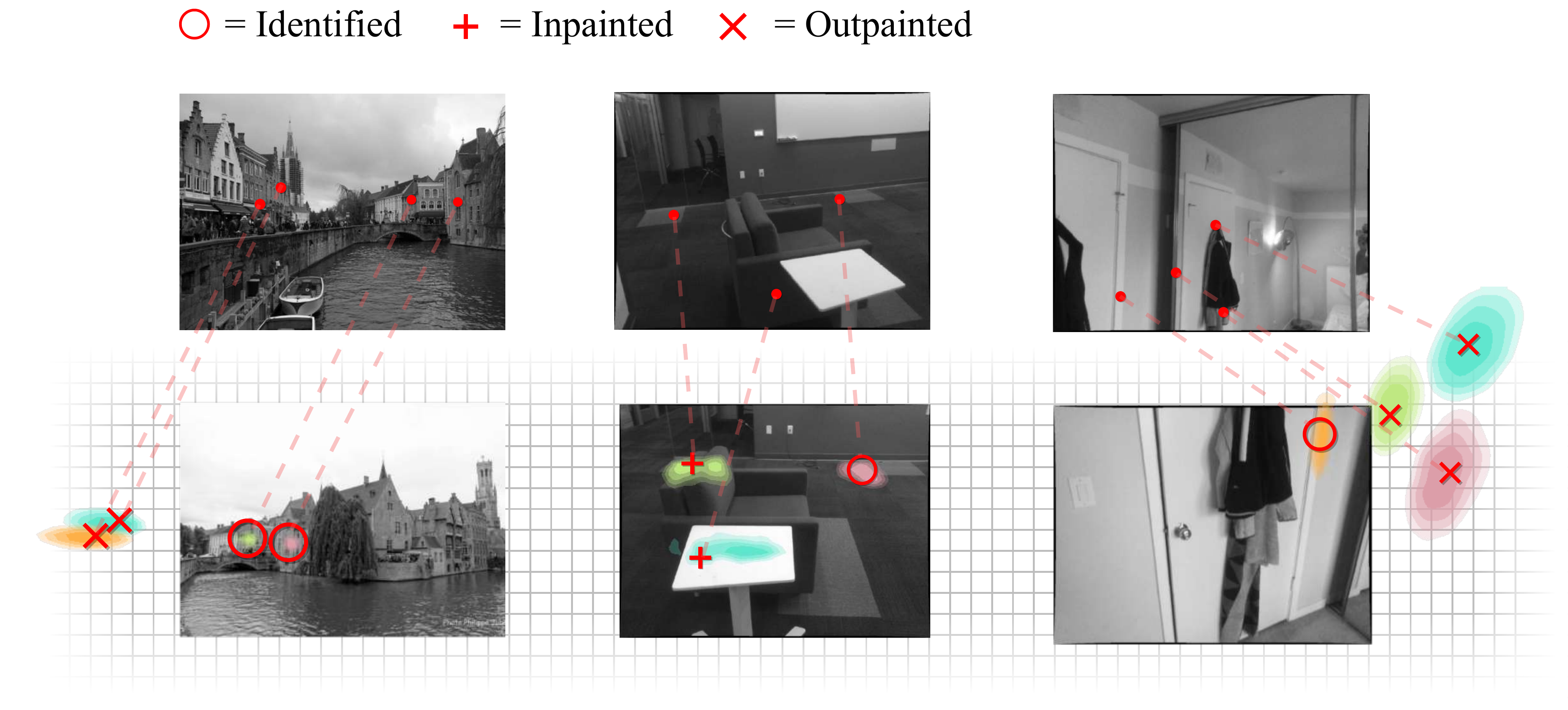}
    \caption{ \textbf{Visual correspondence hallucination.} Our network, called
      \neurhal, takes as input a pair of partially overlapping source and target
      images and a set of keypoints detected in the source image, and outputs
      for each keypoint a probability distribution over its correspondent's
      location in the target image. When the correspondent is actually visible,
      its location can be \emph{identified}; when it is not, its location must
      be \emph{hallucinated}.  Two types of hallucination tasks can be
      distinguished: 1) if the correspondent is occluded, its location has to be \emph{inpainted}; 2) if it is outside the field of view of the target image, its location needs to be \emph{outpainted}.  \neurhal generalizes to scenes not seen during training: For each of these three pairs of source/target images coming from the test scenes of ScanNet~\cite{ScanNet} and MegaDepth~\cite{Megadepth}, we show (top row) the source image with a small subset of keypoints, and (bottom row) the target image with the probability distributions predicted by our network and the ground truth correspondents: $\color{red}\boldsymbol{\circ}$ for the identified correspondents, $\color{red}\boldsymbol{+}$ for the inpainted ones, and $\color{red}\boldsymbol{\times}$ for the outpainted correspondents.
      %
}
    \label{fig:teaser}
  \end{center}
  \vspace{-0.3cm}
\end{figure}

\section{Related work}\label{sec:related}

To the best of our knowledge, aiming at hallucinating visual correspondences has
never been done but the related fields of local feature description and matching
are immensely vast, and we focus here only on recent learning-based approaches.

\paragraph{Learning-based local feature description.} Using deep neural networks
to learn to compute local feature descriptors have shown to bring significant
improvements in invariance to viewpoint and illumination changes compared to
handcrafted methods~\cite{Survey1, Survey2, Survey3, HPatches}. Most
  methods learn descriptors locally around pre-computed \emph{covisible}
  interest regions in both images~\cite{LIFT, SuperPoint, PNNet, ContextDesc},
  using convolutional-based siamese architectures trained with a contrastive
  loss~\cite{Gordo2016DeepIR, SchroffKP15,
  Balntas2016LearningLF,Radenovi2016CNNIR,HardNet, Simonyan2014LearningLF}, or
  using pose~\cite{Wang2020LearningFD,Zhou2021Patch2PixEP} or
self~\cite{Yang2021SelfsupervisedGP} supervision. To further improve the
performances, \cite{D2Net, R2D2} propose to jointly learn to detect and describe
keypoints in both images, while~\citet{S2DNet} only detects in one
image and densely matches descriptors in the other.

\paragraph{Learning-based local feature matching.} All the methods described in
the previous paragraph establish correspondences by comparing descriptors using
a simple operation such as a dot product. Thus the combination of such a simple
matching method with a siamese architecture inevitably produces outlier
correspondences, especially in non-covisible regions. To reduce the amount of
outliers, most approaches employ so-called Mutual Nearest Neighbor~(MNN)
filtering. However, it is possible to go beyond a simple MNN and learn to match
descriptors. Learning-based matching methods~\cite{OANet, NGRANSAC,
moo2018learning,sun2020acne,choy2020high, Choy2016UniversalCN} take as input
local descriptors and/or putative correspondences, and learn to output
correspondences probabilities. However, all these matching methods focus only on
predicting correctly covisible correspondences.

\paragraph{Jointly learning local feature description and matching.} Several
methods have recently proposed to jointly learn to compute and match
descriptors~\cite{SuperGlue,LoFTR,li2020dual,NCNet, SparseNCNet}. All these
methods use a siamese Convolutional Neural Network (CNN) to obtain dense local
descriptors, but they significantly differ regarding the way they establish
matches. They actually fall into two categories.  The first category of
methods~\cite{li2020dual,NCNet, SparseNCNet} computes a 4D correlation tensor
that essentially represents the scores of all the possible correspondences. This
4D correlation tensor is then used as input to a second network that learns to
modify it using soft-MNN and 4D convolutions. Instead of summarizing all the
information into a 4D correlation tensor, the second category of
methods~\cite{SuperGlue,LoFTR} rely on
Transformers~\cite{Vaswani2017AttentionIA, Dosovitskiy2020AnII,
Ramachandran2019StandAloneSI, caron2021emerging, Cordonnier2020OnTR,
Zhao2020ExploringSF,Katharopoulos2020TransformersAR} to let the descriptors of
both images communicate and adapt to each other. All these methods again focus
on identifying correctly covisible correspondences and consider non-covisible
correspondences as noise. While our architecture is closely related to the
second category of methods as we also rely on Transformers, the motivation for
using it is quite different since it is our goal of hallucinating
correspondences that calls for a non-siamese architecture~(see
Sec.\ref{sec:our}). 

\paragraph{Visual content hallucination.} \cite{Yang2019ExtremeRP}
proposes to hallucinate the content of RGB-D scans to perform relative pose
estimation between two images. More recently~\cite{Chen2021WideBaselineRC}
regresses distributions over relative camera poses for spherical images using
joint processing of both images. The work of~\cite{Yang2020ExtremeRP,
Qian2020Associative3DVR, Jin2021PlanarSR} shows that employing a
\emph{hallucinate-then-match} paradigm can be a reliable way of recovering 3D
geometry or relative pose from sparsely sampled images. In this work, we focus
on the problem of \emph{correspondence} hallucination which unlike previously
mentioned approaches does not aim at recovering explicit visual content or
directly regressing a camera pose.
Perhaps closest to our goal
is \citet{Cai2021Extreme} that seeks to estimate a
relative rotation between two non-overlapping images by learning to reason about
``hidden'' cues such as direction of shadows in outdoor scenes, parallel lines
or vanishing points.


\section{Our approach}\label{sec:our} Our goal is to train a network that takes
as input a pair of partially overlapping source/target images and keypoints in
the source image, and outputs for each keypoint a probability distribution over
its correspondent's location in the target image plane, regardless of this
correspondent being visible, occluded, or outside the field of view. While the
problem of learning to find the location of a \emph{visible} correspondent
received a lot of attention in the past few years (see Sec.~\ref{sec:related}),
to the best of our knowledge, this paper is the first attempt of learning to
find the location of a correspondent regardless of this correspondent being
visible, occluded, or outside the field of view.  Since this learning task is
virgin territory, we first analyze its specific features below, before defining
a loss function and a network architecture able to handle these features.

\subsection{Analysis of the problem}\label{sec:analysis}

The task of finding the location of a correspondent regardless of this
correspondent being visible, occluded, or outside the field of view actually
leads to three different problems. Before stating those three problems, let us
first recall the notion of correspondent as it is the keystone of our problem. 

\paragraph{Correspondent.} Given a keypoint $\p_\S \in \Real^2$ in the source
image $\I_\S$, its depth $d_\S\in\Real^+$, and the relative camera pose
$\R_{\T\S}\in\SO$, $\t_{\T\S}\in\Real^3$ between the coordinate systems of
$\I_\S$ and the target image $\I_\T$, the \emph{correspondent} $\p_\T \in \Real^2$ of $\p_\S$ in
the target image plane is obtained by warping $\p_\S$: $\p_\T \defeq \omega
\parr{d_\S, \p_\S,\R_{\T\S},\t_{\T\S}}\defeq\K_\T{\pi\parr{d_\S \R_{\T\S}
\K_\S^{-1}\p_\S+\t_{\T\S}}}$, where $\K_\S$ and $\K_\T$ are the camera calibration
matrices of source and target images
and $\pi\parr{\u}\defeq\croch{\u_x/\u_z,\u_y/\u_z, 1}^\T$ is the projection
function.  In a slight abuse of notation, we do not distinguish a homogeneous 2D
vector from a non-homogeneous 2D vector. Let us highlight that the correspondent
$\p_\T$ of $\p_\S$ may not be \emph{visible}, \ie it may be occluded or outside
the field of view.

\paragraph{Identifying the correspondent.} In the case where a network has to
establish a correspondence between a keypoint $\p_\S$ in $\I_\S$ and its
\emph{visible} correspondent $\p_\T$ in $\I_\T$, standard approaches, such as
comparing a local descriptor computed at $\p_S$ in $\I_\S$ with local
descriptors computed at detected keypoints in $\I_\T$, are applicable to
\emph{identify} the correspondent $\p_\T$. 


\paragraph{Outpainting the correspondent.} When $\p_\T$ is outside the field of
view of $\I_\T$,  there is nothing to identify, \ie neither can $\p_\T$ be
detected as a keypoint nor can a local descriptor be computed at that location.
Here the network first needs to identify correspondences in the region where
$\I_\T$ overlaps with $\I_\S$ and realize that the correspondent $\p_\T$ is
outside the field of view to eventually \emph{outpaint} it (see
Fig.~\ref{fig:teaser}). We call this operation "outpainting  the correspondent"
as the network needs to predict the location of $\p_\T$ outside the field of
view of $\I_\T$.

\paragraph{Inpainting the correspondent.} When $\p_\T$ is occluded in $\I_\T$,
the problem is even more difficult since local features can be computed at that
location but will not match the local descriptor computed at $\p_\S$ in $\I_\S$.
As in the outpainting case, the network needs to identify correspondences in the
region where $\I_\T$ overlaps with $\I_\S$ and realize that the correspondent
$\p_\T$ is occluded to eventually \emph{inpaint} the correspondent $\p_\T$ (see
Fig.~\ref{fig:teaser}). We call this operation "inpainting  the correspondent"
as the network needs to predict the location of $\p_\T$ behind the occluding
object.

Let us now introduce a loss function and an architecture that are able to unify
the identifying, inpainting and outpainting tasks.

\subsection{Loss function}\label{sec:loss}

The distinction we made between the identifying, inpainting and outpainting
tasks come from the fact that the source image $\I_\S$ and the target image
$\I_\T$ are the projections of the same 3D environment from two different camera
poses.  In order to integrate this idea and obtain a unified correspondence
learning task, we rely on the \emph{Neural Reprojection Error}~(NRE)
 introduced by \cite{NeuralReprojection}.
In order to properly present the NRE, we first recall the notion of \emph
{correspondence map}.

\paragraph{Correspondence map.} Given $\I_\S$, $\I_\T$ and a keypoint $\p_\S$ in
the image plane of $\I_\S$, the \emph{correspondence map} $\C_\T$ of $\p_\S$ in
the image plane of $\I_\T$ is a 2D tensor of size $H_\C \times W_\C$ such that
$\C_\T\parr{\p_\T}\defeq p\parr{\p_\T|\p_\S, \I_\S, \I_\T}$ is the likelihood of
$\p_\T$ being the correspondent of $\p_\S$. The likelihood can only be evaluated
for $\p_\T\in\Omega_{\C_\T}$ where $\Omega_{\C_\T}$ is the set of all the pixel
locations in $\C_\T$. Here, we implicitly defined that the likelihood of $\p_\T$
falling outside the boundaries of $\C_{\T}$ is zero. In practice, a
correspondence map $\C_\T$ is implemented as a neural network that takes as
input $\p_\S$, $\I_\S$ and $\I_\T$, and outputs a softmaxed 2D tensor.  A
correspondence map $\C_\T$ may not have the same number of lines and columns
than $\I_\T$ especially when the goal is to outpaint a correspondence.  Thus, in
the general case, to transform a 2D point from the image plane of $\I_\T$ to the
correspondence plane of $\C_\T$, we will need another affine transformation matrix
$\K_\C$.  Let us highlight that this likelihood is obtained using the visual
content of $\I_\S$ and $\I_\T$ only.

\paragraph{Neural Reprojection Error.}
The NRE \cite{NeuralReprojection} is a loss function that warps a keypoint $\p_\S$ into the image plane of $\I_\T$ and evaluates the negative log-likelihood at this location. In our context, the NRE can be written as:
\begin{align}
  \text{NRE} \parr{\p_{\S},\C_\T,\R_{\T\S},\t_{\T\S}, d_\S}\defeq -\ln\C_{\T}\parr{\x_\T} \text{ where } \x_\T=\K_\C\omega \parr{d_\S, \p_{\S},\R_{\T\S},\t_{\T\S}} \>.  \label{eq:NRE}
\end{align}
In general, $\x_\T$ does not have integer coordinates and the notation
$\ln\C_{\T}\parr{\x_\T}$ corresponds to performing a bilinear interpolation
\emph{after} the logarithm. For more details concerning the derivation of the
NRE, the reader is referred to~\citet{NeuralReprojection}.

The NRE provides us with a framework to learn to identify, inpaint or outpaint
the correspondent of $\p_\S$ in $\I_\T$ in a unified manner since
Eq.~\eqref{eq:NRE} is differentiable \wrt $\C_{\T}$ and there is no assumption
regarding covisibility. The main difficulty to overcome is the definition of a
network architecture able to output a consistent $\C_{\T}$ being given only
$\p_\S$, $\I_\S$ and $\I_\T$ as inputs,  \ie the network must figure out whether
the correspondent of $\p_\S$ in $\I_\T$ can be identified or has to be inpainted
or outpainted.

%

\subsection{Network architecture}\label{sec:architecture}

The analysis from Sec.~\ref{sec:analysis} and the use of the NRE as a loss
(Sec.~\ref{sec:loss}) call for:\\
• a non-siamese architecture to be able to link the information from $\I_\S$
with the information from $\I_\T$ to \emph{outpaint} or \emph{inpaint} the
correspondent if needed;\\
• an architecture that outputs a matching score for all the possible locations in $\I_\T$ as well as locations beyond the field of view of $\I_\T$ as the network could decide to identify, inpaint or outpaint a correspondent at these locations.

To fulfill these requirements, we propose the following: Our network takes as
input $\I_\S$ and $\I_\T$ as well as a set of keypoints
$\curl{\p_{\S,n}}_{n=1...N}$ in the source image plane of $\I_\S$. A siamese CNN
backbone is applied to $\I_\S$ and $\I_\T$ to produce compact dense local
descriptor maps $\H_\S$ and $\H_\T$.  In order to be able to \emph{outpaint}
correspondents in the target image plane, we pad $\H_\T$ with a learnable fixed
vector $\boldsymbol{\lambda}$. This padding step allows to \emph{initialize}
descriptors at locations outside the field of view of $\I_\T$. We note $\gamma$
the relative output-to-input correspondence map resolution ratio.

The dense descriptor maps $\H_\S$ and $\H_{\T,\text{pad}}$, and the keypoints
$\curl{\p_{\S,n}}_{n=1...N}$ are then used as inputs of a cross-attention-based
backbone $\mathcal{F}$ with positional encoding. This part of the network
outputs a feature vector $\d_{\S,n}$ for each keypoint $\p_{\S,n}$ and dense
feature vectors $\D_{\T,\text{pad}}$ of the size of $\H_{\T,\text{pad}}$. This
cross-attention-based backbone allows the local descriptors $\H_\S$ and
$\H_{\T,\text{pad}}$ to \emph{communicate} with each other. Thus, during
training, the network will be able to leverage this ability to communicate, to
learn to \emph{hallucinate} peaked \emph{inpainted}
and \emph{outpainted} correspondence maps.

\begin{wrapfigure}{r}{0.6\linewidth}
\centering
\includegraphics[width=\linewidth]{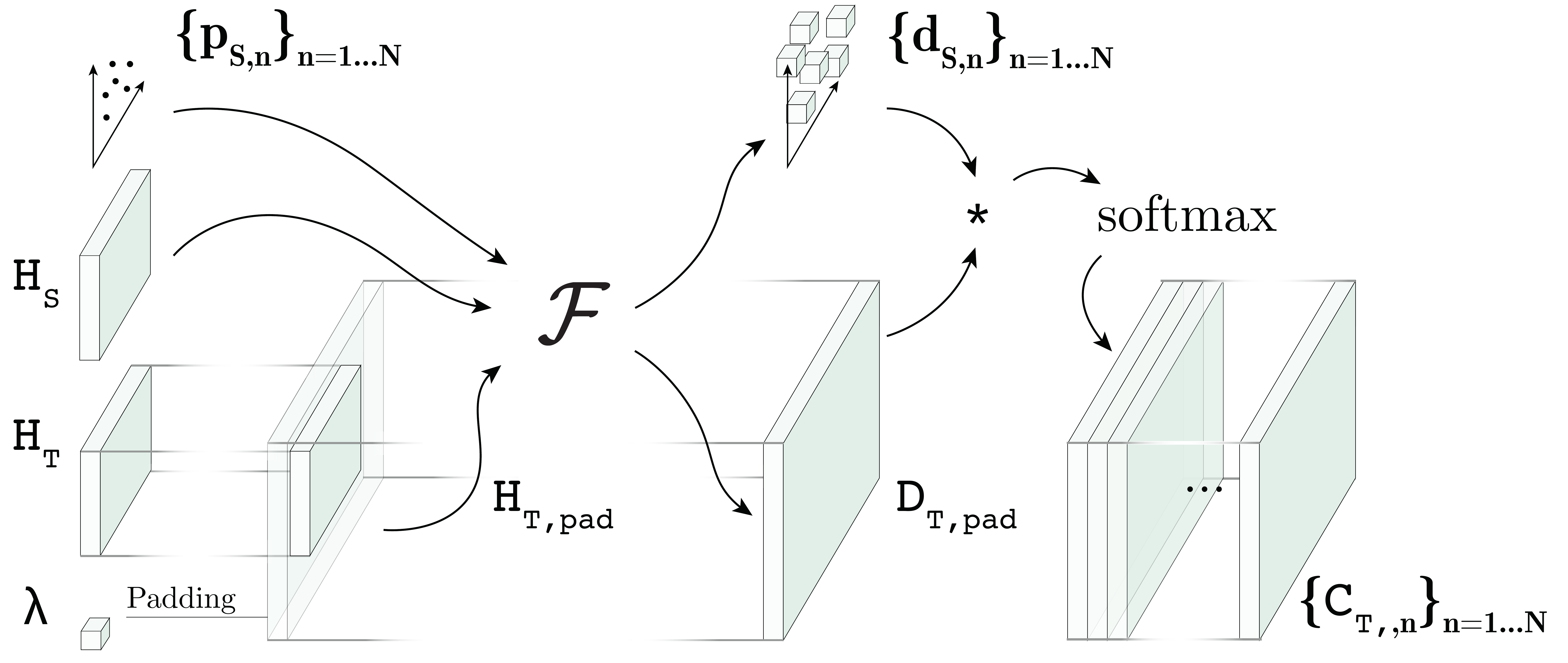}
\caption{
\textbf{Overview of NeurHal:} See text for details.}
\vspace{-0.4cm}
\label{fig:overview_neurhal}
\end{wrapfigure}

The correspondence map $\C_{\T,n}$ of $\p_{\S,n}$ in the image plane of $\I_\T$
is computed by applying a 1$\times$1 convolution to $\D_{\T,\text{pad}}$ using
$\d_{\S,n}$ as filter, followed by a 2D softmax. 

An overview of our architecture, that we call \neurhal, is presented in
Fig.~\ref{fig:overview_neurhal}. In practice, in order to keep the required
amount of memory and the computational time reasonably low, the correspondence
maps $\curl{\C_{\T,n}}_{n=1...N}$ have a low resolution, \ie for a target image
of size $640\times480$, we use a CNN with an effective stride of $s=8$ and
consequently the resulting correspondence maps (with $\gamma=50\%$) are of size
$160\times120$. Producing low resolution correspondence maps prevents \neurhal
from predicting accurate correspondences. But as we show in the experiments,
this low resolution is sufficient to hallucinate correspondences and have an
\emph{affirmative answer} to both questions: (i) can we derive a network
architecture able to learn to hallucinate correspondences? and (ii) is
correspondence hallucination beneficial for absolute pose estimation? Thus, we
leave the question of the accuracy of hallucinated correspondences for future
research.
%
%
Additional details concerning the architecture are provided in
Sec.~\ref{sec:architecture_details} of the appendix.

\subsection{Training-time}\label{sec:training} Given a pair of partially
overlapping images $\parr{\I_\S,\I_\T}$, a set of keypoints with ground truth
depths $\curl{\p_{\S,n},d_{\S,n}}_{n=1...N}$ as well as the ground truth
relative camera pose $\parr{ \R_{\T\S},\t_{\T\S}}$, the corresponding sum of NRE
terms (Eq.~\ref{eq:NRE}) can be minimized \wrt the parameters of the network
that produces the correspondence maps. Thus, we train our network using
stochastic gradient descent and early stopping by providing pairs of overlapping
images along with the aforementioned ground truth information. Let us also
highlight that there is no distinction in the training process between the
identifying, inpainting and outpainting tasks since the only thing our network
outputs are correspondence maps. Moreover there is no need for labeling
keypoints with ground truth labels such as "identify/visible",
"inpaint/occluded" or "outpaint/outside the field of view".  Additional
information concerning the training are provided in
Sec.~\ref{sec:dataset_details} of the appendix.

\subsection{Test-time} 
At test-time, our network only requires a pair of
partially overlapping images $\parr{\I_\S,\I_\T}$ as well as keypoints
$\curl{\p_{\S,n}}_{n=1...N}$ in $\I_\S$, and outputs a correspondence map
$\C_{\T,n}$ in the image plane of $\I_\T$ for each keypoint, regardless of its
correspondent being visible, occluded or outside the field of view.

\begin{figure}[t]
 \begin{center}
    \centering
    \raisebox{0.45in}{\rotatebox{90}{\small{ScanNet}}}
    \includegraphics[height=3.3cm]{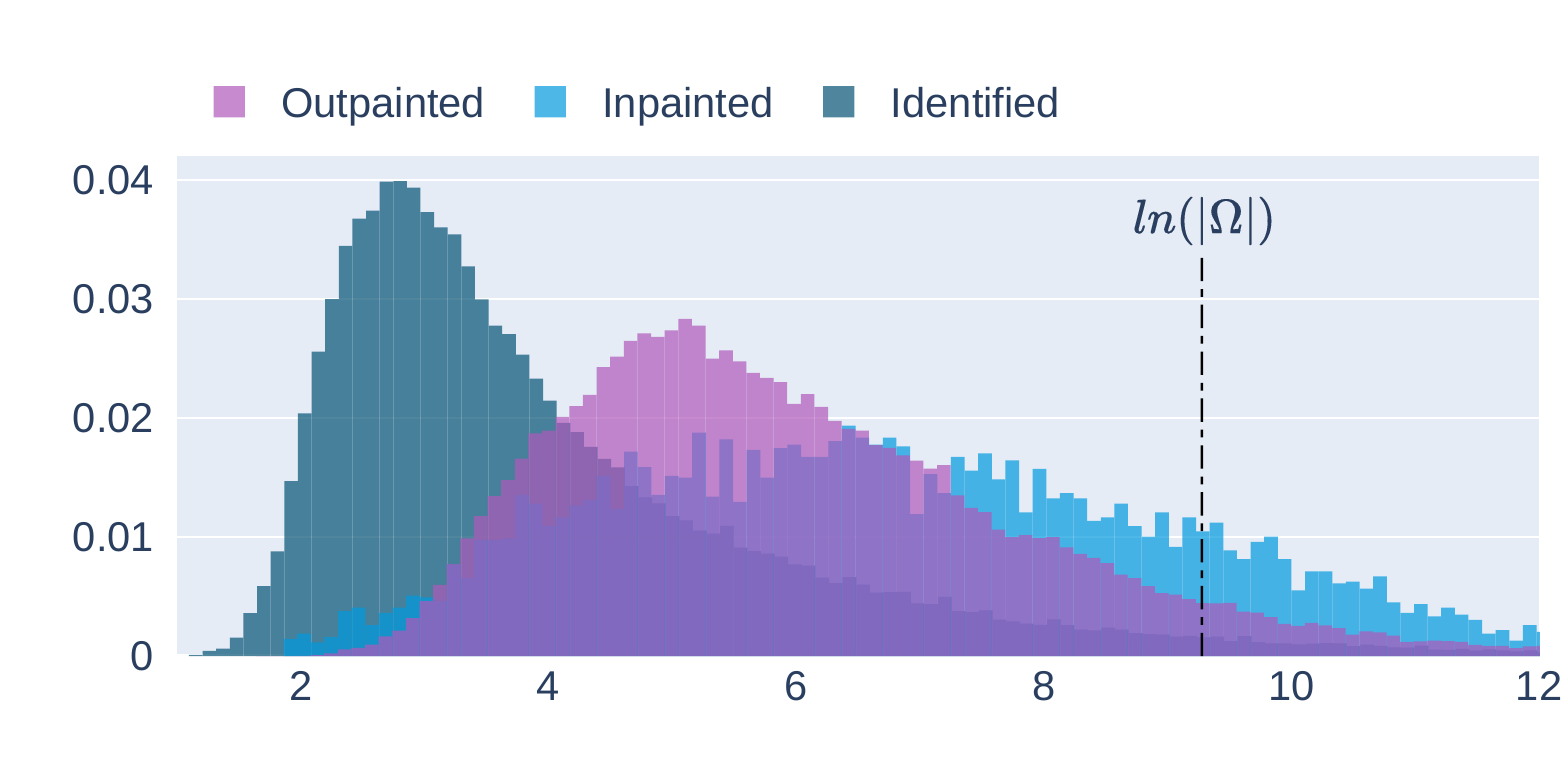}
    \includegraphics[height=3.3cm]{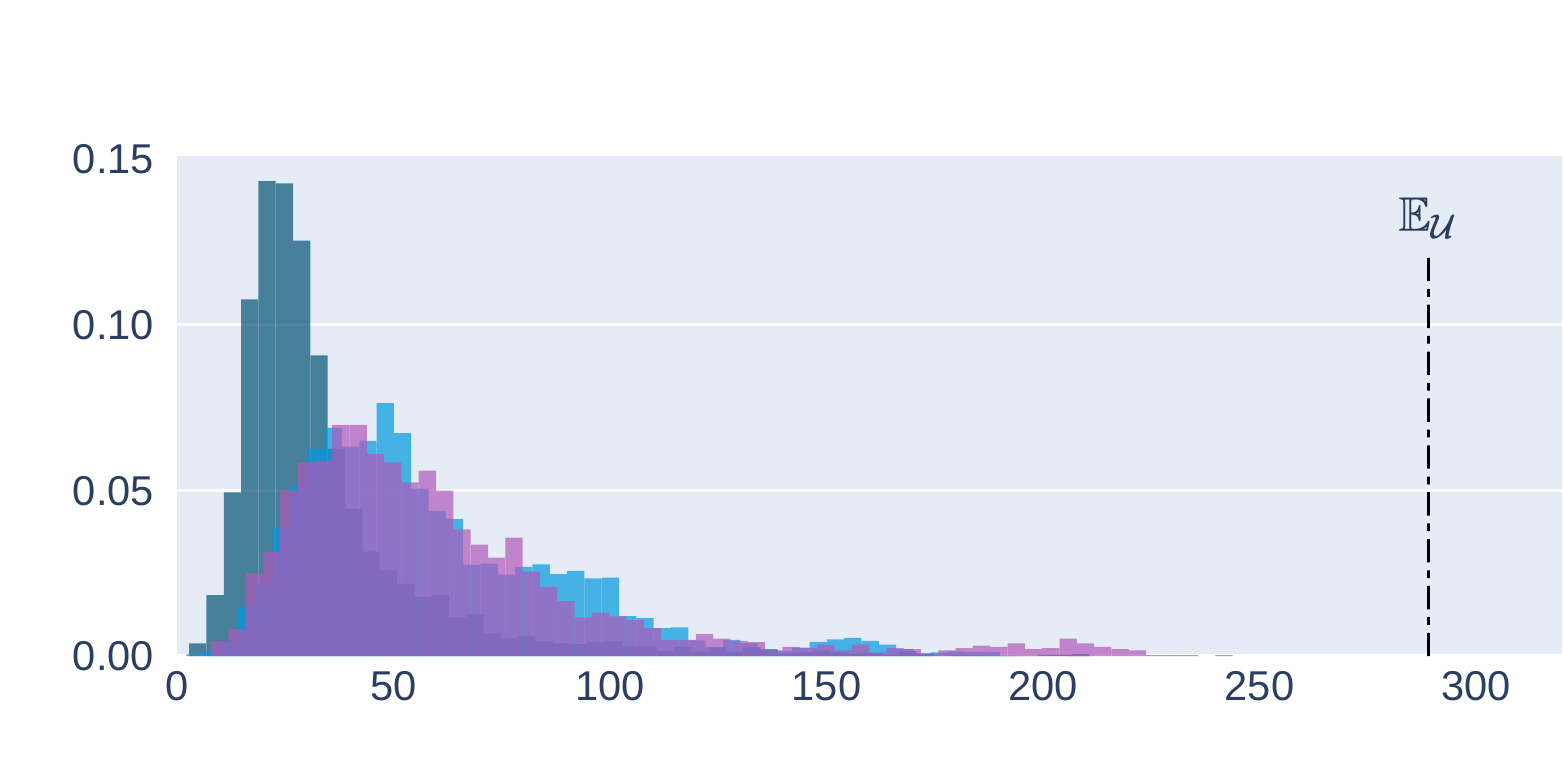}\\
    \vspace{-0.35cm}
    \raisebox{0.35in}{\rotatebox{90}{\small{Megadepth}}}
    \includegraphics[height=3.3cm]{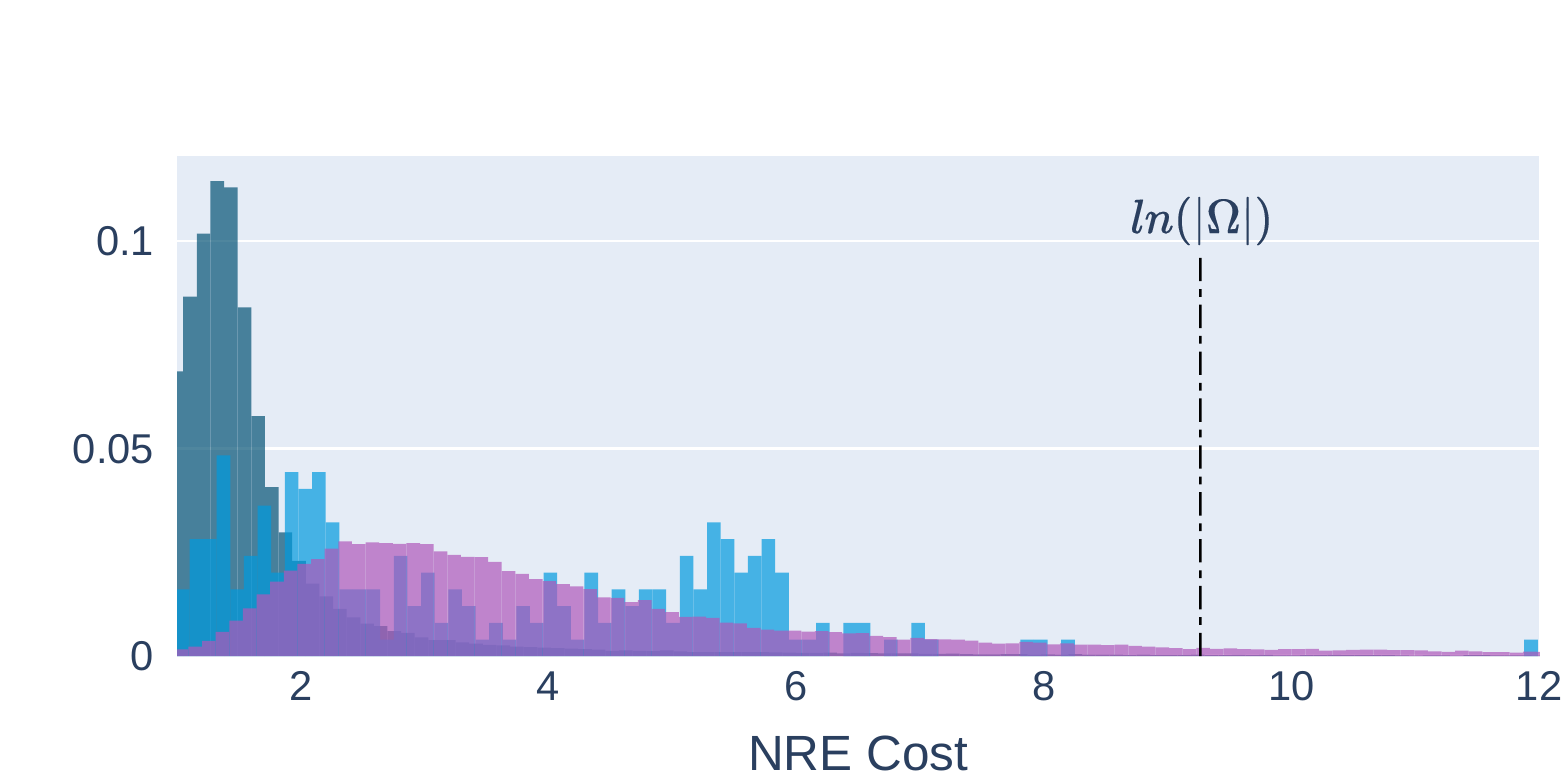}
    \includegraphics[height=3.3cm]{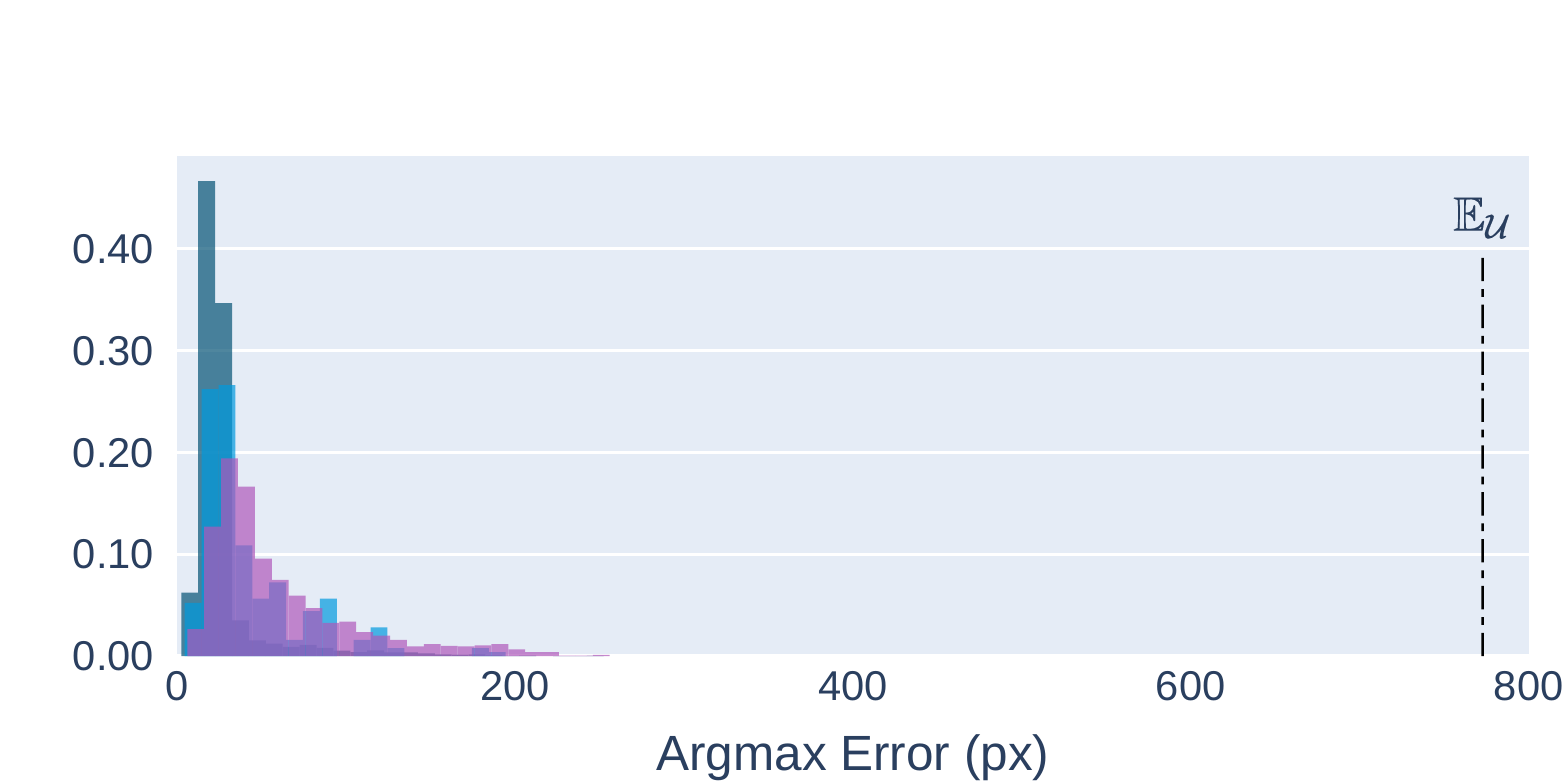}
    \vspace{0.1cm}
    \caption{ \textbf{Evaluation of the ability of NeurHal to hallucinate
      correspondences on the test scenes of ScanNet and MegaDepth.}  (left) Histograms of the NRE (see Eq.~\ref{eq:NRE}) for each task (identifying, outpainting, inpainting), computed on correspondence maps produced by \neurhal . The value $\ln|\Omega_{\C_\T}|$ is the NRE of a uniform correspondence map.  (right) Histograms of the errors between the argmax (mode) of a correspondence map and the ground truth correspondent's location, for each task.  The value $\mathbb{E}_\mathcal{U}$ is the average error of a random prediction. 
}
    \label{fig:histograms}
  \end{center}
  \vspace{-0.4cm}
\end{figure}

\section{Experiments}\label{sec:experiments}
In these experiments, we seek to answer two questions: 1) "Is the proposed
\neurhal approach presented in Sec.~\ref{sec:our} capable of
hallucinating correspondences?" and 2) "In the context of absolute camera pose
estimation, does the ability to hallucinate correspondences bring further
robustness?".

\subsection{Evaluation of the ability to hallucinate correspondences}
\label{sec:evaluation} We evaluate the ability of our network to hallucinate
correspondences on four datasets: the indoor datasets ScanNet~\cite{ScanNet} and
NYU~\cite{NYU}, and the outdoor datasets MegaDepth~\cite{Megadepth} and
ETH-3D~\cite{ETH3D}. 
For the indoor setting (outdoor setting, respectively), we train \neurhal on
ScanNet (Megadepth, respectively) on the
training scenes as described in
Sec.~\ref{sec:training}, and evaluate it on the \emph{disjoint} set of
validation scenes. Thus, all the qualitative and quantitative results presented
in this section cannot be ascribed to scene memorization.
%
For each dataset, we run predictions over $2,500$ source and target image pairs
sampled from the test set, with overlaps between $2\%$ and $80\%$. For every
image pair, we also feed as input to \neurhal~keypoints in the source image.
These keypoints have known ground truth correspondents in the target image and
labels (visible, occluded, outside the field of view) that we use to evaluate
the ability of our network to hallucinate correspondences. For more details on
the settings of our experiment see Sec.~\ref{sec:dataset_details}.  For this
experiment, we use $\gamma=50\%$.

We report in Fig.~\ref{fig:histograms} two histograms computed over more than
one million keypoints for each task we seek to validate: identification,
inpainting, and outpainting.  The first histogram Fig.~\ref{fig:histograms}
(left) is obtained by evaluating for each correspondence map the NRE
cost~(Eq.~\ref{eq:NRE}) at the ground truth correspondent's location.  In order
to draw conclusions, we also report the negative log-likelihood of a uniform
correspondence map~($\ln|\Omega_{\C_\T}|$).  We find that for each task and for
both datasets, the predicted probability mass lies significantly below
$\ln|\Omega_{\C_\T}|$, which demonstrates \neurhal's ability to perform
identification, inpainting and outpainting.  On ScanNet, we also observe that
identification is a simpler task than outpainting while inpainting is the
hardest task: On average, the NRE cost of inpainted correspondents is higher
than the average NRE cost of outpainted correspondents, which indicates the
predicted correspondence maps are less peaked for inpainting than they are for
outpainting.  This corroborates what we empirically observed on qualitative
results in Fig.~\ref{fig:teaser}, and supports our analysis in
Sec.~\ref{sec:analysis}. On Megadepth, outpainting and inpainting histograms
have a similar shape which does not reflect the previous statement, but we
believe this is due to the fact that inpainting labels are noisy for this
dataset, as explained in Sec.~\ref{sec:dataset_details}.


On the right histogram of Fig.~\ref{fig:histograms}, we report the distribution
of the distance between the argmax of a correspondence map and the ground truth
correspondent's location.  We also report the average error of a random
prediction.  We find the histogram mass lies significantly to the left of the
random prediction average error, indicating our model is able to place modes
correctly in the correspondence maps, regardless of the task at hand.  On
ScanNet, we observe that the inpainting and outpainting histograms
are very similar, indicating the predicted argmax is equally good for both
tasks. As mentioned above, the correspondence maps produced by \neurhal have a
low resolution (see Sec.~\ref{sec:architecture}) which explains why the "argmax
error" is not closer to zero pixel.

\begin{figure}[t]
  \captionsetup[subfigure]{position=b}
    \centering
    \resizebox{0.75\columnwidth}{!}{
      \subcaptionbox{Inpainting - S}{
      \includegraphics[height=3.4cm]{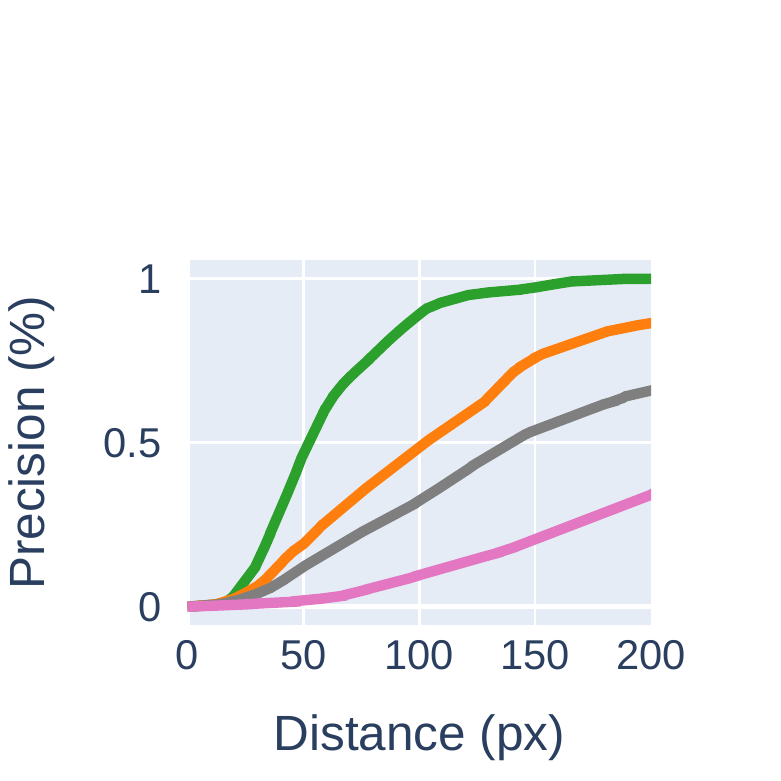}
    }
    \subcaptionbox{Outpainting - S}{
      \includegraphics[height=3.4cm]{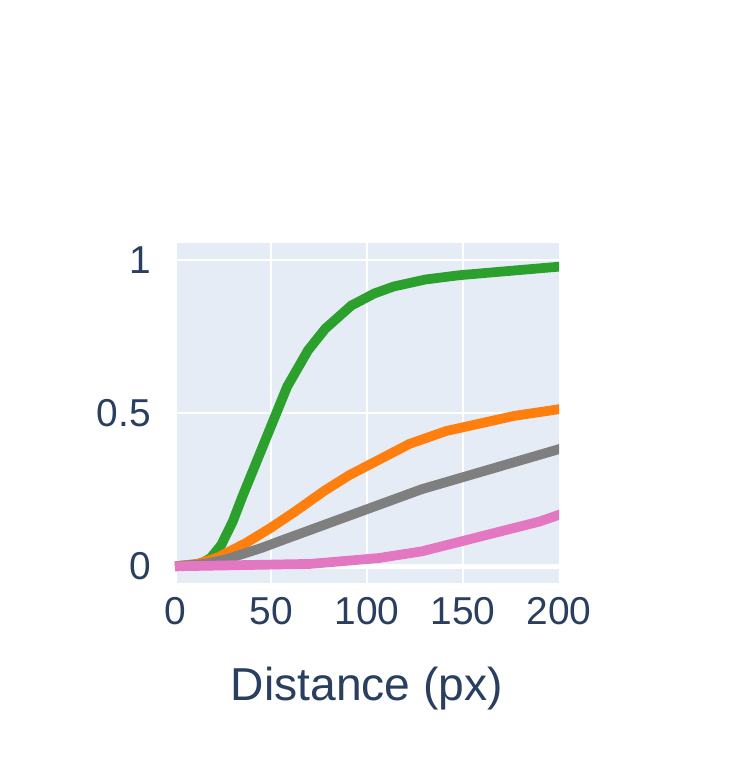}
    }
    \subcaptionbox{Inpainting - M}{
      \includegraphics[height=3.4cm]{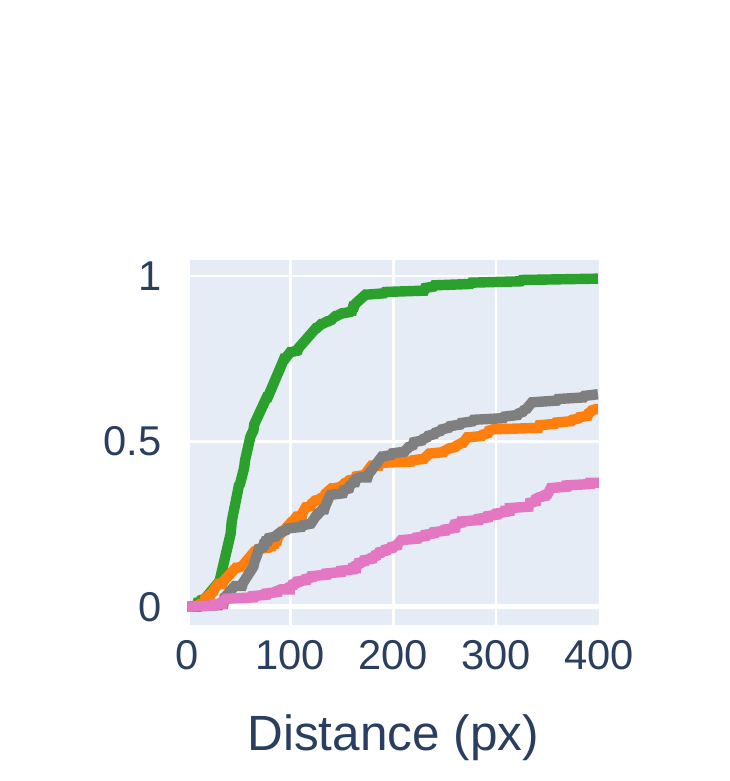}
    }
    \subcaptionbox{Outpainting - M}{
      \includegraphics[height=3.4cm]{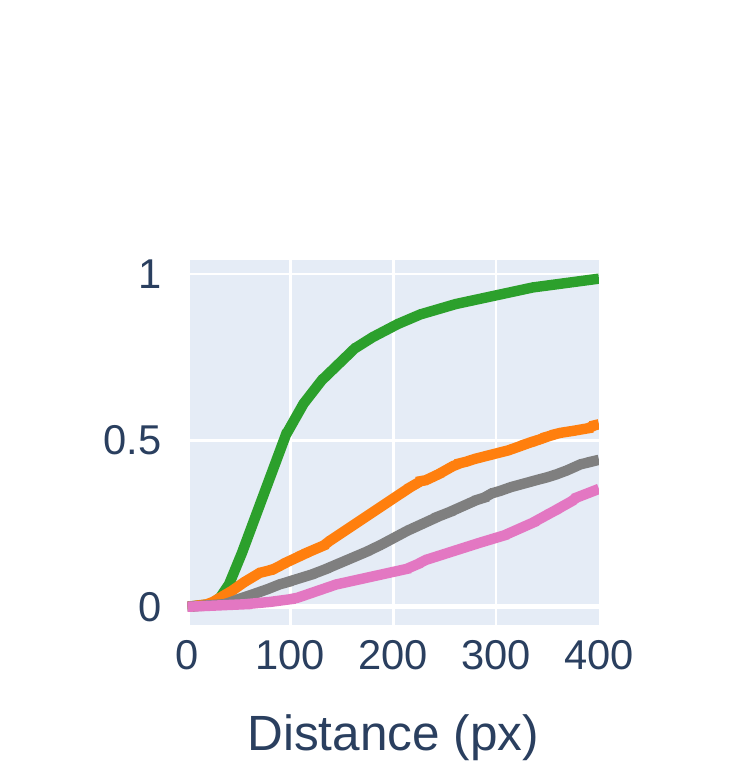}
      }
    }
    \raisebox{0.0cm}{
    \resizebox{0.2\columnwidth}{!}{
      \ra{1.2}
      \begin{tabular}[b]{l}\toprule
      \textbf{Method}\\ \midrule
      \textcolor{color_LoFTR}{\rule{0.5cm}{0.6mm}} LoFTR~\cite{LoFTR}\\
      \textcolor{color_DRCNet}{\rule{0.5cm}{0.6mm}} DRCNet~\cite{li2020dual}\\
      \textcolor{color_Mixed6e}{\rule{0.5cm}{0.6mm}} S2D~\cite{NeuralReprojection}\\
      \textcolor{color_S2DLoc}{\rule{0.5cm}{0.6mm}} \neurhal~\\
      \bottomrule
      \end{tabular}
      }
    }
  \caption{\textbf{Ability to hallucinate - comparison against state-of-the-art
    local feature matching methods on ScanNet (S) and Megadepth (M).} For each method, we
    report the percentage of keypoint's correspondents whose distance \wrt the
    ground truth location is lower than $x$ pixels, as a function of
  $x$, for (a-c) the inpainting task and (b-d) the outpainting
task.}
  \label{fig:hallucination_sota}
  \vspace{-0.6cm}
\end{figure}

In Fig.~\ref{fig:hallucination_sota}, we compare the hallucination performances
of \neurhal against state-of-the-art local feature matching methods. Since all
these local feature matching methods were designed and trained on pairs of
images with significant overlap to perform only {identification}, they obtain
poor {inpainting} results. Concerning the outpainting task, these methods seek
to find a correspondent within the image boundaries, consequently they cannot
outpaint correspondences and obtain very poor results.

\begin{figure}[t]
\centering
\rule{2.2cm}{0.3mm}~~ScanNet~~\rule{2.2cm}{0.3mm}
~~~~~
\rule{2.2cm}{0.3mm}~~Megadepth~~\rule{2.2cm}{0.3mm}
\hfill
\\
\raisebox{0.25in}{\rotatebox{90}{\small{\!\!\!$\text{-}\ln\C_\T$ ~~~~~~~~~~ Target ~~~~~~~ Source}}}
\includegraphics[width=0.95\textwidth]{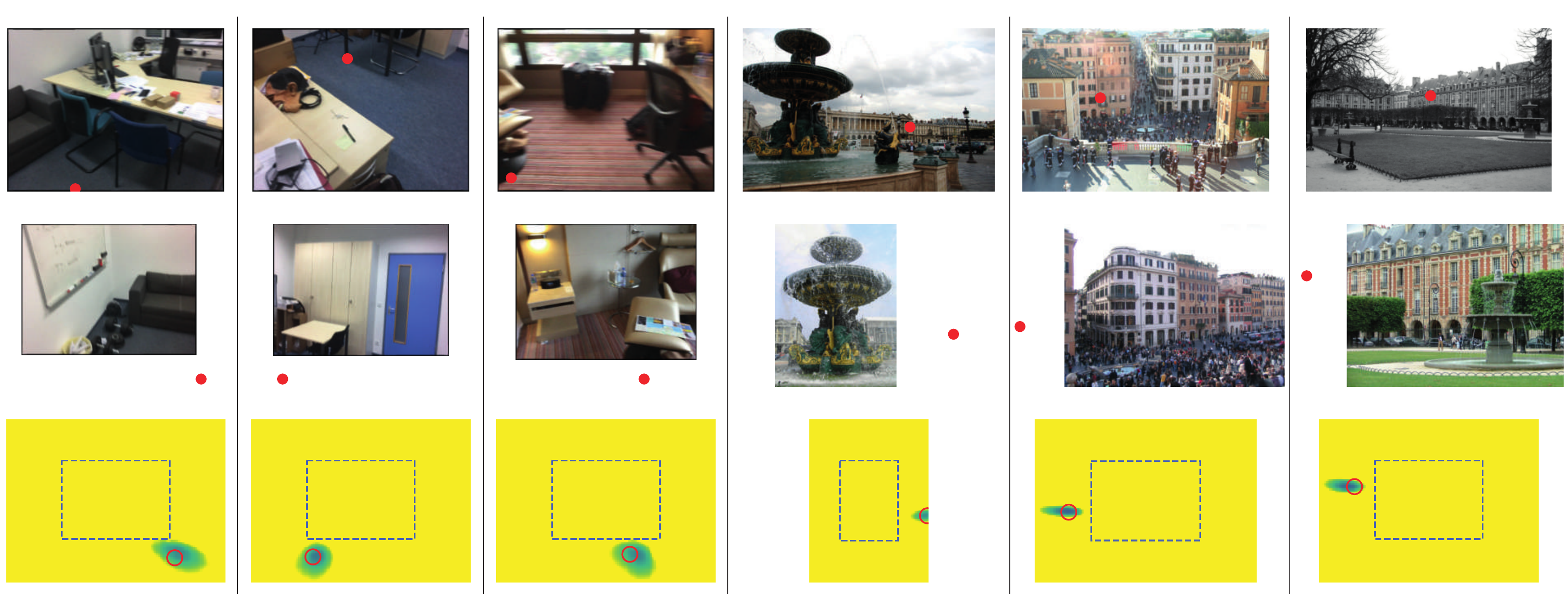}
\hrule
\raisebox{0.25in}{\,\rotatebox{90}{\small{\!\!\!$\text{-}\ln\C_\T$ ~~~~~~~~~~ Target ~~~~~~~ Source}}}
\includegraphics[width=0.95\textwidth]{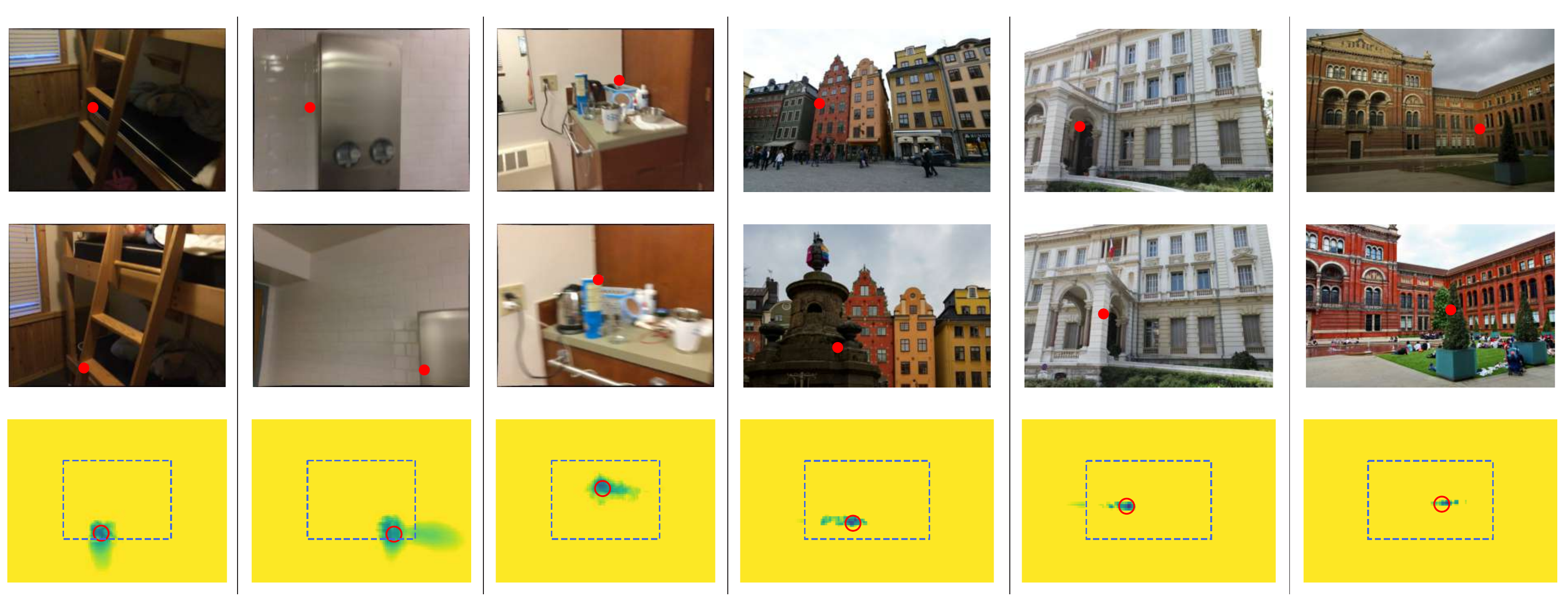}
\vspace{0.2cm}
\caption{\textbf{Ability to hallucinate - Qualitative inpainting/outpainting
results.}  To illustrate the ability of \neurhal to hallucinate correspondents, we display correspondence maps predicted by
\neurhal~on image pairs (captured in scenes that were not seen
at training-time): (top row) outpainting examples, (bottom row) inpainting
examples. In the source image, the red dot is a keypoint. In the target image
and in the (negative-log) correspondence map, the red dot represents the ground
truth keypoint's correspondent.  The dashed rectangles represent the borders of
the target images. More results on the NYU and ETH-3D datasets can be found in the
appendix~\ref{sec:generalization}.}
\label{fig:qualitative}
\vspace{-0.4cm}
\end{figure}

In Fig.~\ref{fig:qualitative} we show several qualitative inpainting/outpainting
results on ScanNet and MegaDepth datasets. In the appendix, we also report
qualitative results obtained on the NYU Depth dataset~(Fig.~\ref{fig:nyu}) and
on the ETH-3D dataset~(Fig.~\ref{fig:eth3D}).

These results allow us to conclude that \neurhal is able to hallucinate
correspondences with a strong generalization capacity. Additional experiments
concerning the ability to hallucinate correspondences are provided in
Sec.~\ref{sec:appendix_hallucination} as well as technical details regarding the
evaluation protocol in Sec.~\ref{sec:evaluation_details}.


\subsection{Application to absolute camera pose estimation}\label{sec:application}

In the previous experiment, we showed that our network is able to hallucinate
correspondences.  We now evaluate whether this ability helps improving the
robustness of an absolute camera pose estimator.  We run this evaluation on
the test set of ScanNet over 2,500 source and target image pairs captured in
scenes that were not used at training time. For each
source/target image pair, we employ \neurhal to produce correspondence
maps.  As in the previous experiment, we use $\gamma=50\%$. Given these
correspondence maps and the depth map of the source image, we estimate the
absolute camera pose between the target image and the source image using
the method proposed in~\citet{NeuralReprojection}.

\begin{figure}
  \captionsetup[subfigure]{position=b}
    \centering
      \includegraphics[width=0.4\textwidth]{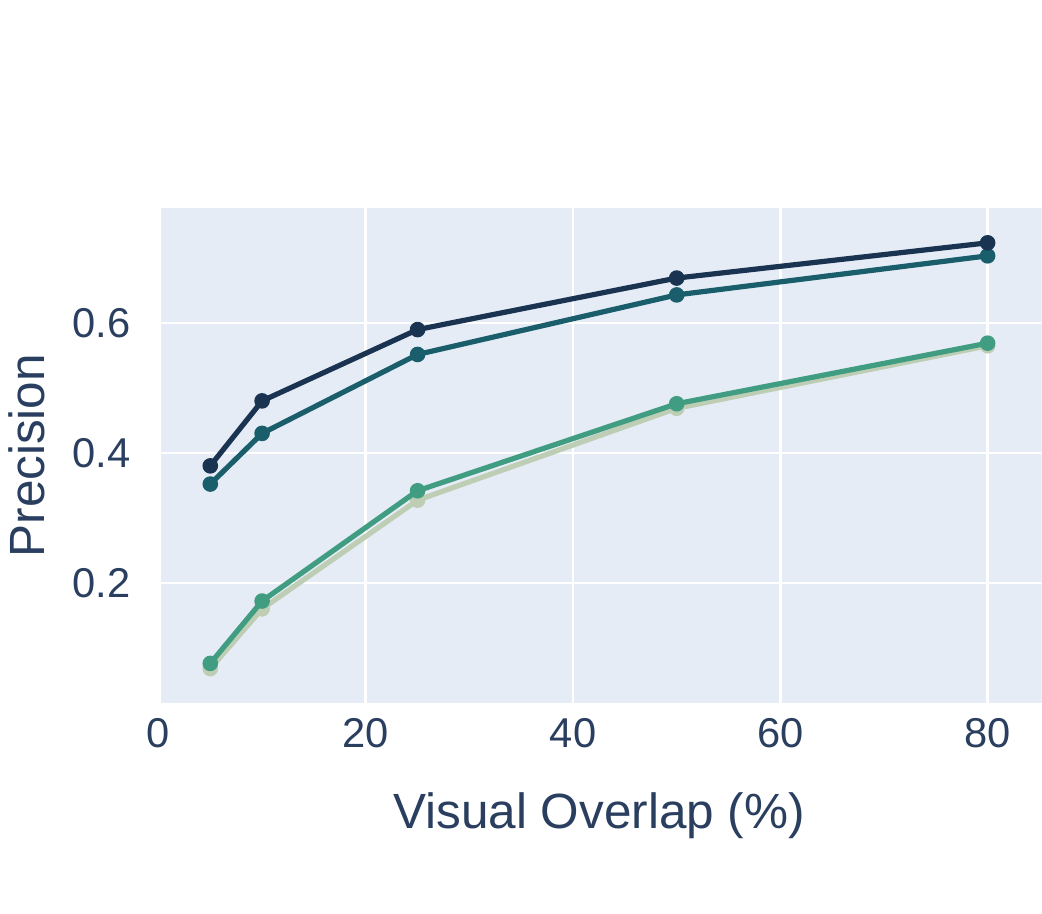}
      ~~~
      \raisebox{2.3cm}{
      \begin{adjustbox}{max width=0.45\textwidth}
      \ra{1.2}
        \begin{tabular}{cccc}%
        \toprule
        \multirow{2}{*}{Method} & \multicolumn{3}{c}{Training correspondences}\\
        \cmidrule(lr){2-4}
         & Identified & Inpainted & Outpainted\\
        \midrule
        \textcolor{color_VCH}{\rule{0.5cm}{0.6mm}} & \checkmark & &\\
        \textcolor{color_VCH_I}{\rule{0.5cm}{0.6mm}} & \checkmark & \checkmark & \\
        \textcolor{color_VCH_O}{\rule{0.5cm}{0.6mm}} & \checkmark & & \checkmark\\
        \textcolor{color_VCH_IO}{\rule{0.5cm}{0.6mm}} & \checkmark & \checkmark & \checkmark\\
        \bottomrule
      \end{tabular}%
      \end{adjustbox}
    }
\caption{\textbf{Ablation study - Impact of learning to hallucinate for absolute
  camera pose estimation.} We compare the influence of adding inpainting and
  outpainting ($\gamma=50\%$) tasks when training \neurhal. We report the percentage
  of camera poses being correctly estimated for image pairs having an overlap between $2\%$ and $x\%$, as a function of $x$, on
  ScanNet~\cite{ScanNet}, with thresholds for translation and rotation
  errors of $\tau_t=1.5m$ and $\tau_r=20.0\degree$.  Learning to hallucinate
  correspondences (especially outpainting) significantly improves the amount of correctly estimated poses.
}\label{fig:ablation_localization_data}
\vspace{-0.5cm}
\end{figure}

In Fig.~\ref{fig:ablation_localization_data}, we show the results of an ablation
study conducted on ScanNet.  In this study, we focus on the robustness of the
camera pose estimate for various combinations of training data, \ie we consider
a pose is "correct" if the rotation error is lower than 20 degrees and the
translation error is below 1.5 meters (see Sec.~\ref{sec:evaluation_details}). We find that training our network to
perform the three tasks (identification, inpainting, and outpainting) produces
the best results. In particular, we find that adding outpainting plays a
critical role in improving localization of low-overlap image pairs.  We also
find that learning to inpaint does not bring much improvement to the absolute
camera pose estimation.

\begin{figure}[t]
 \begin{center}
    \centering
    \subcaptionbox{$\tau_t=1.5m,\tau_r=20\degree$}{
      \includegraphics[width=0.35\textwidth]{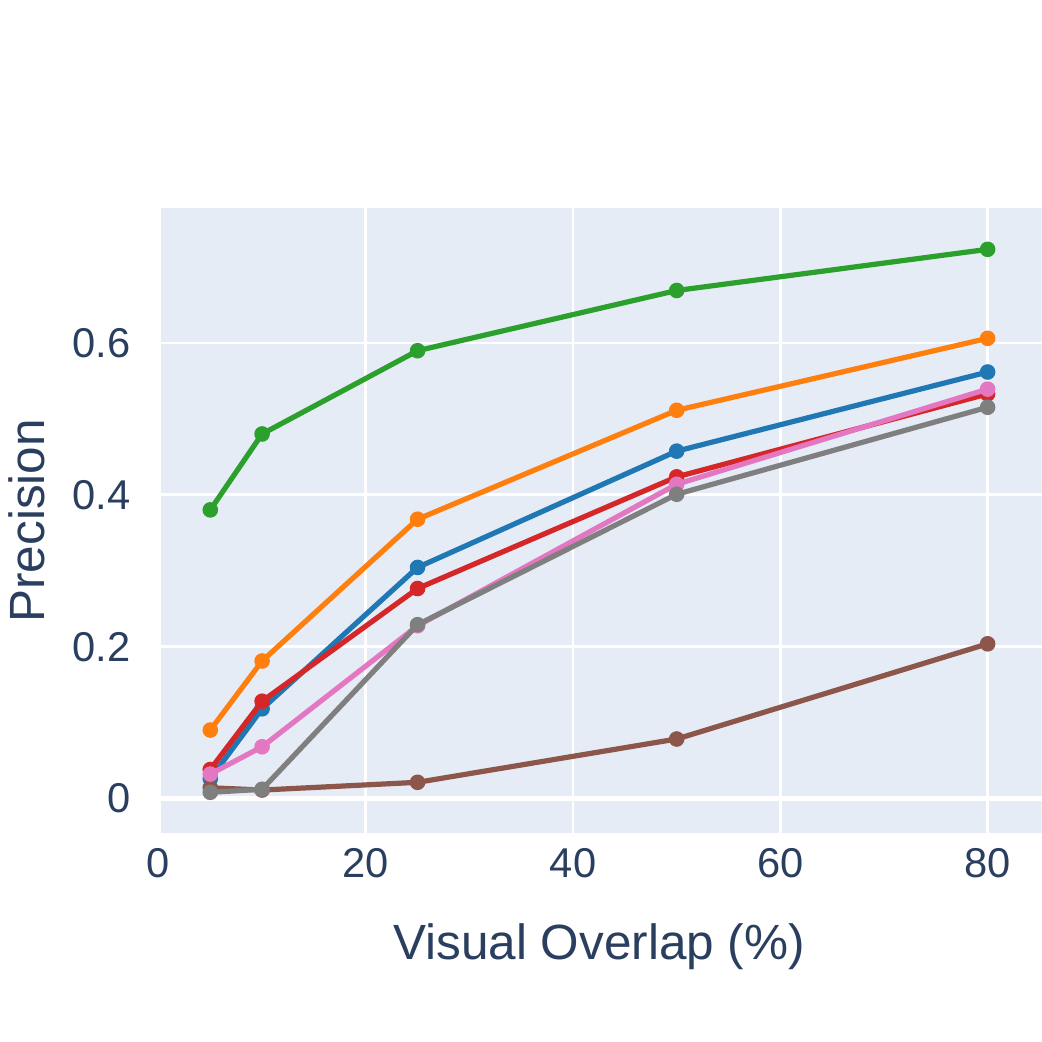}
    }
    \subcaptionbox{$\tau_t=1.0m,\tau_r=15\degree$}{
      \includegraphics[width=0.35\textwidth]{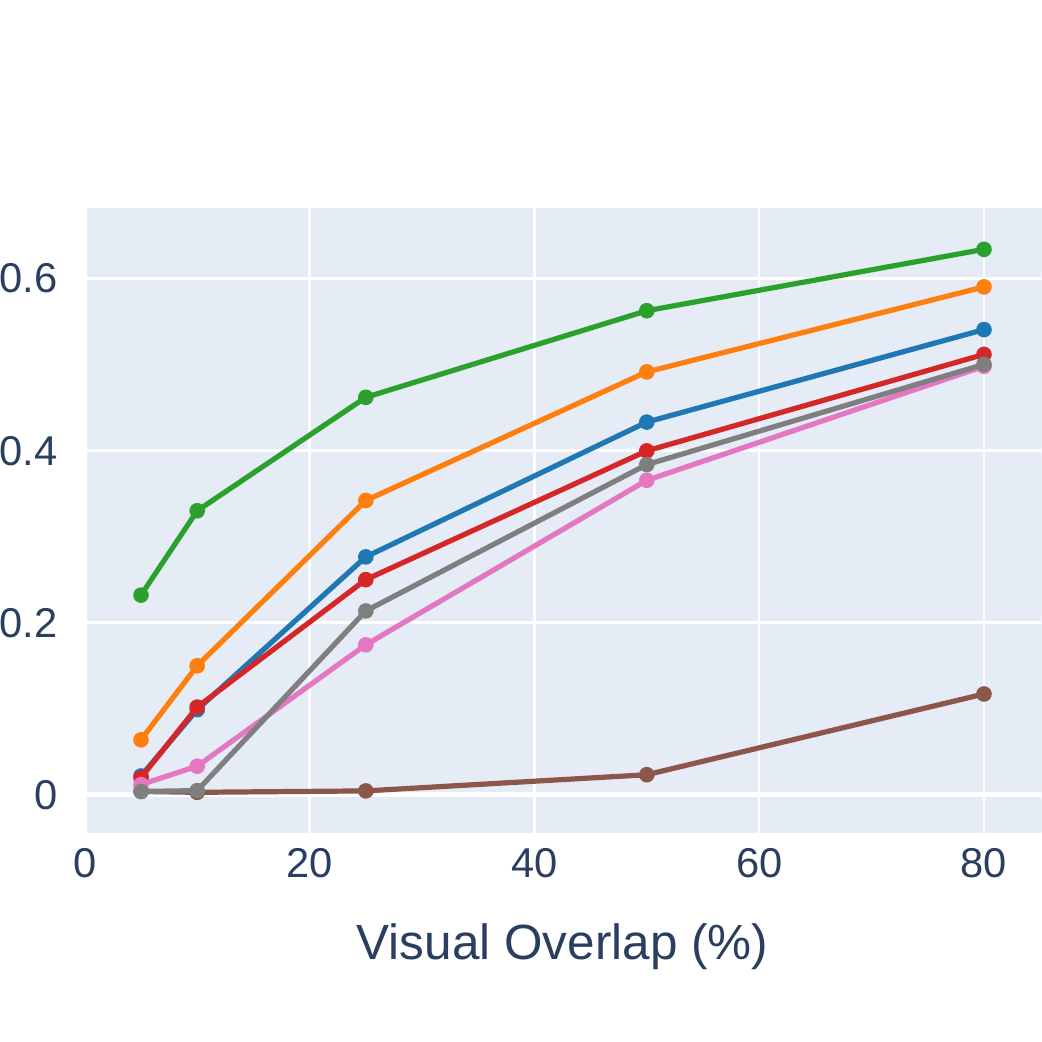}
    }
    \raisebox{1.5cm}{
    \resizebox{0.15\columnwidth}{!}{
      \ra{1.5}
      \begin{tabular}[b]{l}\toprule
      \textbf{Method}\\ \midrule
      \textcolor{color_Identity}{\rule{0.5cm}{0.6mm}} Identity\\
      \textcolor{color_R2D2}{\rule{0.5cm}{0.6mm}} R2D2~\cite{R2D2}\\
      \textcolor{color_SPSG}{\rule{0.5cm}{0.6mm}} SP+SG~\cite{SuperGlue}\\
      \textcolor{color_LoFTR}{\rule{0.5cm}{0.6mm}} LoFTR~\cite{LoFTR}\\
      \textcolor{color_DRCNet}{\rule{0.5cm}{0.6mm}} DRCNet~\cite{li2020dual}\\
      \textcolor{color_Mixed6e}{\rule{0.5cm}{0.6mm}} S2D~\cite{NeuralReprojection}\\
      \textcolor{color_S2DLoc}{\rule{0.5cm}{0.6mm}} \neurhal~\\
      \bottomrule
      \end{tabular}
      }
    }
    \caption{\textbf{Absolute camera pose experiment.}  We compare the
      performance of \neurhal against state-of-the-art local feature matching
      methods on ScanNet~\cite{ScanNet}.  The "identity" method consists in systematically predicting the identity pose.  We report the percentage of camera poses being correctly estimated for pairs of images that have an overlap between $2\%$ and $x\%$, as a function of $x$, for two rotation and translation error thresholds.
See discussion in Sec.~\ref{sec:application}. }
    \label{fig:overlaps}
    \vspace{-0.5cm}
  \end{center}
\end{figure}

In Fig.~\ref{fig:overlaps}, we compare the results of \neurhal against
state-of-the-art local feature matching methods.  In low-overlap settings,
  very few keypoints' correspondents can be identified and many keypoints'
  correspondents have to be outpainted. In this case, we find that \neurhal is
  able to estimate the camera pose correctly significantly more often than any
  other method, since \neurhal is the only method able to
  outpaint correspondences (see Fig.~\ref{fig:hallucination_sota}). For
  high-overlap image pairs, the ability to hallucinate is not useful since many
  keypoints' correspondents can be identified.  In this case, we find that
  state-of-the-art local feature matching methods to be slightly better than
  \neurhal.  This is likely due to the fact that \neurhal outputs low resolution
  correspondences maps while the other methods output high resolution
  correspondences.  The overall performance shows that \neurhal significantly
  outperforms all the competitors, which allows us to conclude that the ability
  of \neurhal to outpaint correspondences is beneficial for absolute pose
  estimation. Technical details concerning the previous experiment as well as
  additional experiments concerning the application to absolute camera pose
  estimation are provided in Sec.~\ref{sec:appendix_pose_estimation}).


\section{Limitations}\label{sec:limitations} We identified the following
limitations for our approach: $(i)$ - The previous experiments showed that
\neurhal is able to inpaint correspondences but the inpainted correspondence
maps are much less peaked compared to the outpainted correspondence maps. This
is likely due to the fact that inpainting correspondences is much more difficult
than outpainting correspondences~(see Sec~\ref{sec:analysis}). $(ii)$ - The
proposed architecture outputs low resolution correspondence maps~(see
Sec.~\ref{sec:architecture}), \eg $160\times120$ for input images of size
$640\times480$ and  an amount of padding $\gamma=50\%$. This is essentially due
to the quadratic complexity of attention layers we use (see
Sec.~\ref{sec:architecture_details} of the appendix). $(iii)$ - Our approach is
able to outpaint correspondences but our correspondence maps have a finite size.
Thus, in the case where a keypoint's correspondent falls outside the
correspondence map, the resulting correspondence map would be erroneous.  We
believe these three limitations are interesting future research directions.


\section{Conclusion} To the best of our knowledge, this paper is the first
attempt to learn to inpaint and outpaint correspondences. We proposed an
analysis of this novel learning task, which has guided us towards employing an
appropriate loss function and designing the architecture of our network.  We
experimentally demonstrated that our network is indeed able to inpaint and
outpaint correspondences on pairs of images captured in scenes that were not
seen at training-time, in both indoor~(ScanNet) and outdoor~(Megadepth)
settings.  We also tested our network on other datasets~(ETH3D and NYU) and
discovered that our model has strong generalization ability.  We then tried to
experimentally illustrate that hallucinating correspondences is not just a
fundamental AI problem but is also interesting from a practical point of view.
We applied our network to an absolute camera pose estimation problem and found
that hallucinating correspondences, especially outpainting correspondences,
allowed to significantly outperform the state-of-the-art feature matching
methods in terms of robustness of the resulting pose estimate. Beyond this
absolute pose estimation application, this work points to new research
directions such as integrating correspondence hallucination into
Structure-from-Motion pipelines to make them more robust when few images are
available.

\section{Ethics Statement}

The method described in this paper has the potential to greatly improve many
computer vision-based industrial applications, especially those involving visual
localization in GPS-denied or cluttered environments. For example robotics or
augmented reality applications could benefit from our algorithm to better
relocalize within their surroundings, which could lead to more reliable and
overall safer behaviours. If this was to be applied to autonomous driving or
drone-based search and rescue, one could appreciate the positive societal impact
of our method. On the other hand like many computer vision algorithms, it could
be applied to improve robustness of malicious devices such as weaponized UAVs,
or invade citizens privacy through environment re-identification. Thankfully as
AI technology advances, discussions and regulations are brought forward by
governments and public entities. 

These ethical debates pave the way for a
brighter future and can only make us think NeurHal will more bring benefits than
harms to society.

\section{Reproducibility}

We provide the \neurhal model architecture and weights in the supplementary
material. We also release a simple evaluation script that generates qualitative
results, and show in a notebook the results obtained on an image pair captured
indoors using a smartphone.

\section*{Acknowledgement}
The authors would like to thank Matthieu Vilain and Rémi Giraud for their insight
on visual correspondence hallucination.  This project has received funding from
the Bosch Research Foundation (\textit{Bosch Forschungsstiftung}). This work was
granted access to the HPC resources of IDRIS under the allocation
2021-AD011011682R1 made by GENCI.

\bibliography{string,ref}{}
\bibliographystyle{iclr2022_conference}

\clearpage

\appendix


\section*{Appendix}

In the following pages, we present additional experiments and technical details
about our visual correspondence hallucination method NeurHal. We present
additional experiments on the ability to hallucinate in
Sec.~\ref{sec:appendix_hallucination} and on camera pose estimation in
Sec.~\ref{sec:appendix_pose_estimation}. We describe technical details in
Sec.~\ref{sec:technical_details} and provide additional qualitative results in
Sec.~\ref{sec:qualitative_results}.

\vspace{-1.5cm}

\addcontentsline{toc}{section}{Appendix} 
\part{} 
\parttoc 

\section{Additional experiments concerning the ability to hallucinate correspondences}\label{sec:appendix_hallucination}

In this section we first present an additional ablation study on the ability to
hallucinate, followed by additional related work on correspondence
hallucination.

\subsection{Impact of learning to inpaint and outpaint}

To supplement the study made in Sec.~\ref{sec:evaluation}, we now aim at
evaluating the impact of learning to inpaint and outpaint specifically. To do
so, we isolate keypoints with the \textit{identified}, \textit{inpainted} and
\textit{outpainted} labels in our ScanNet~\cite{ScanNet} evaluation set. 

\begin{figure}
  \captionsetup[subfigure]{position=b}
  \centering

\subcaptionbox{Identified}{
    \includegraphics[height=3.5cm]{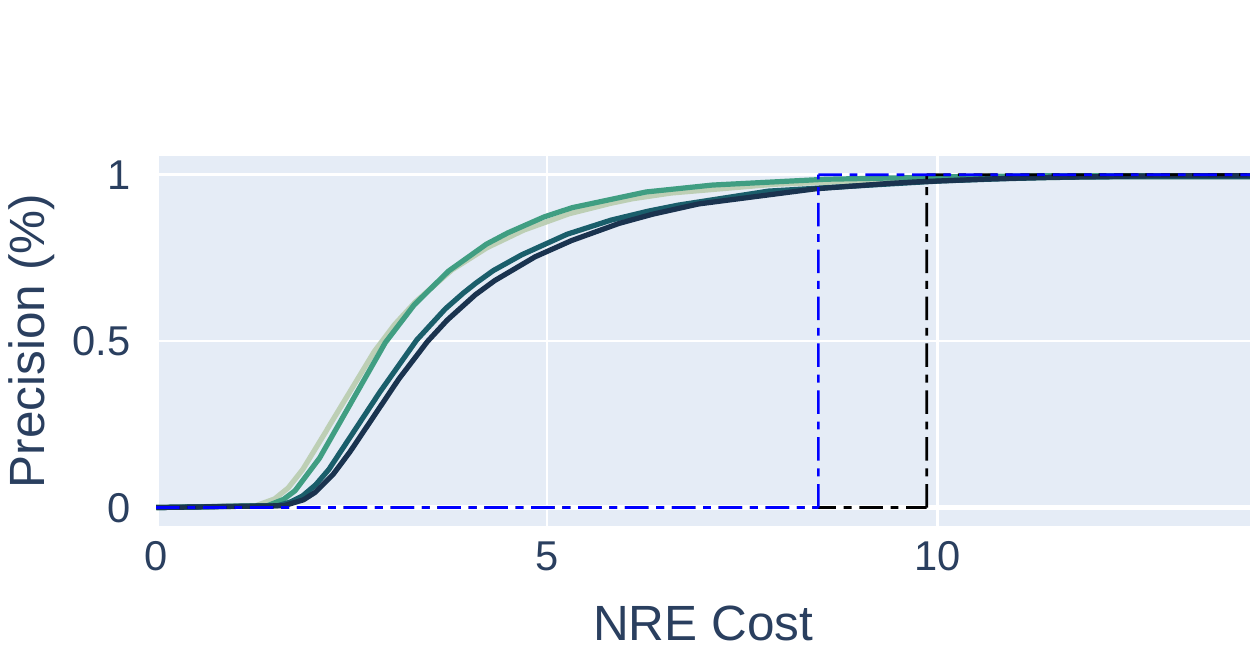}
    \includegraphics[height=3.5cm]{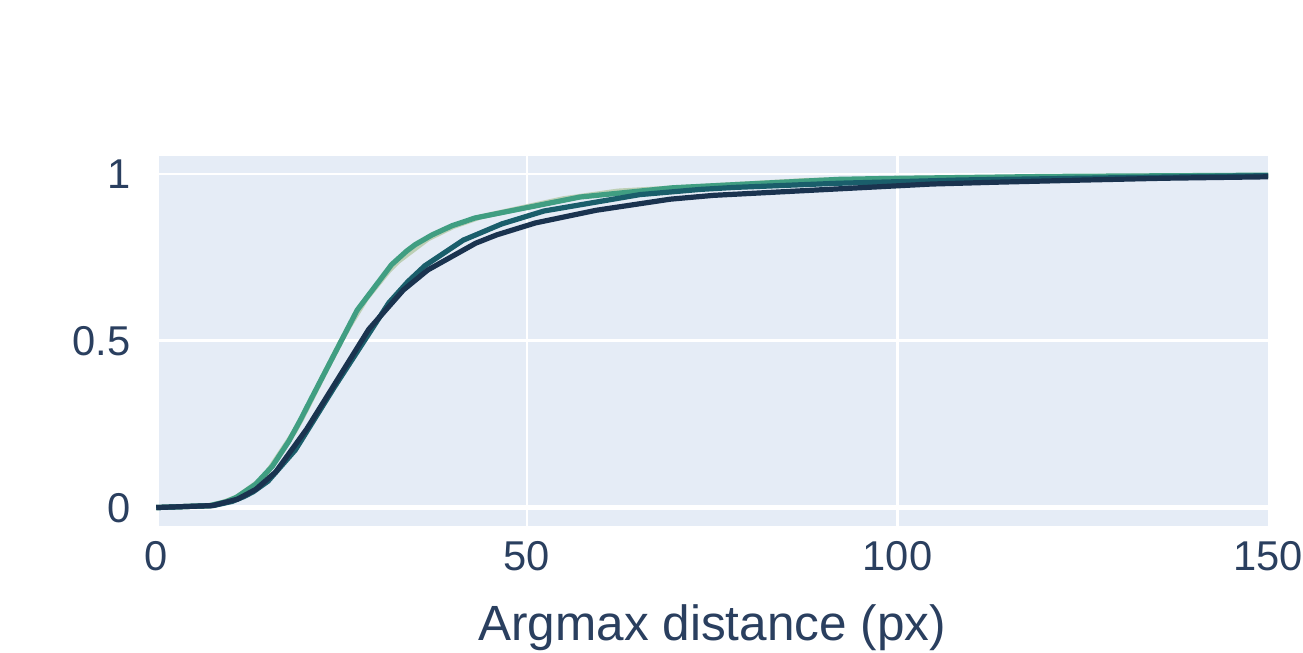}
}
\subcaptionbox{Inpainted}{
    \includegraphics[height=3.5cm]{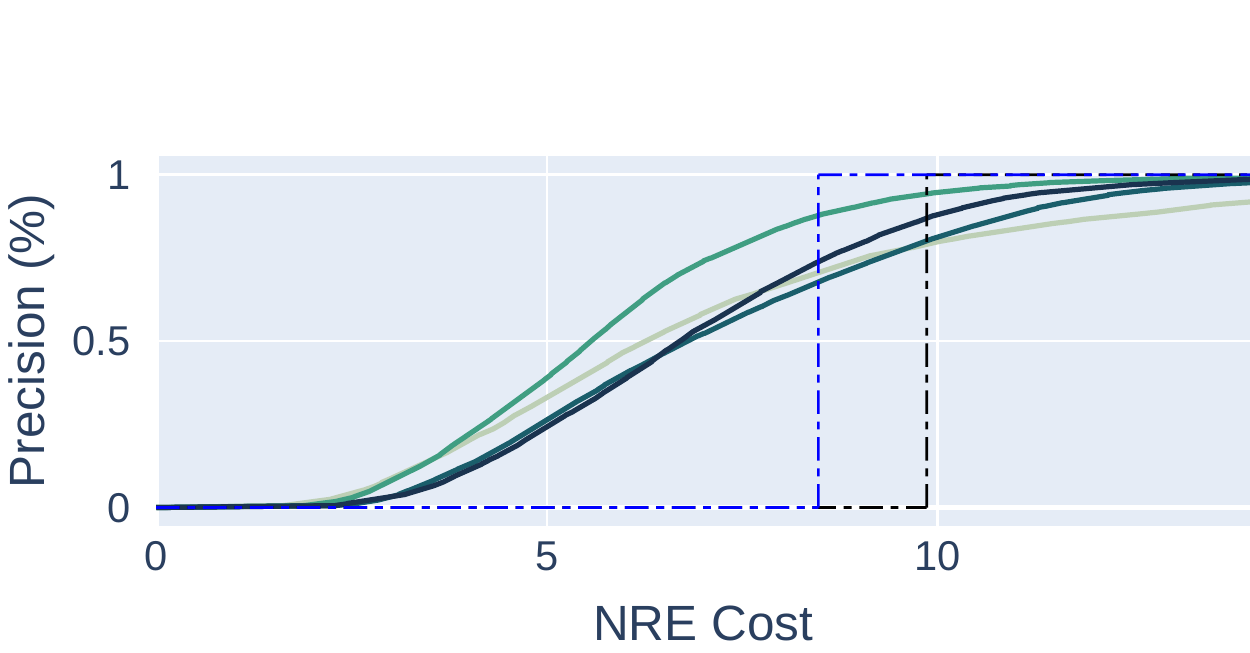}
    \includegraphics[height=3.5cm]{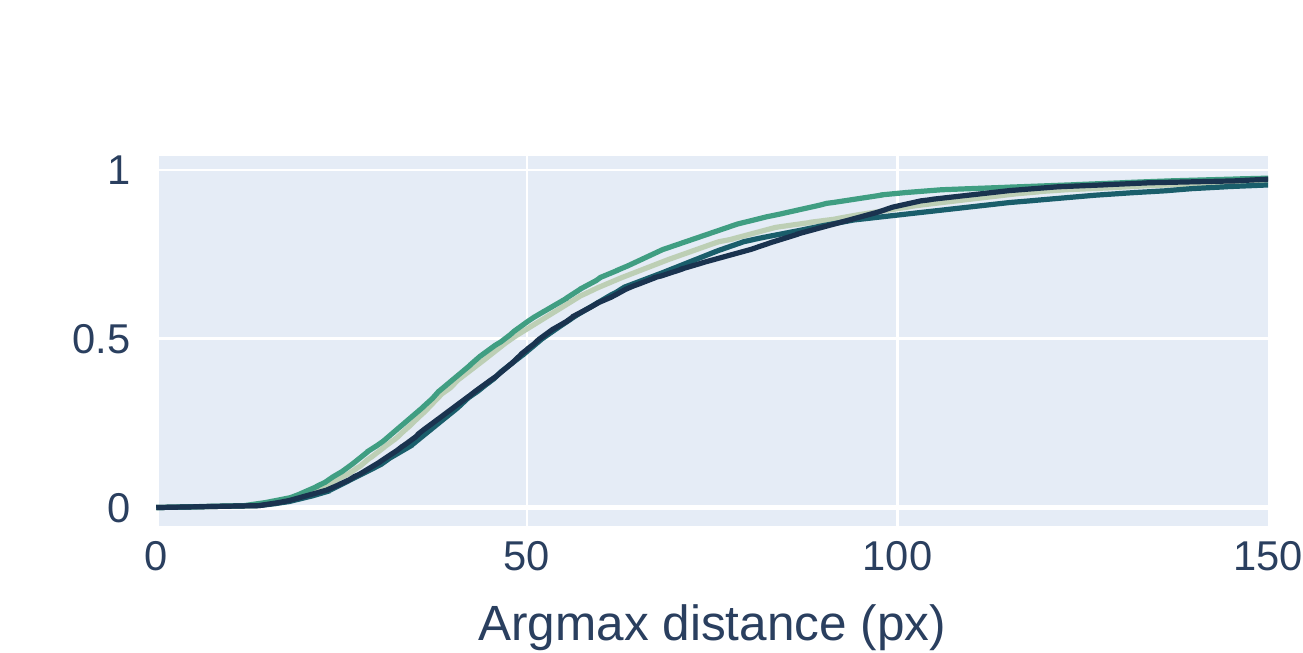}
}
\subcaptionbox{Outpainted}{
    \includegraphics[height=3.5cm]{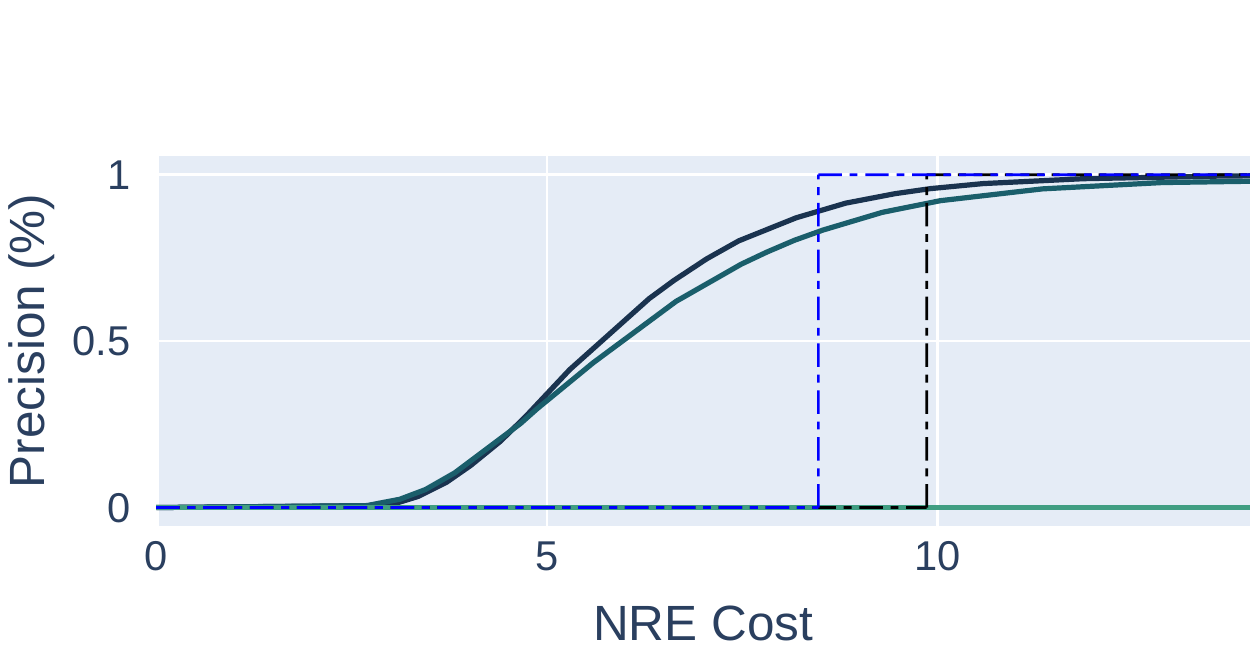}
    \includegraphics[height=3.5cm]{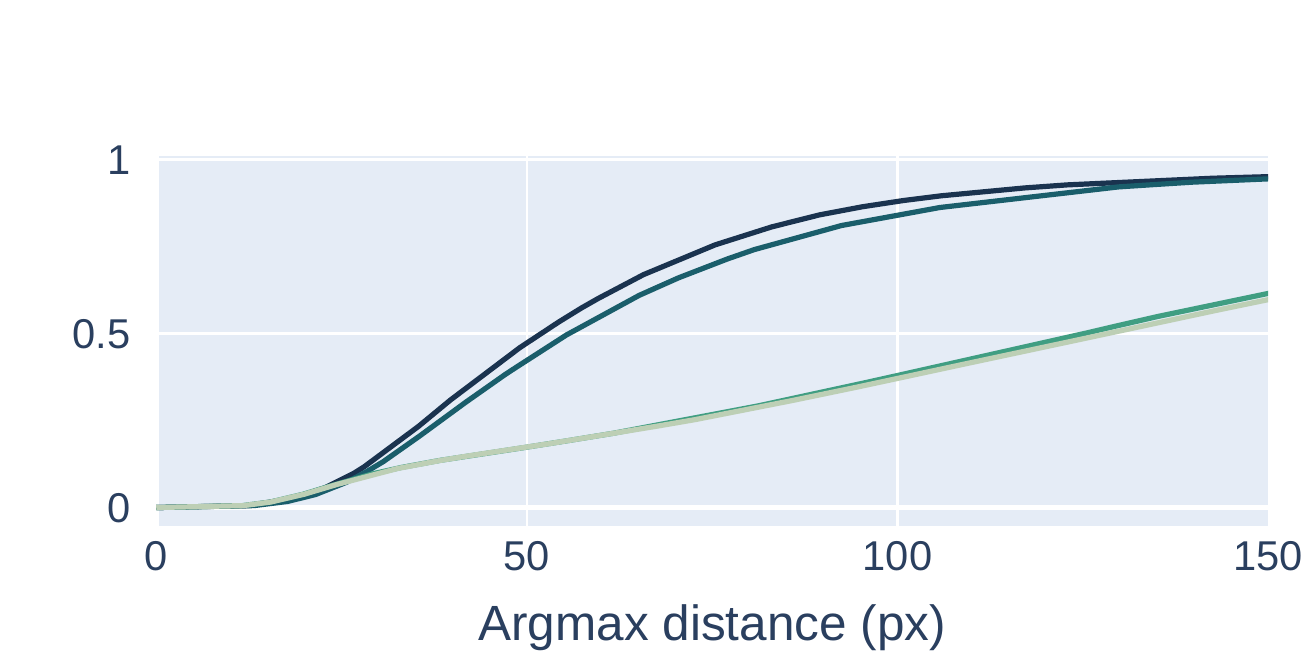}
}\\[2ex]
\begin{adjustbox}{max width=0.4\textwidth}
\ra{1.2}
    \begin{tabular}{cccc}%
    \toprule
    \multirow{2}{*}{Method} & \multicolumn{3}{c}{Training correspondences}\\
    \cmidrule(lr){2-4}
      & Identified & Inpainted & Outpainted\\
    \midrule
    \textcolor{color_VCH}{\rule{0.5cm}{0.6mm}} & \checkmark & &\\
    \textcolor{color_VCH_I}{\rule{0.5cm}{0.6mm}} & \checkmark & \checkmark & \\
    \textcolor{color_VCH_O}{\rule{0.5cm}{0.6mm}} & \checkmark & & \checkmark\\
    \textcolor{color_VCH_IO}{\rule{0.5cm}{0.6mm}} & \checkmark & \checkmark & \checkmark\\
    \bottomrule\\
    \multicolumn{4}{c}{
     \tikz\draw [blue,thick,dash dot] (0,0) -- (0.5,0);
     $ln(|\Omega|)_{\gamma=0\%}$
     ~~~~~~
     \tikz\draw [thick,dash dot] (0,0) -- (0.5,0);
     $ln(|\Omega|)_{\gamma=50\%}$
    }\\
    \end{tabular}
  \end{adjustbox}\\[2ex]

%
  \caption{ \textbf{Ability to hallucinate - Ablation study on ScanNet.} We
    compare the influence of adding inpainting and outpainting when training
    \neurhal. \textbf{(left column)} We report the percentage of keypoint's
    correspondents whose NRE cost is lower than $x$, as a function of $x$, for
    (a) identified (b) inpainted and (c) outpainted keypoints. \textbf{(right
    column)} We report the percentage of keypoint's correspondents whose
    distance \wrt the ground truth is lower than $x$ pixels, as a function of $x$, for the same categories.%
}\label{fig:ablation_hallucination} \vspace{0.4cm}
\end{figure}

In Fig.~\ref{fig:ablation_hallucination}, we show the results of an ablation
study on \neurhal's training setup. We report for the identification, inpainting
and outpainting tasks two sets of cumulative histograms: 1) the NRE costs at
ground truth keypoint correspondents' locations, and 2) distances between the
argmax of the correspondence map and the ground truth location. On NRE cost
cumulative histograms, we also report the results from the uniform distribution,
for models trained both with and without outpainting ($\gamma=0\%$ and
$\gamma=50\%$ respectively).


For the identification task (Fig.~\ref{fig:ablation_hallucination} (a)) we find
that all methods yield a consistent performance. The left figure reveals that
\neurhal predictions are significantly above the uniform distribution,
indicating peaky maps and thus confident predictions. The right figure shows
that the distance of the argmax location \wrt the ground truth is also robust
(\neurhal predicts at 1/8th of the original resolution but the histogram is
computed at full resolution).

For the inpainting task (Fig.~\ref{fig:ablation_hallucination} (b)) we can draw
similar conclusions. We find however that correspondence maps are overall less
peaky and closer to their respective uniform distribution, which indicates that
predictions are less confident. We also find that even though it was not trained
to inpaint, the identification baseline is surprisingly able to inpaint
correspondences as its performance is not far from the identification+inpainting
model.

Lastly for the outpainting task (Fig.~\ref{fig:ablation_hallucination}(c)), we
find that learning to outpaint gives a significant boost in performance on both
the NRE distribution and correspondents locations. We also find that jointly
learning to inpaint and outpaint is beneficial to the quality of the outpainted
cost maps, which implies that both objectives are complementary.

\subsection{Additional related work on correspondence hallucination}

In addition to the related work presented in Sec.~\ref{sec:related}, let us
mention some recent work that touches upon the problem of visual content
hallucination and relative pose estimation under very limited visual overlap.
The work of~\citet{Yang2019ExtremeRP} proposes to hallucinate the content of
RGB-D scans to perform relative pose estimation between two images. More
recently~\cite{Chen2021WideBaselineRC} regresses distributions over relative
camera poses for spherical images using joint processing of both images, and
manages to recover relative poses despite very limited visual overlap.  The work
of~\cite{Yang2020ExtremeRP, Qian2020Associative3DVR, Jin2021PlanarSR} shows that
employing a \emph{hallucinate-then-match} paradigm can be a reliable way of
recovering 3D geometry or relative pose from sparsely sampled images. In this
work, we focus on the problem of \emph{correspondence} hallucination which
unlike previously mentioned approaches does not aim at recovering explicit
visual content or directly regressing a relative camera pose.

\section{Additional experiments concerning the application to camera pose estimation}\label{sec:appendix_pose_estimation}

In this section, we present additional experiments on correspondence
hallucination for camera pose estimation. We begin with a study on the impact of
the pose estimator in Sec.~\ref{sec:impact_of_pose}, followed by a study on the
impact of the padding value $\gamma$ in Sec.~\ref{sec:impact_of_gamma}. Lastly,
we present in Sec.~\ref{sec:additional_pose_indoor} additional results on
indoor camera pose estimation.

\subsection{Influence of the pose estimator: \cite{NeuralReprojection} vs.
\cite{Chum2003LocallyOR}}\label{sec:impact_of_pose}

\cite{NeuralReprojection} provides a pose estimation framework which leverages
dense keypoint matching uncertainties to predict more accurate and robust camera
poses. Compared to the standard pose estimator presented
in~\cite{Chum2003LocallyOR} which relies on sparse 2D-to-3D correspondences, the
method from~\cite{NeuralReprojection} preserves rich information in the form of
dense loss maps that is particularly suited for ambiguous matches. For the
problem of correspondence hallucination we find the loss maps of both outpainted
and inpainted correspondences are usually unimodal but quite diffuse, and are
thus particularly suited for this pose estimator.

\begin{figure}[t]
\centering
\includegraphics[width=0.4\textwidth]{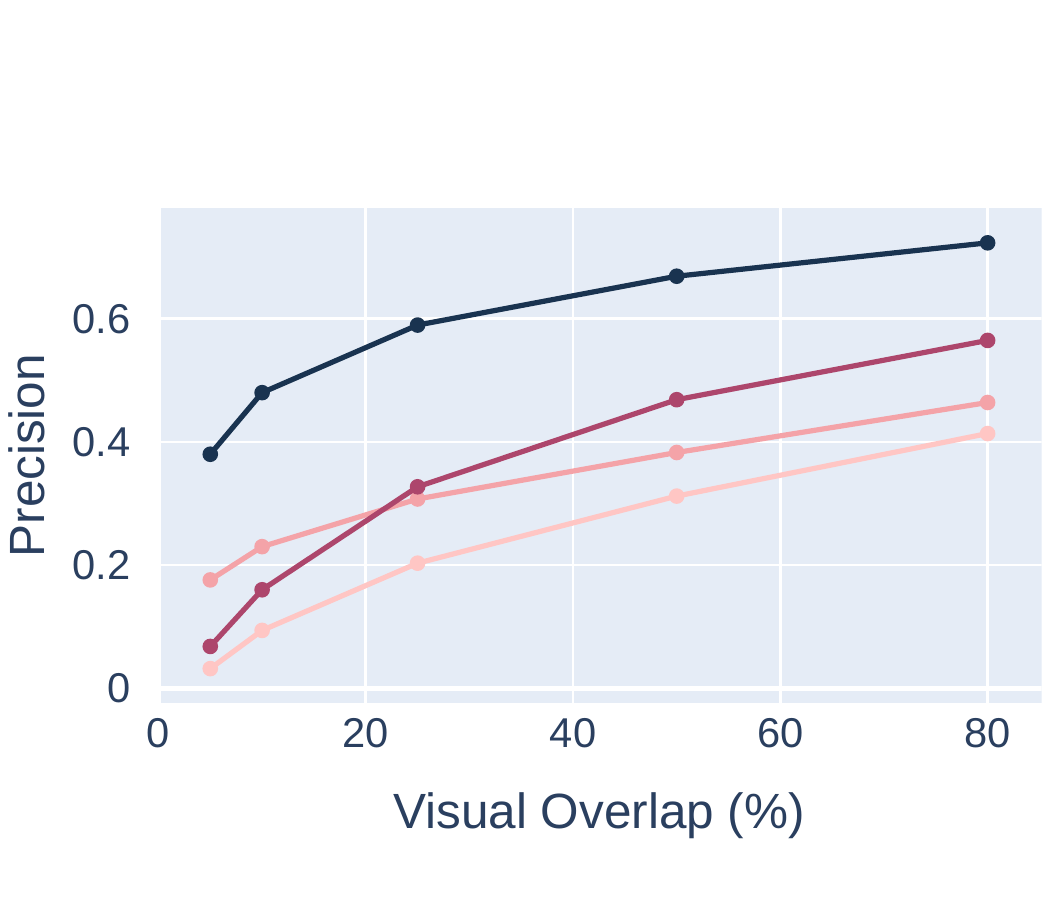}
\hspace{0.2cm}
\raisebox{2.35cm}{
  \begin{adjustbox}{max width=0.55\textwidth}
  \ra{1.2}
  \begin{tabular}{lcc}%
    \toprule
    Method & Hallucination & Pose estimator\\
    \midrule
    \textcolor{color_NRE_RE}{\rule{0.5cm}{0.6mm}} & & \cite{Chum2003LocallyOR} \\
   \textcolor{color_NeurHal_RE}{\rule{0.5cm}{0.6mm}} & \checkmark & \cite{Chum2003LocallyOR} \\
    \midrule
   \textcolor{color_NRE_NRE}{\rule{0.5cm}{0.6mm}} & & \cite{NeuralReprojection}\\
   \textcolor{color_NeurHal_NRE}{\rule{0.5cm}{0.6mm}} & \checkmark & \cite{NeuralReprojection}\\
    \bottomrule
  \end{tabular}%
  \end{adjustbox}
}
\caption{\textbf{Influence of the pose estimator: \cite{NeuralReprojection} vs. \cite{Chum2003LocallyOR}:} To study the influence of using the pose estimator proposed in~\cite{NeuralReprojection} compared to using the pose estimator from~\cite{Chum2003LocallyOR}, we report the performance of \neurhal
  with both estimators. We also include, for both estimators, the performance of \neurhal trained with
  identified correspondences only (\ie without hallucination). We report the percentage of camera poses being correctly estimated for pairs of ScanNet~\cite{ScanNet} images that have an overlap between $2\%$ and $x\%$ (as
  a function of $x$).} 
  \label{fig:nre_vs_re}
\end{figure}

To study the influence of the pose estimator, we report in
Fig.~\ref{fig:nre_vs_re} the performance of \neurhal~+~\cite{NeuralReprojection}
vs. \neurhal~+~\cite{Chum2003LocallyOR}. To estimate the camera pose using the
method presented in~\cite{Chum2003LocallyOR}, we simply take the argmax of each
correspondence map and treat it as a sparse 2D correspondent in the query image.
We also include the performance of \neurhal when trained without visual
correspondence hallucination (\ie trained using only identified ground truth
correspondences.)

We find that the two methods trained without hallucination have poor
performances for very low-overlap image pairs which underlines the importance of
correspondence hallucination in such cases. 

Concerning \neurhal trained with hallucination and using the pose
estimator~\cite{Chum2003LocallyOR}, taking the argmax of a very coarse
correspondence map
prevents the pose estimator from achieving good results.

\neurhal trained with hallucination and coupled with the pose estimator of
~\citet{NeuralReprojection} achieves the best results which shows that to obtain
robust absolute camera estimates it is important to \emph{combine} the ability
to hallucinate correspondences of \neurhal with the pose estimator
from~\cite{NeuralReprojection}.

%
%
%


\subsection{Impact of the value of $\gamma$}\label{sec:impact_of_gamma}

\begin{figure}[t]
\centering
  \includegraphics[width=0.35\textwidth]{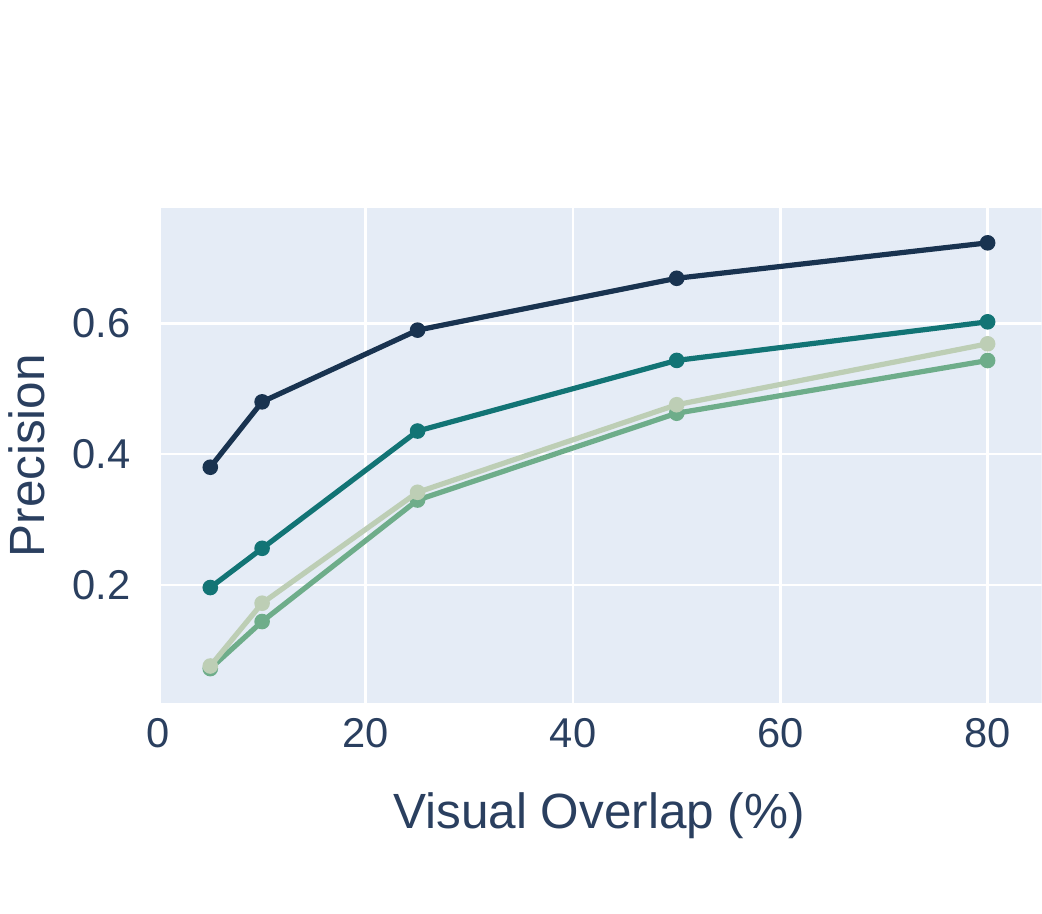}
\raisebox{2.1cm}{
  \begin{adjustbox}{max width=0.2\textwidth}
  \ra{1.2}
  \begin{tabular}{lc}%
    \toprule
    Method & $\gamma$ \\
    \midrule
    \textcolor{color_gamma_50}{\rule{0.5cm}{0.6mm}} & $\gamma=50$\% \\
    \textcolor{color_gamma_25}{\rule{0.5cm}{0.6mm}} & $\gamma=25$\% \\
    \textcolor{color_gamma_10}{\rule{0.5cm}{0.6mm}} & $\gamma=10$\% \\
    \textcolor{color_gamma_00}{\rule{0.5cm}{0.6mm}} & $\gamma=0$\% \\
    \bottomrule
  \end{tabular}%
  \end{adjustbox}
}

\caption{\textbf{Impact of the value of $\gamma$:} For increasing values of $\gamma$, we
  report the percentage of camera poses being correctly estimated for pairs of
  ScanNet images that have an overlap between $2\%$ and $x\%$ (as
  a function of $x$), for $\tau_t=1.5m$ and $\tau_r=20.0\degree$.  We find that a
  small value of $\gamma=10\%$ yields no benefit and even damages performance,
  while values of $\gamma=25\%$ and $\gamma=50\%$ bring significant
improvements, especially at small visual overlaps.}
  \label{fig:impact_of_outpainting}

\end{figure}

We report in Fig.~\ref{fig:impact_of_outpainting} the absolute camera pose estimation
performance for varying values of $\gamma$. We compute the percentage of camera
poses being correctly estimated for ScanNet~\cite{ScanNet} test images pairs
that have an overlap between $2\%$ and $x\%$ (as a function of $x$) for a
translation threshold of $1.5m$ and a rotation threshold of $20.0\degree$.

We find that using only a small percentage of outpainting such as
$\gamma=10\%$ does not improve the performance which is most likely
due to the small amount of added training keypoints. For higher $\gamma$ values
however significant gains are visible, especially at small visual overlaps. This
experiment demonstrates the benefit of learning to outpaint correspondences
beyond image borders, and broaden the extent of usable source keypoints to
perform camera pose estimation.

\begin{figure}
  \centering
  \subcaptionbox{Field-of-view \wrt $\gamma$}{
    \includegraphics[width=0.4\textwidth]{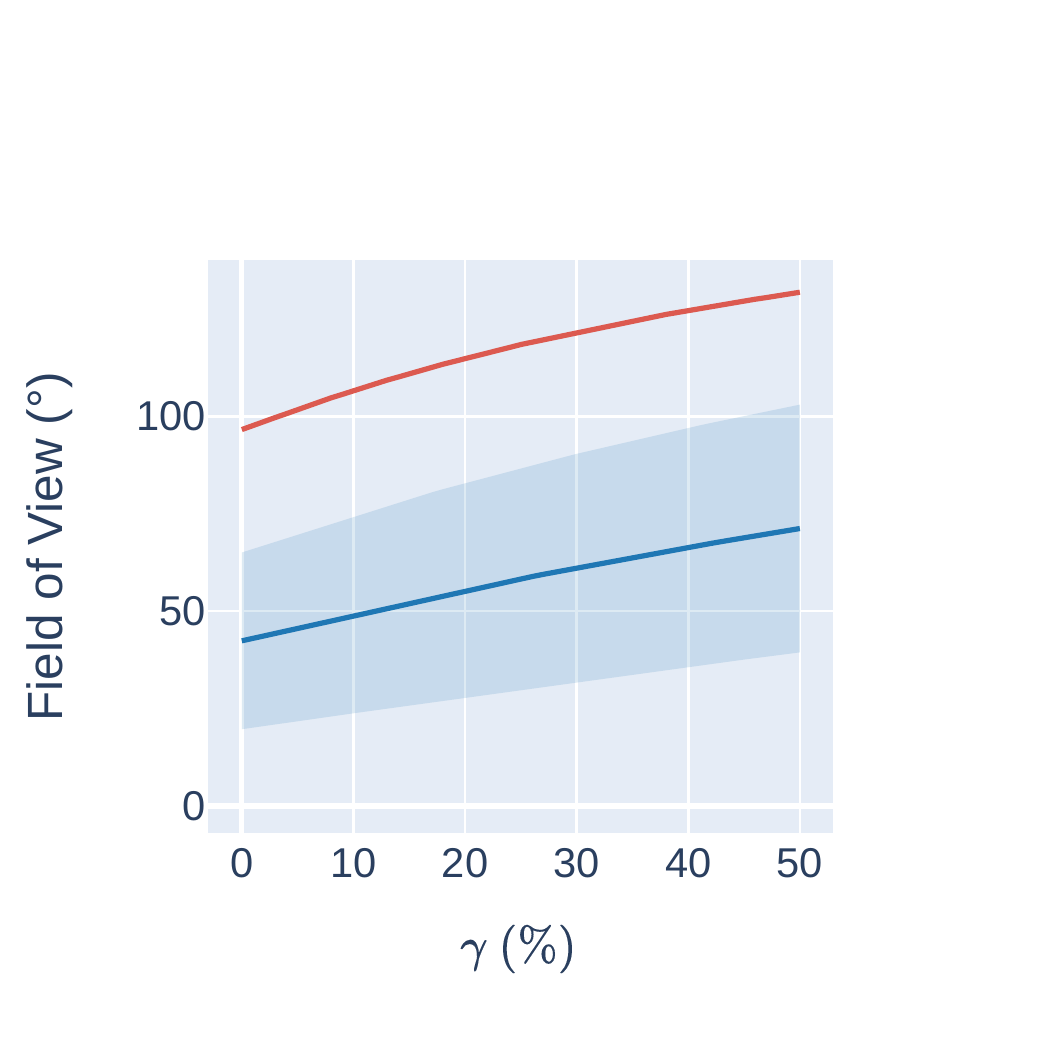}
  }
  \subcaptionbox{Relative viewpoint statistics}{
    \raisebox{2.3cm}{
      \begin{adjustbox}{max width=0.5\textwidth}
      \ra{1.2}
      \begin{tabular}{lcccccccc}%
        \toprule
        Dataset & $|\Delta r_x|$ & $|\Delta r_y|$ & $|\Delta r_z|$ & $|\Delta
        \theta|$ & $|\Delta f|$\\
        \midrule
        \textcolor{color_scannet}{\rule{0.3cm}{0.6mm}} ScanNet & $29.21$\degree
                                                               & $38.72$\degree
                                                               & $25.68$\degree
                                                               & $55.20$\degree
                                                               & $0.00$mm\\
        \textcolor{color_megadepth}{\rule{0.3cm}{0.6mm}} Megadepth &
        $4.73$\degree & $6.91$\degree & $1.64$\degree & $20.25$\degree & $376.69$mm\\
        \bottomrule
      \end{tabular}
      \end{adjustbox}
    }
  }
  \vspace{0.1cm}
  \subcaptionbox{Histogram of absolute relative angle norm $|\Delta\theta|$}{
    \includegraphics[width=0.47\textwidth]{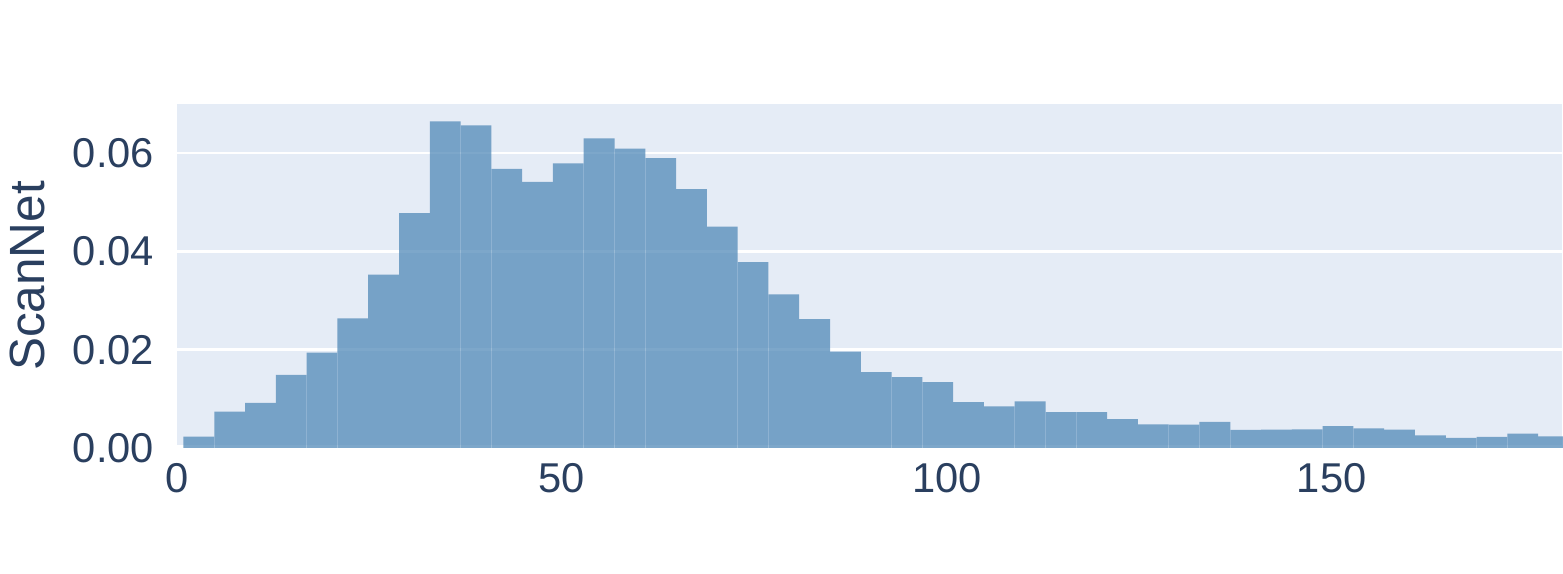}
    \includegraphics[width=0.47\textwidth]{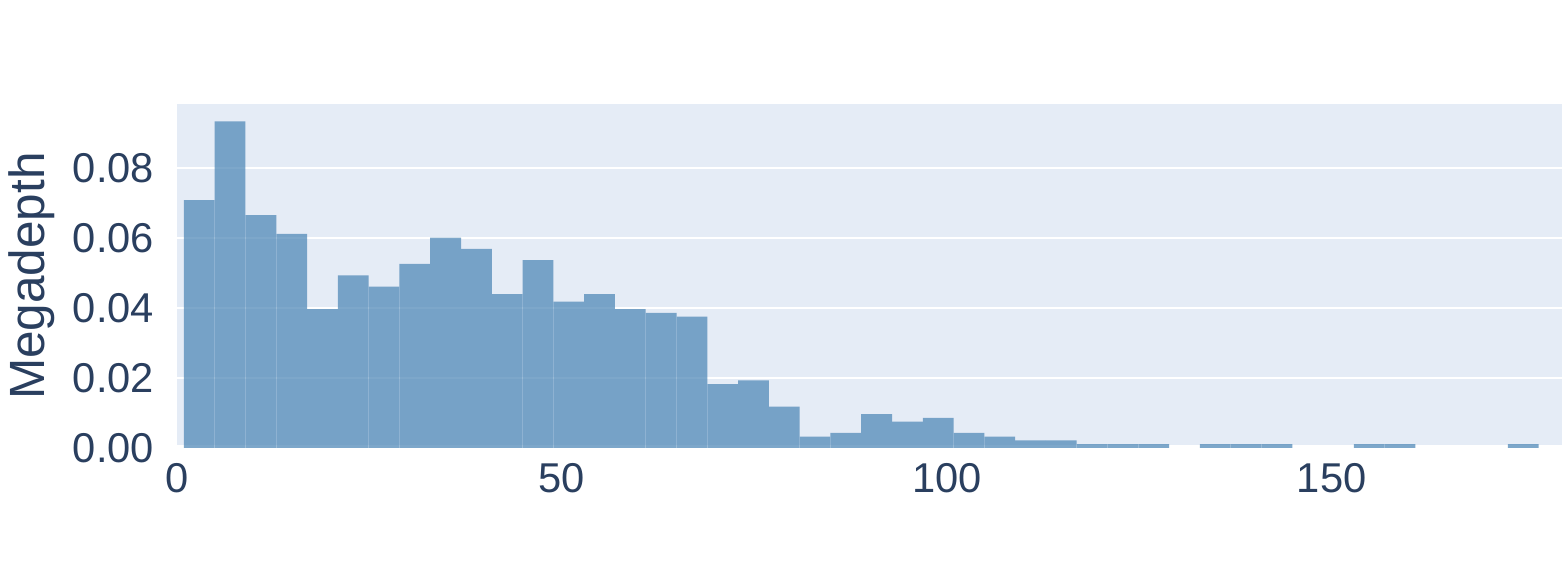}
  }
\caption{\textbf{Field-of-view as a function of $\gamma$ and relative viewpoint
  statistics:} We report in $(a)$ the average camera field-of-view as a function of $\gamma$
  on ScanNet~\cite{ScanNet} and Megadepth~\cite{Megadepth} images.  We find that
  $\gamma=50\%$ enables a significant amount of additional visual content to
  reproject within the image boundaries. We report in $(b)$ the median absolute
  difference in rotation along the $x$, $y$ and $z$ axis, norm of the relative
  rotation, along with the difference in focal length on low-overlap image pairs
  for ScanNet~\cite{ScanNet} and Megadepth~\cite{Megadepth}. We report in (c)
  the histogram of absolute relative angle norm on both datasets. We find ScanNet
  image pairs exhibit strong relative angular motion while Megadepth image pairs
display predominantly zoom-ins and zoom-outs.}
  \label{fig:fov}
  \label{tab:stats}
\end{figure}

We report in Fig.~\ref{fig:fov} the camera field-of-view as a function
of the padding parameter. We find that $\gamma=50\%$ provides $130\degree$ and
$71\degree$ of field-of-view on average on ScanNet and Megadepth respectively,
which is significantly wider than $\gamma=0\%$.

\subsection{Additional indoor pose estimation
results}\label{sec:additional_pose_indoor}

\begin{figure}
\captionsetup[subfigure]{position=b}
\centering

\footnotesize
\stackon[-8pt]{\includegraphics[width=0.135\textwidth]{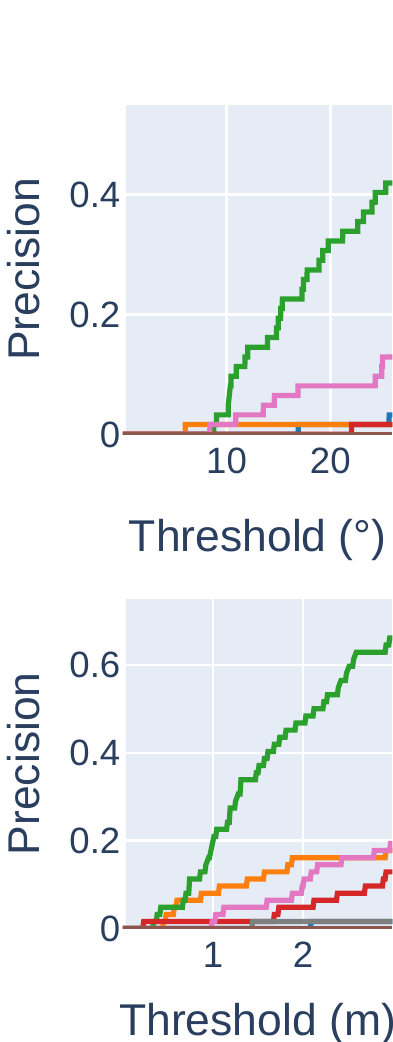}}{~~\tiny{Identity}}
\stackon[-8pt]{\includegraphics[width=0.135\textwidth]{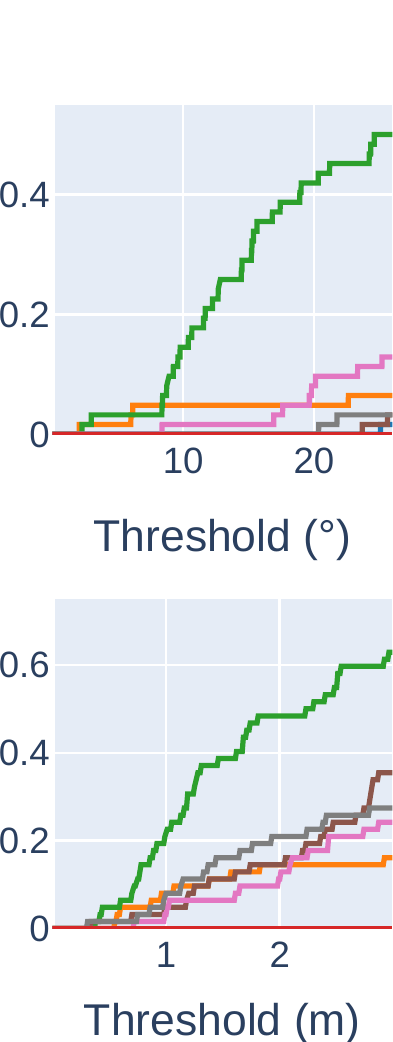}}{\tiny{R2D2}}
\stackon[-8pt]{\includegraphics[width=0.135\textwidth]{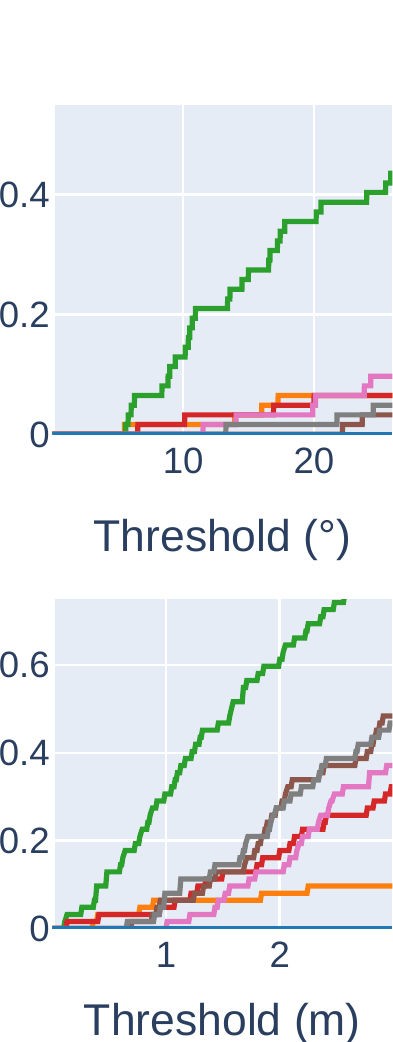}}{\tiny{SP+SG}}
\stackon[-8pt]{\includegraphics[width=0.135\textwidth]{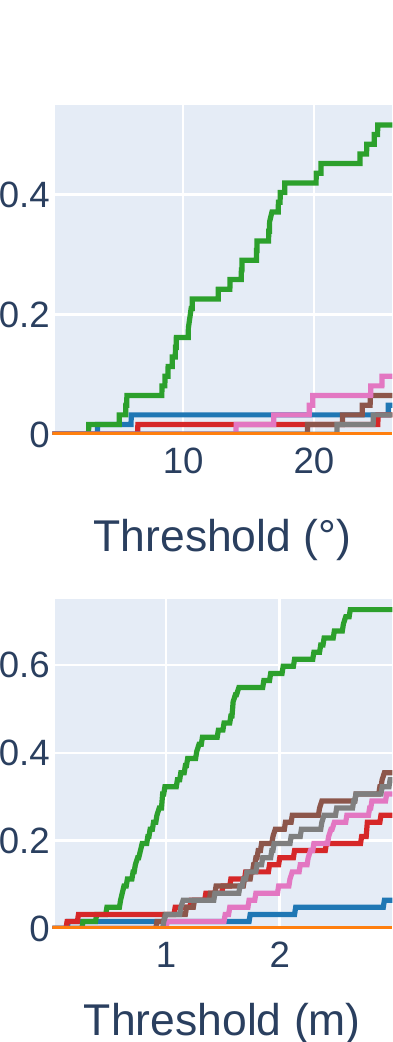}}{\tiny{LoFTR}}
\stackon[-8pt]{\includegraphics[width=0.135\textwidth]{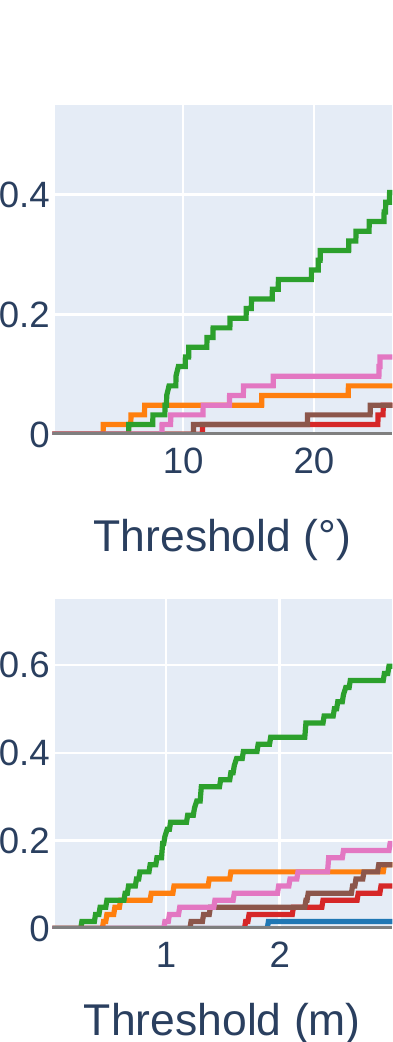}}{\tiny{DRCNet}}
\stackon[-8pt]{\includegraphics[width=0.135\textwidth]{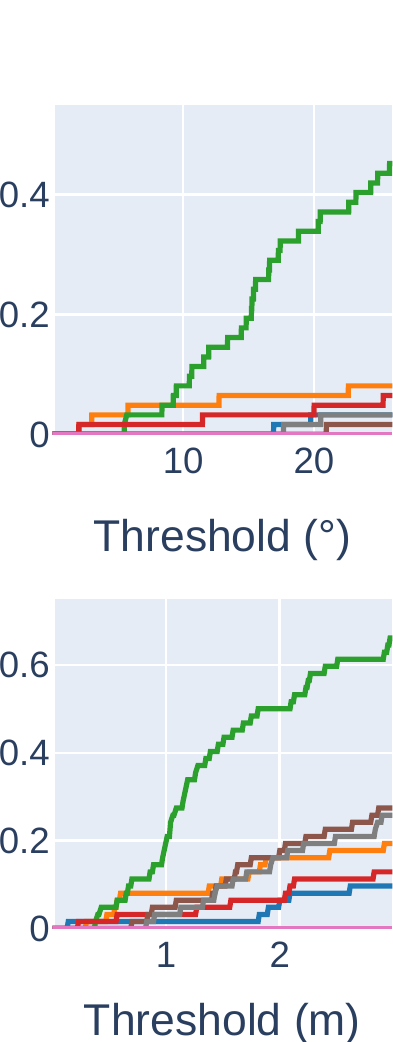}}{\tiny{S2D}}
\stackon[-8pt]{\includegraphics[width=0.135\textwidth]{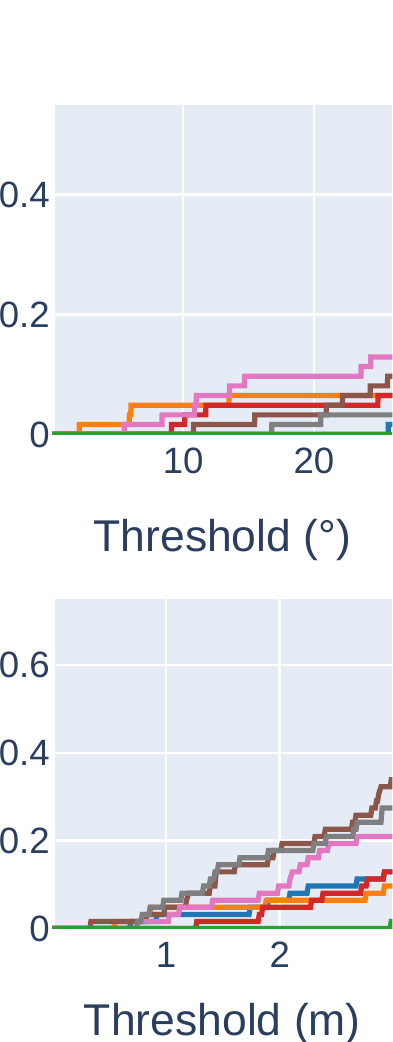}}{\tiny{NeurHal}}

\vspace{0.1cm}

\small{
\textcolor{color_Identity}{\rule{0.3cm}{0.6mm}} Identity 
\hfill
\textcolor{color_R2D2}{\rule{0.3cm}{0.6mm}} R2D2~\cite{R2D2} 
\hfill
\textcolor{color_SPSG}{\rule{0.3cm}{0.6mm}} SP+SG~\cite{SuperGlue}
\hfill
\textcolor{color_LoFTR}{\rule{0.3cm}{0.6mm}} LoFTR~\cite{LoFTR}

\textcolor{color_DRCNet}{\rule{0.3cm}{0.6mm}} DRCNet~\cite{li2020dual}
\hfill
\textcolor{color_Mixed6e}{\rule{0.3cm}{0.6mm}} S2D~\cite{NeuralReprojection}
\hfill
\textcolor{color_S2DLoc}{\rule{0.3cm}{0.6mm}} \neurhal
\hfill
}

\caption{\textbf{Camera pose
  estimation experiment - Worst cases:} We report the performance of \neurhal and state-of-the-art
  feature matching methods on ScanNet~\cite{ScanNet} image pairs with visual
  overlaps between 2\% and 5\%. For every column, we subselect the 25\% of
  images pairs with the worst predictions for a given method. We find that in
  all cases, \neurhal strongly outperforms its competitors. On the contrary, on
  the worst \neurhal predictions state-of-the-art methods achieve a much lower
performance, which is either on par or lower than the predictions obtained using
the Identity.}
\label{fig:worst_cases_all}

\end{figure}

In addition to the results presented in Fig.~\ref{fig:overlaps}, we report in
Fig.~\ref{fig:worst_cases_all} the performance of \neurhal and state-of-the-art
feature matching methods on ScanNet~\cite{ScanNet} image pairs with visual
overlaps between 2\% and 5\%.  For every method, we subselect the 25\% of images
pairs with the worst predictions, and compare it with the performance of its
competitors.  We find that in all cases, \neurhal strongly outperforms its
competitors. On the worst \neurhal predictions, state-of-the-art methods achieve
a much lower performance. For this category we can observe that all \neurhal
competitors are either on par or achieve a lower performance than the Identity
predictions.

This figure highlights the fact that when \neurhal fails to correctly estimate
the camera pose, all the competitors also fail since all the methods perform
similarly to the "identity" method, \ie the method that consists in
systematically predicting the identity pose.

\begin{figure}
\captionsetup[subfigure]{position=b}
\centering
\includegraphics[width=0.55\textwidth]{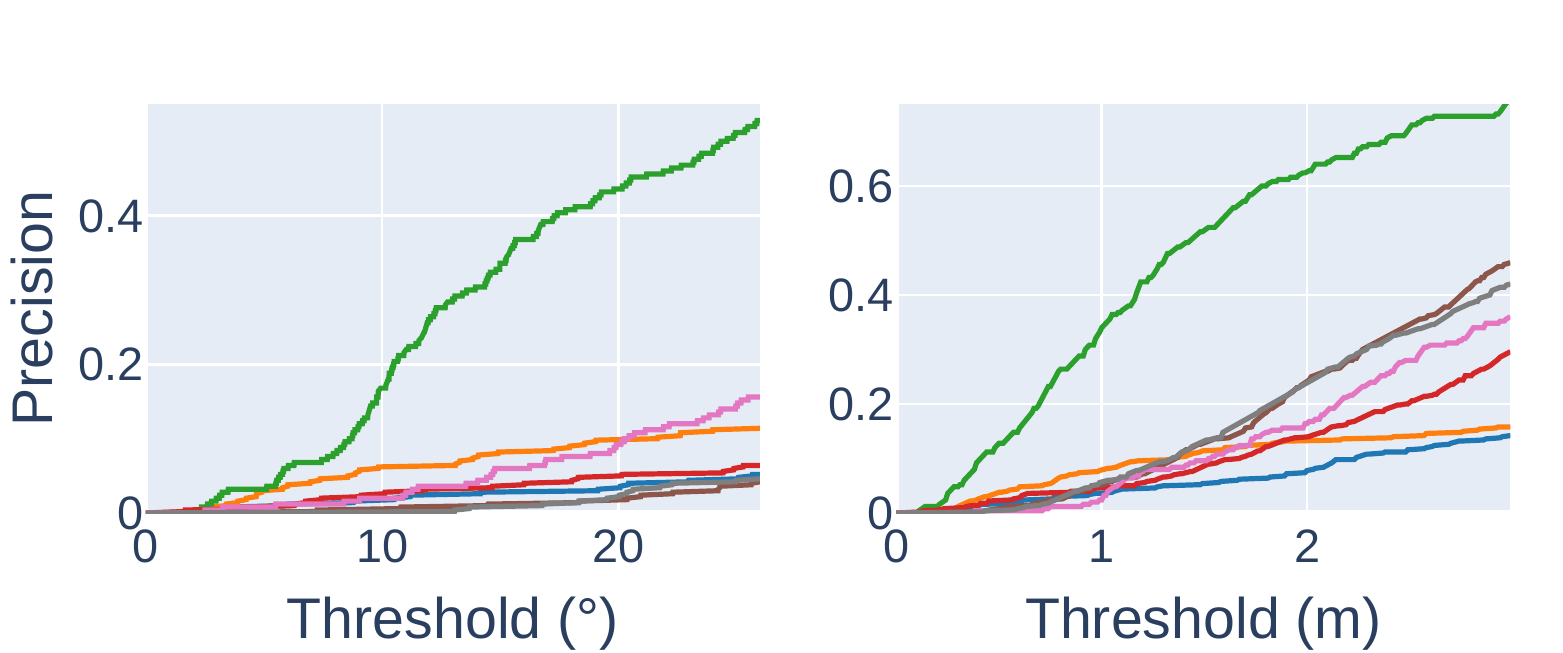}
\vspace{0.1cm}

\small{
\textcolor{color_Identity}{\rule{0.3cm}{0.6mm}} Identity 
\hfill
\textcolor{color_R2D2}{\rule{0.3cm}{0.6mm}} R2D2~\cite{R2D2} 
\hfill
\textcolor{color_SPSG}{\rule{0.3cm}{0.6mm}} SP+SG~\cite{SuperGlue}
\hfill
\textcolor{color_LoFTR}{\rule{0.3cm}{0.6mm}} LoFTR~\cite{LoFTR}

\textcolor{color_DRCNet}{\rule{0.3cm}{0.6mm}} DRCNet~\cite{li2020dual}
\hfill
\textcolor{color_Mixed6e}{\rule{0.3cm}{0.6mm}} S2D~\cite{NeuralReprojection}
\hfill
\textcolor{color_S2DLoc}{\rule{0.3cm}{0.6mm}} \neurhal
\hfill
}

\caption{\textbf{Camera pose estimation experiment - cumulative precision:} We report the performance of \neurhal and state-of-the-art
  feature matching methods on ScanNet~\cite{ScanNet} image pairs with visual
  overlaps between 2\% and 5\%. For various angular and translation thresholds
  we report the percentage of correctly localized images. We find that in
  all cases, \neurhal strongly outperforms its competitors.}
\label{fig:cumulative_localization}
\end{figure}

Fig.~\ref{fig:cumulative_localization} shows that \neurhal is much more robust than
state-of-the-art local feature matching methods for pairs of images with a low
overlap.

\section{Technical details}\label{sec:technical_details}

\subsection{Architecture details}
\label{sec:architecture_details}

\neurhal's architecture can be separated in two building blocks: the convolutional
backbone and the multi-head attention block.

\paragraph{Convolutional backbone.} The convolutional backbone consists of a
truncated Inceptionv3~\cite{Inceptionv3} model (up to Mixed-6a, 768-dimensional
descriptors), modified as per~\citet{NeuralReprojection} to provide, 
in the case of ScanNet~\cite{ScanNet}, a $1/8$ output-to-input
resolution ratio. To help with memory consumption we apply a simple 2D
convolutional layer to compress the descriptor size to 384. In the case where
$\gamma>0$, we subsequently pad $\H_\T$ with the learned vector $\boldsymbol{\lambda}$,
producing $\H_{\T,\text{pad}}$.

\paragraph{Positional encoding.} After computing $\H_\S$ and
$\H_{\T,\text{pad}}$ with the convolutional backbone, positional encoding is
applied to both dense feature maps. Similarly to SuperGlue~\cite{SuperGlue}, we
use a 6-layer MLP of size (32, 64, 128, 256, 384), mapping a positional meshgrid
between $(-1, 1)$ (centered around the image center) to higher dimensionalities.
BatchNorm and ReLU layers are placed between every module. In our experiments,
we tried adding more positional encoding layers but found it did not make a
difference in performance. After applying the positional encoding, sparse
descriptors $\curl{\d_{\S,n}}_{n=1...N}$ are bilinearly interpolated at
$\curl{\p_{\S,n}}_{n=1...N}$ in $\H_\S$.

\paragraph{Self-attention.} Following the positional encoding, a single
multi-head attention layer is applied on $\H_{\T,\text{pad}}$, with 4 heads. It
consists of a standard dot-product attention~\cite{Vaswani2017AttentionIA},
coupled with a gating mechanism. For a given query $Q$, key $K$ and value $V$,
we compute the attention as $\text{Attention}(Q, K, V) =
\text{softmax}(g*QK^T)V$ where $g=\sigma(max(QK))$.  To mitigate the quadratic
cost of the dot-product attention, we also apply a max-pooling operator on keys
and values with a stride of 2, as we empirically found it had very little impact
on performance.
We also tried using a Linear Transformer (\textit{e.g.}
LinFormer~\cite{Katharopoulos2020TransformersAR}) architecture, but despite
trying numerous variants we found it consistently damaged the convergence of the
model.

\paragraph{Cross-attention.} Using the same attention-layer design, we
subsequently apply it once between $\curl{\d_{\S,n}}_{n=1...N}$ and $\H_\S$.
This layer allows for communication between the interpolated source descriptors
which will be used to produce the final correspondence maps, and the original dense source image
content. Then, we apply $k$ cross-attention layers between
$\curl{\d_{\S,n}}_{n=1...N}$ and $\H_{\T,\text{pad}}$. We empirically found
these layers to be most important, as they allow for direct communication
between the sparse source descriptors and the dense target feature maps, prior
to the correspondence maps computation. After trying different values for $k$
and with memory consumption in mind, we settled for $k=4$ in all our
experiments.


\paragraph{Implementation.} The model is implemented in
PyTorch~\cite{Paszke2017automatic}. For an indoor sample with $2000$ keypoints
it has an average throughput of $8.84$ image/s on an NVIDIA RTX 3070 GPU. We
report the number of parameters in our model in Table~\ref{table:parameters}.

\subsection{Datasets and Training details}\label{sec:dataset_details}

\paragraph{ScanNet.} The ScanNet~\cite{ScanNet} dataset is a large-scale indoor
dataset containing monocular RGB videos and dense depth images, along with
ground truth absolute camera poses. As SuperGlue~\cite{SuperGlue} and
LoFTR~\cite{LoFTR}, we pre-compute the visual overlaps between all image pairs for
both training and test scenes. For the training set we sample images with a
visual overlap between 2\% and 50\% from the ScanNet training scenes, which
provides us with challenging images to handle. We assemble $6M$ image pairs and
randomly subsample $200k$ pairs at every training epoch. For testing images, we
sample $2,500$ image pairs with overlaps between 2\% and 80\% from the
ScanNet testing scenes, using several bins to ensure the sampling is close to
being uniform. For both training and testing images, we sample keypoints in the
source image along a regular grid with cell sizes of $16$ pixels. We remove
keypoints with invalid depth, as well as those where the local depth gradient is
too high, as the depth information might not be reliable. We mark keypoints
falling outside the target image plane as being outpainted, and we automatically
detect the keypoints to inpaint through a cyclic projection of the source keypoints
to the target image and back. The remaining keypoints are labeled as
identifiable.  For all ScanNet experiments, \neurhal uses a $1/8$
output-to-input resolution ratio, with a target correspondence map maximum edge
size of $80$ pixels (when $\gamma=0\%$).

\paragraph{Megadepth.}\label{sec:megadepth}

We use Megadepth~\cite{Megadepth} to train and evaluate \neurhal on outdoor
images. This dataset contains over one million images captured in touristic
places, and split in $196$ scenes.
To train \neurhal and following~\citet{NeuralReprojection} guidelines, we
use the provided SIFT~\cite{SIFT}-based 3D reconstruction which was made with
COLMAP~\cite{Schoenberger2016sfm}. Because the sparse 3D point cloud comes from
SfM, we find however that very little keypoints can be marked as inpainted.
Indeed, no 3D reconstruction is applied to objects or people occluding the
scene.  To allow for a wide variety of image pairs we use the sparse
reconstruction to estimate the visual overlap and sample pairs with an overlap
between 20\% and 100\%. We however find this overlap estimation to be quite
unreliable, as only part of the scene is usually reconstructed. Since
Megadepth~\cite{Megadepth} images are of much higher resolution than
ScanNet~\cite{ScanNet}, we configure \neurhal to use a $1/16$ output-to-input
resolution (with a simple max-pooling layer in the CNN). We set the target
correspondence map maximum edge size of $60$ pixels (when $\gamma=0\%$), to
allow for space in memory when $\gamma=50\%$.

\paragraph{Overlap estimation.} For a given pair of images, we approximate the
visual overlap by computing the covisibility ratio of keypoints for every image
pair. For a given source and target image pair, we first compute the
source-to-target and target-to-source covisibility ratios using ground truth
depth data and camera poses. We then define
the visual overlap as the minimum between both ratios. On Megadepth we find
this overlap estimation to be fairly noisy, as depth is only partially known. 



\begin{table}[t]
\centering
\begin{adjustbox}{max width=0.55\textwidth}
\ra{1.0}
\begin{tabular}{lc}%
\toprule
Layer & \# of parameters  \\
\midrule
CNN & 2.4 M \\
Positional Encoding & 142 K \\
Self-Attention & 1.9 M\\
Cross-Attention & 7.2 M\\
\midrule
Total & 11.7 M\\
\bottomrule
\end{tabular}%
\end{adjustbox}
\vspace{0.1cm}
\caption{\textbf{Number of parameters in NeurHal}}
\label{table:parameters}
\end{table}

\paragraph{Optimizers and scheduling.} On both datasets \neurhal is trained for
a maximum of 40 epochs. We use an initial learning rate of $10^{−3}$, with a
linear learning rate warm-up in 3 epochs from 0.1 of the initial learning rate.
As~\citet{LoFTR}, we decay the learning rate by 0.5 every 8 epochs starting from
the 8th epoch. We apply the linear scaling rule and use a batch size of 8 over 8
NVIDIA V100 GPUs.  We use the AdamW~\cite{Loshchilov2019DecoupledWD} optimizer,
with a weight decay of 0.1. In all training procedures, we randomly initialize
the model weights.

\subsection{Evaluation Details}\label{sec:evaluation_details}

\paragraph{Evaluation protocol.} All baselines follow the same standard protocol
in which we: 1) Compute 2D-2D correspondences between the reference image and
the query image, 2) Lift these 2D-2D correspondences to 2D-3D correspondences
using the available 3D information for the reference image, 3) Estimate the
camera pose given these 2D-3D correspondences by minimizing the Reprojection
Error (RE), i.e. applying LO-RANSAC+PnP~\cite{Chum2003LocallyOR} followed by a
non-linear iterative refinement. This approach is widely used and leads to
state-of-the-art results in visual localization benchmarks. We also include results
for~\citet{NeuralReprojection} which we call S2D. For the evaluation of
Fig.~\ref{fig:hallucination_sota}, we find the inpainted and outpainted
correspondents for LoFTR~\cite{LoFTR} and DRCNet~\cite{li2020dual} by fetching
the argmax 2D coordinates in the 4D matching confidence volume. For S2D and
\neurhal, we simply take the argmax in correspondence maps for the same set of
keypoints.



\paragraph{\cite{Chum2003LocallyOR}-based pose estimator.} For all
\cite{Chum2003LocallyOR}-based methods, we estimate the camera pose using the
\href{https://github.com/mihaidusmanu/pycolmap}{pycolmap} python binding. We
tune the RANSAC threshold for optimal performance, and mark all cases where less
than 3 valid correspondences (\textit{i.e.} with a valid depth value) as failure
cases (infinite pose error). The remaining parameters are left as default. We
follow the evaluation instructions provided by each method, and use indoor
weights for SP+SG~\cite{SuperGlue} and the dual-softmax indoor weights for
LoFTR~\cite{LoFTR}. In the case of \neurhal+~\cite{Chum2003LocallyOR}, we
simply read the argmax of the predicted correspondence maps to obtain explicit
2D-to-3D correspondences.

\paragraph{\cite{NeuralReprojection}-based pose estimator.} For both S2D~\cite{NeuralReprojection} and
\neurhal we only use coarse models, which operate at either 1/8th or 1/16th of
the original input resolution. We first retrain the S2D coarse model
(fully-convolutional Inceptionv3~\cite{Inceptionv3}, up to Mixed-6e) on the same
training set as our method, with the same target resolution of 80 pixels. We
refer to this model as S2D.
Given correspondence maps and the depth map of the source image, we estimate the
camera pose between the target image and the source image using the method
proposed in~\citet{NeuralReprojection}.
For both S2D and \neurhal we use the same set of regularly sampled source
keypoints (see Sec.~\ref{sec:architecture_details}), and we perform camera pose
estimation first using P3P inside an MSAC~\cite{torr2000mlesac} loop. We run P3P
for a maximum of $5,000$ iterations over the top-20\% correspondences. We then
apply a coarse GNC~\cite{blake1987visual} over all source keypoints with
$\sigma_{max}=2.0$ and $\sigma_{min}=0.6$.  Let us highlight that in all the
camera pose experiments, the performances of NeurHal are obtained by predicting
\emph{only} low resolution correspondence maps (see
Sec.~\ref{sec:architecture_details}).

\section{Additional qualitative results}\label{sec:qualitative_results}

\subsection{Generalization to new datasets}\label{sec:generalization}

\begin{figure}
\centering
~~~~~~~~~~~~~~~~~~
\small{Source}
\hfill
\small{Target}
~~
\hfill
~~
\small{Source}
\hfill
\small{Target}
~~~~~~~~~~~~~~~~~~~~
\\
\includegraphics[width=0.47\linewidth]{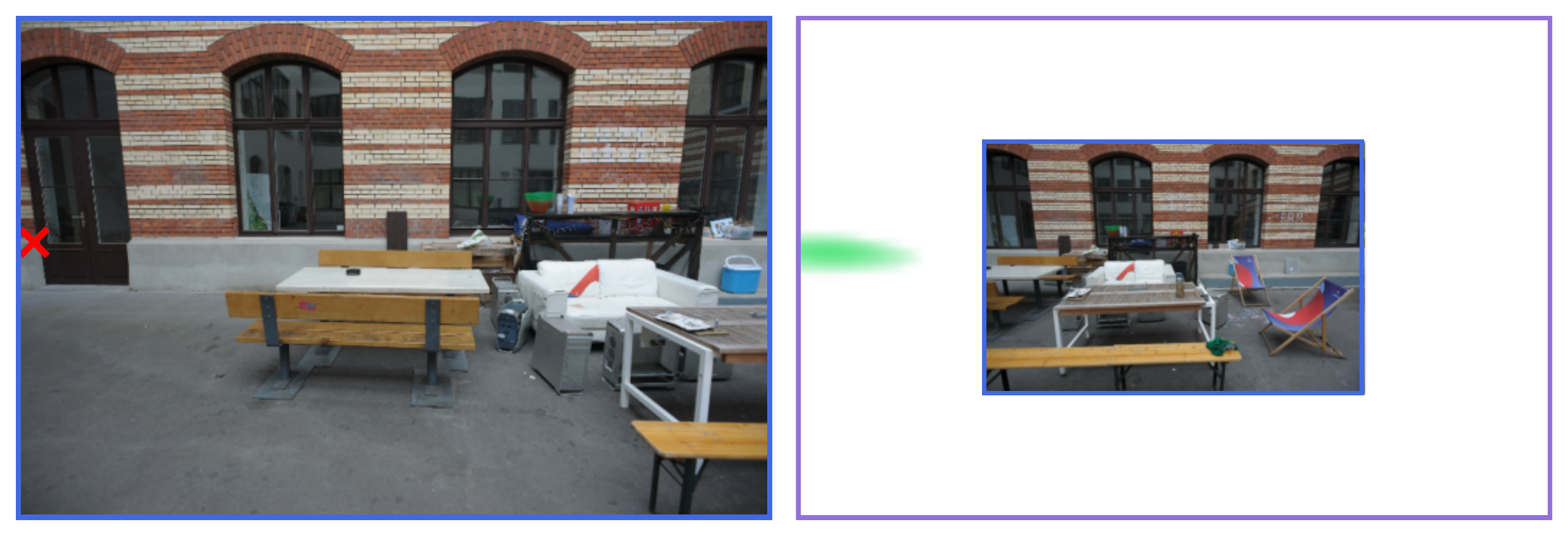}~
\includegraphics[width=0.47\linewidth]{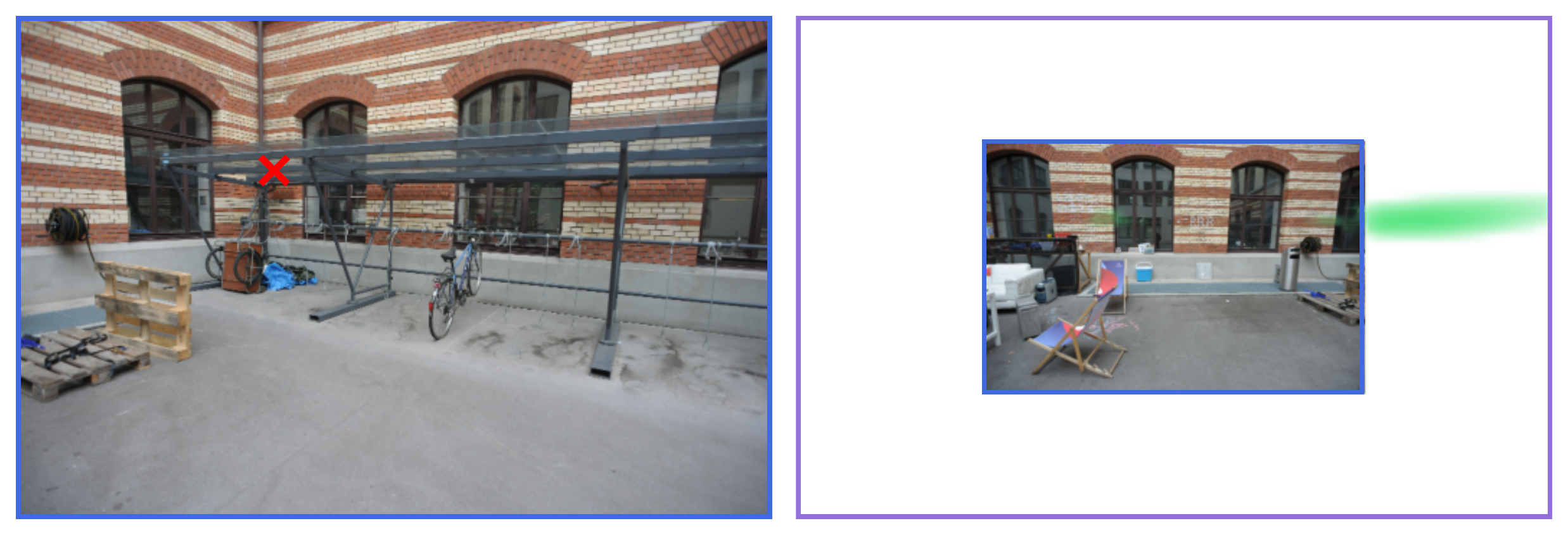}
\\
\includegraphics[width=0.47\linewidth]{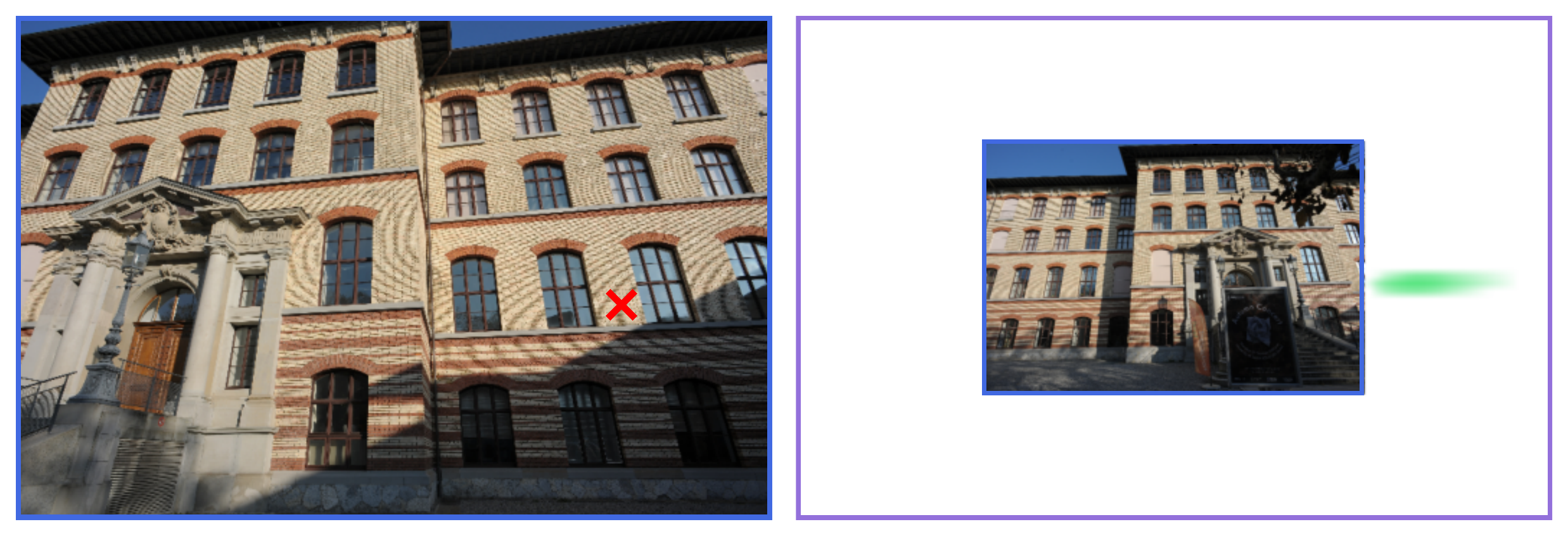}~
\includegraphics[width=0.47\linewidth]{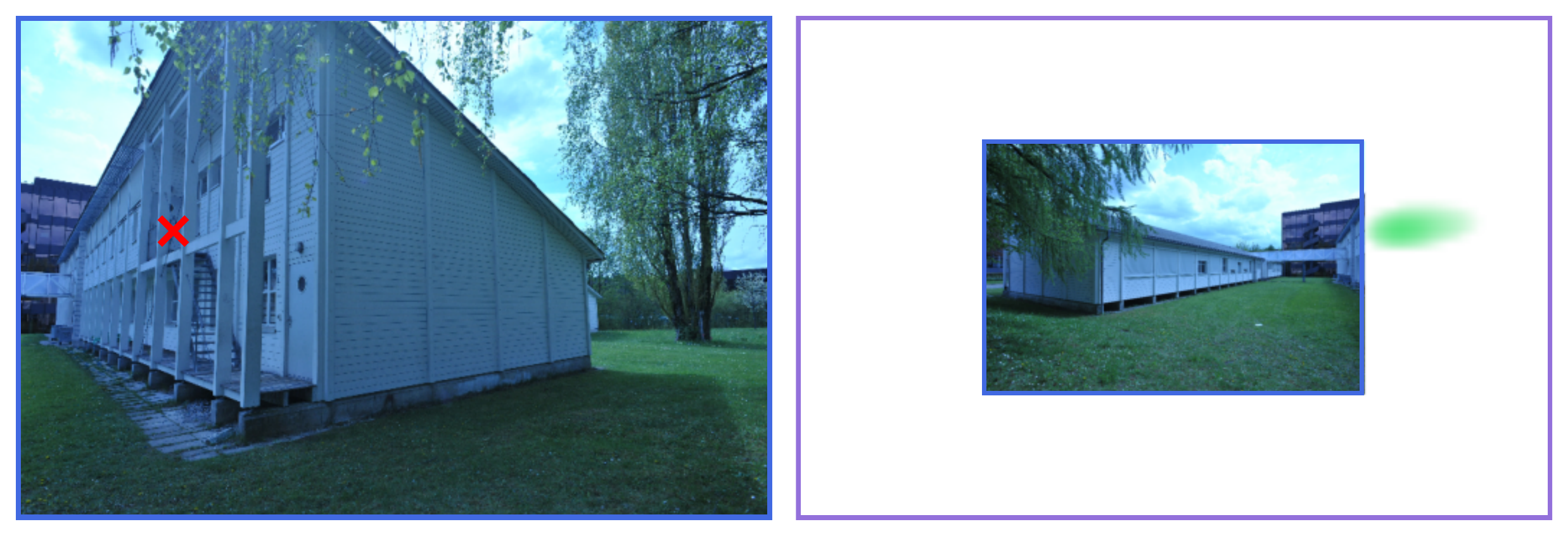}
\\
\includegraphics[width=0.47\linewidth]{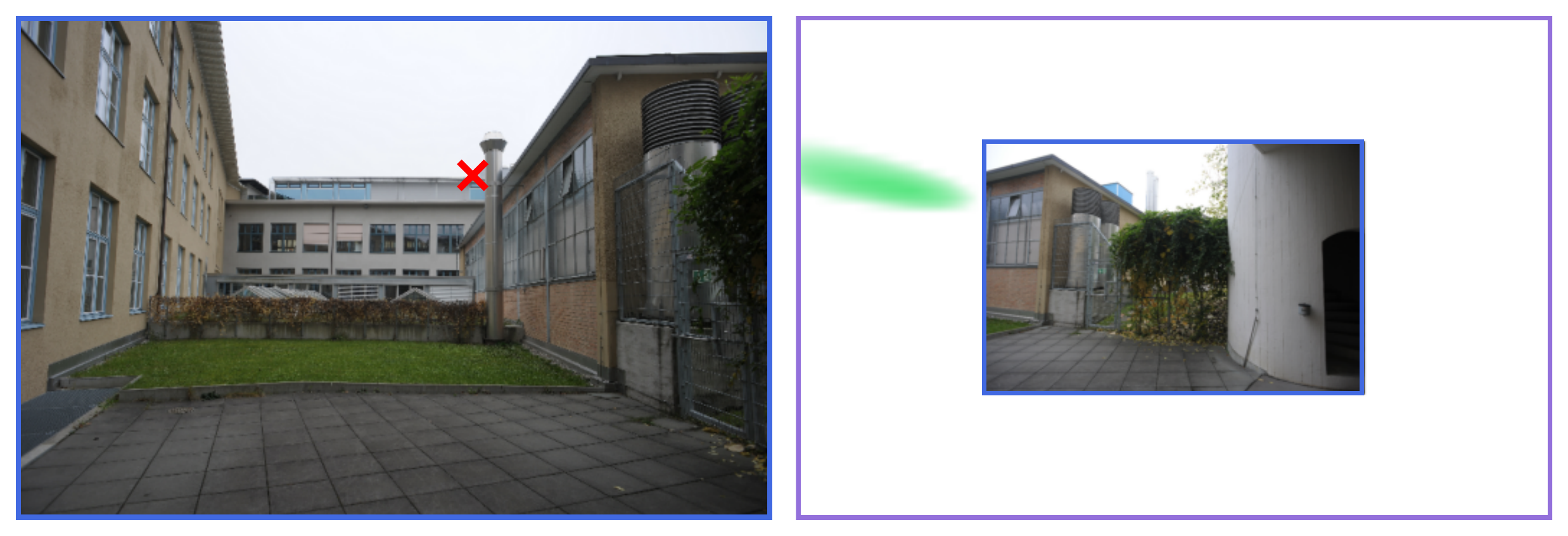}~
\includegraphics[width=0.47\linewidth]{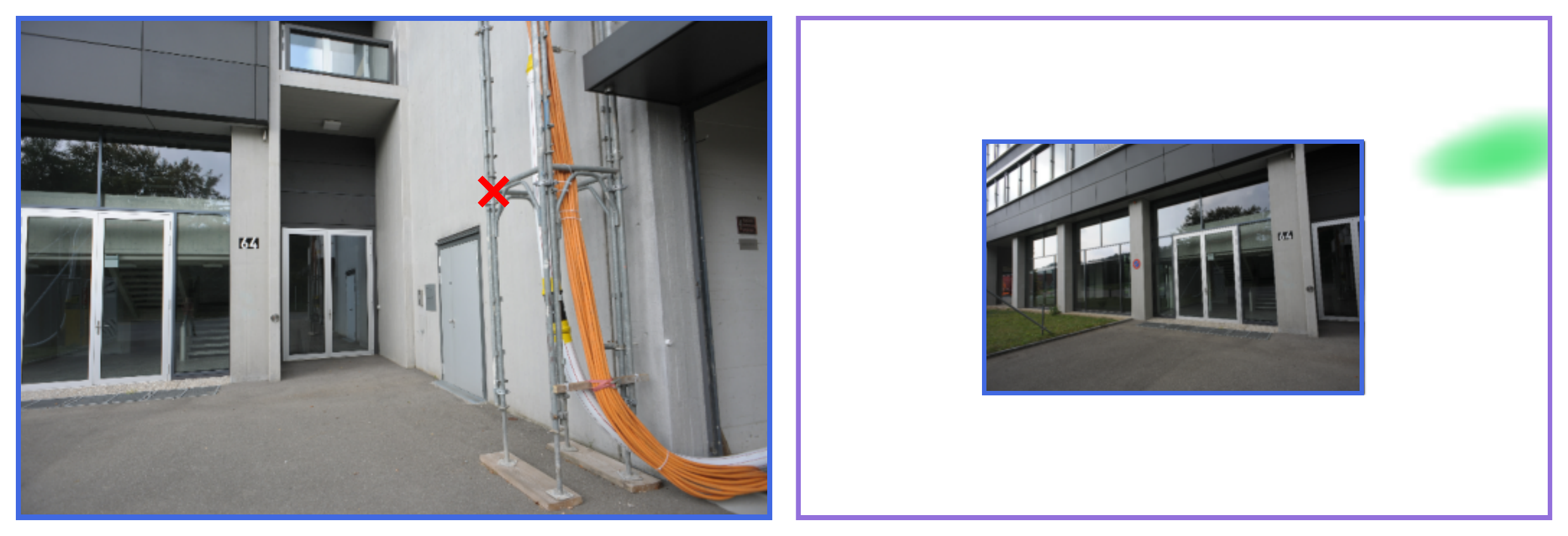}
\\
\includegraphics[width=0.47\linewidth]{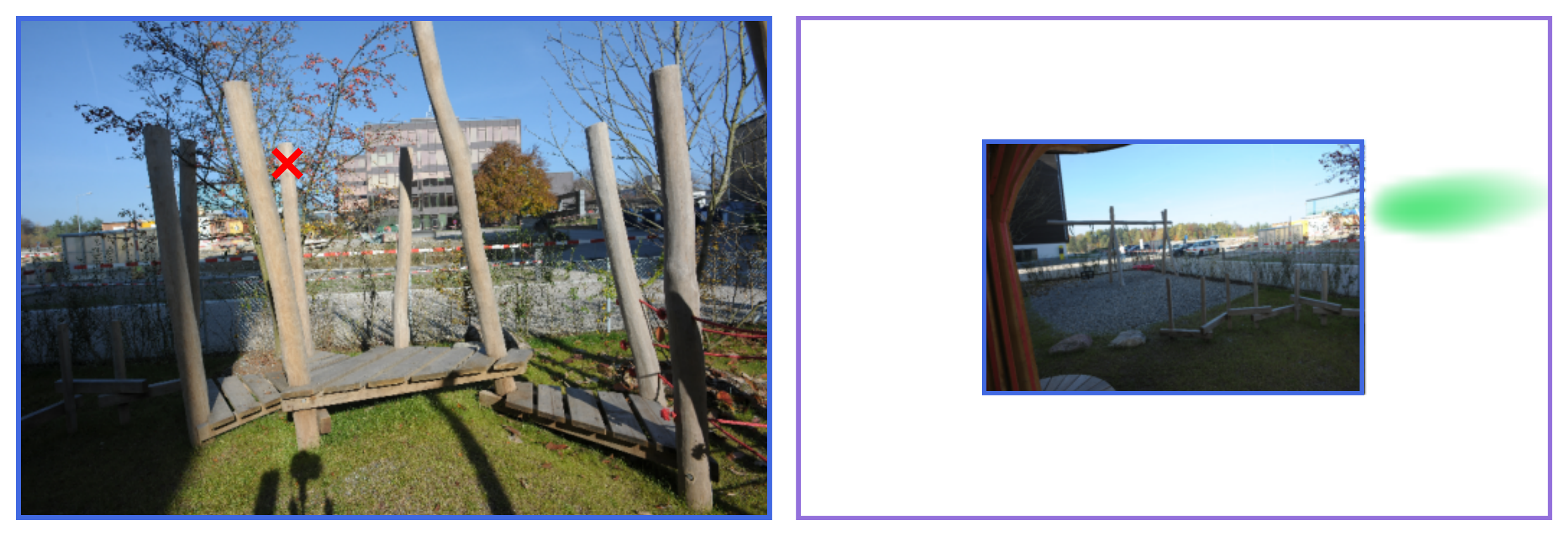}~
\includegraphics[width=0.47\linewidth]{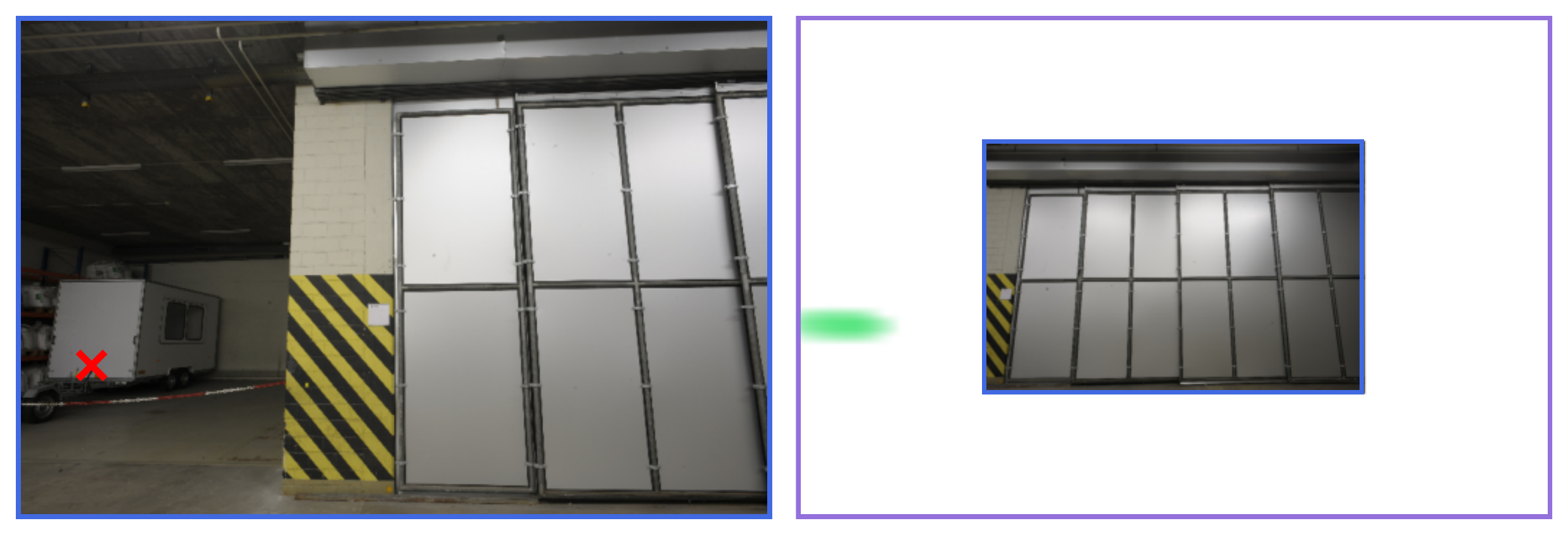}
\caption{\textbf{Qualitative results on the ETH3D dataset:} We evaluate \neurhal
on outdoor image pairs from the ETH-3D~\cite{ETH3D} dataset and find it is able to outpaint correspondences despite low visual overlaps. We report pairs of
source and target images and overlay the upsampled coarse loss map corresponding
to the source detection (in red) on the target image.}
\label{fig:eth3D}
\end{figure}

\begin{figure}
\centering
~~~~~~~~~~~~~~~~~~
\small{Source}
\hfill
\small{Target}
~~
\hfill
~~
\small{Source}
\hfill
\small{Target}
~~~~~~~~~~~~~~~~~~~~
\\
\includegraphics[width=0.47\linewidth]{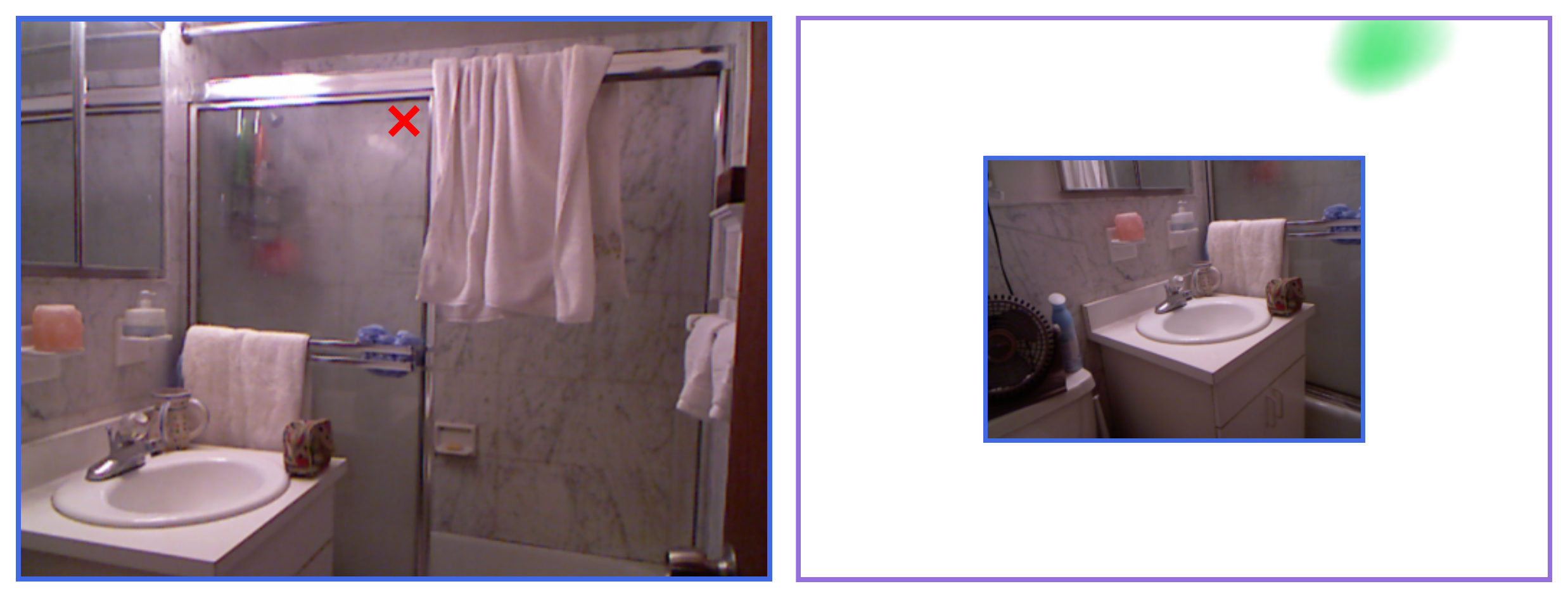}~
\includegraphics[width=0.47\linewidth]{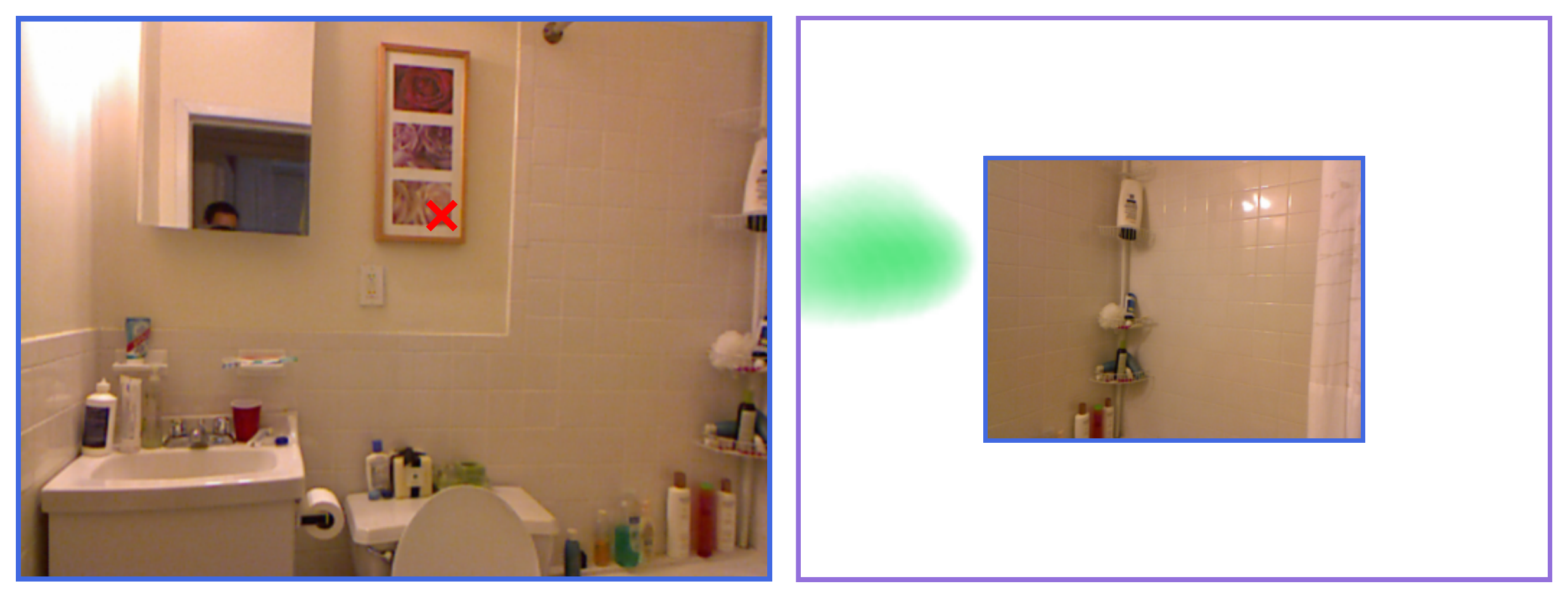}
\\
\includegraphics[width=0.47\linewidth]{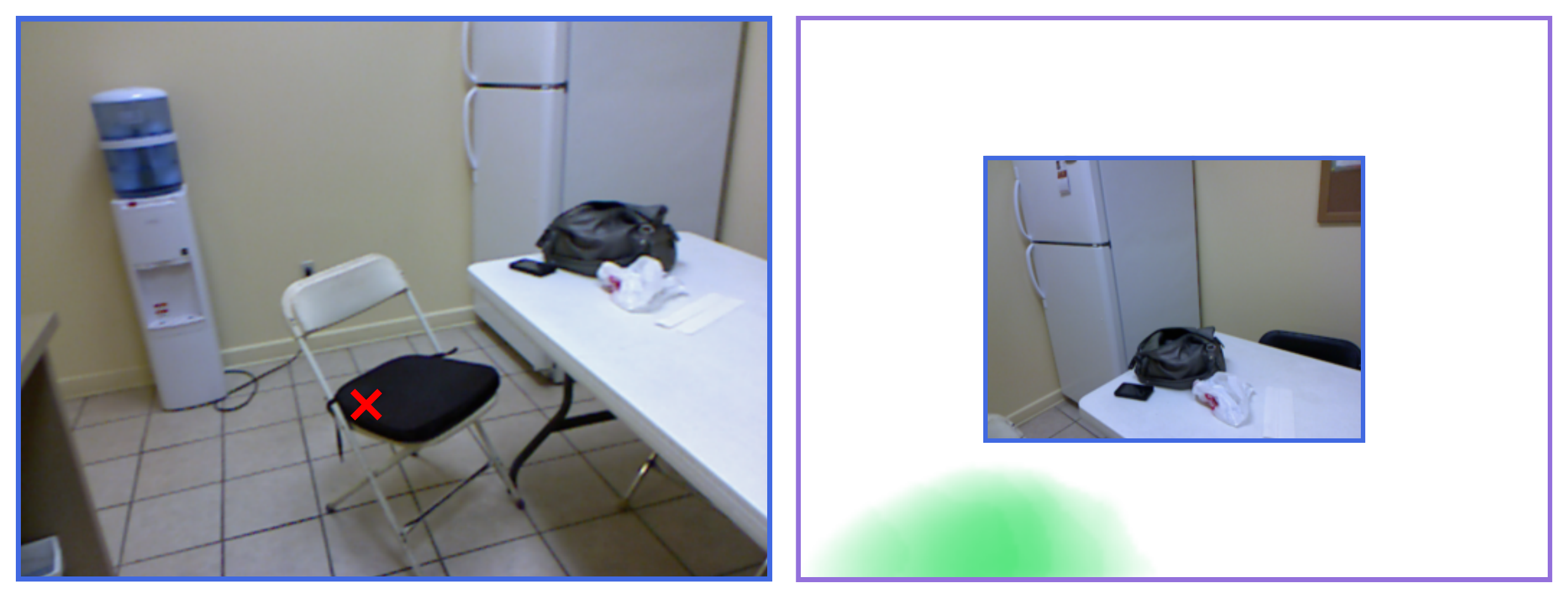}~
\includegraphics[width=0.47\linewidth]{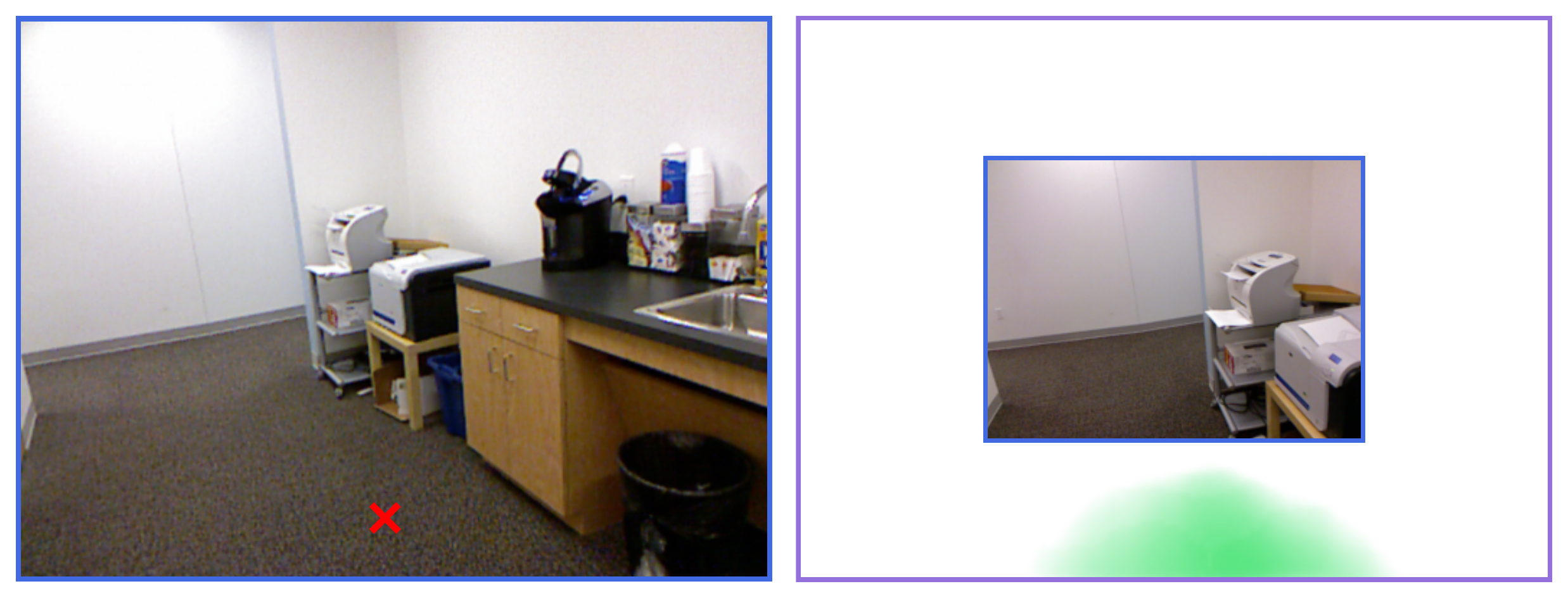}
\\
\includegraphics[width=0.47\linewidth]{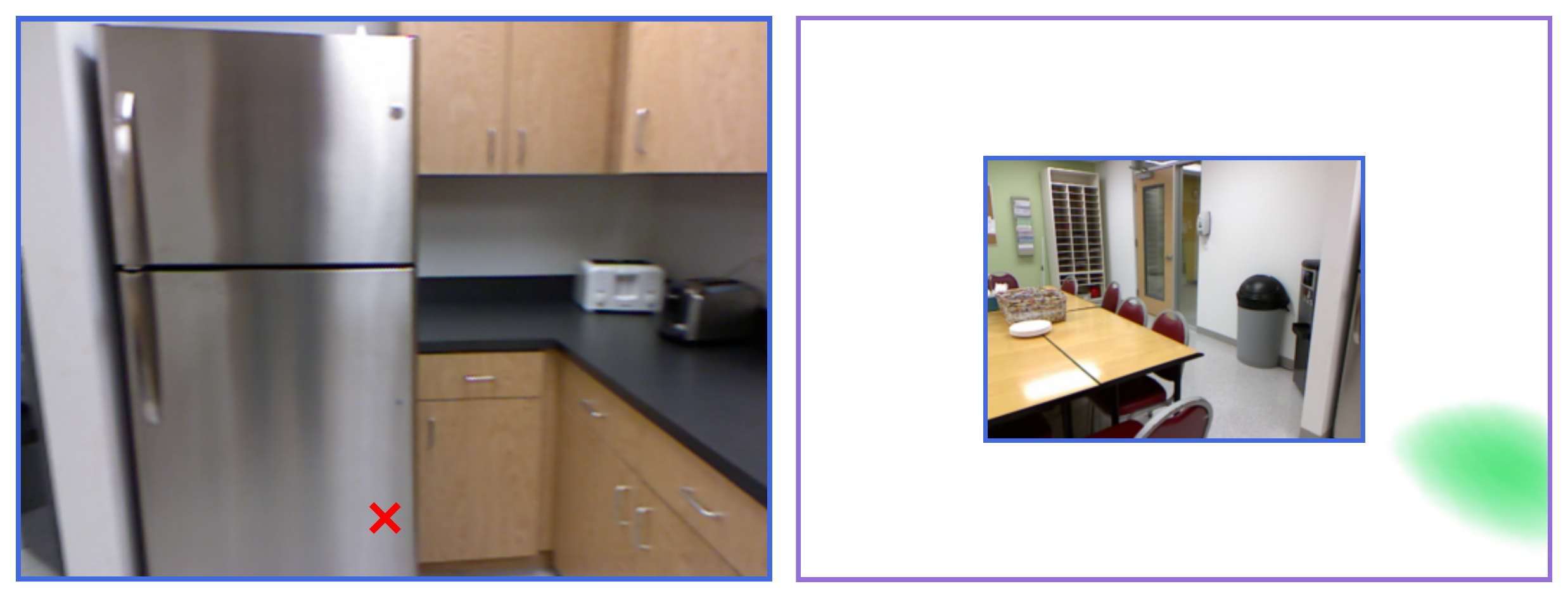}~
\includegraphics[width=0.47\linewidth]{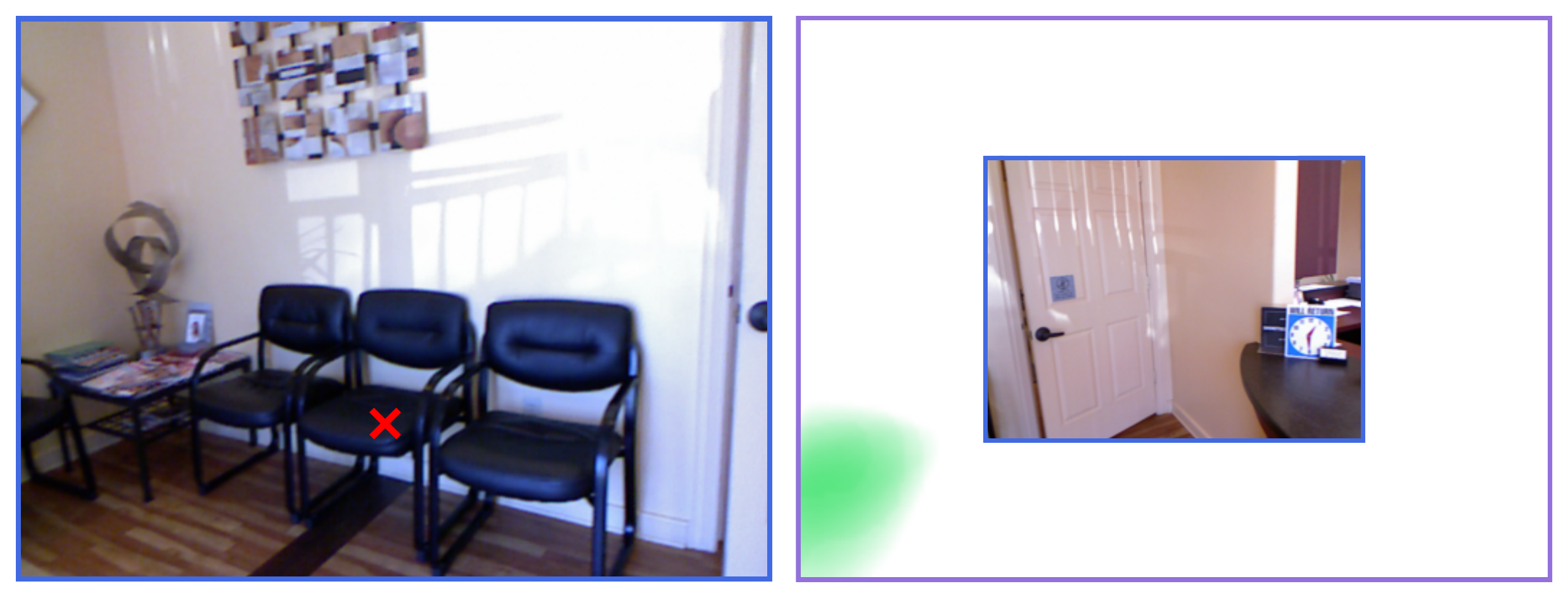}
\\
\includegraphics[width=0.47\linewidth]{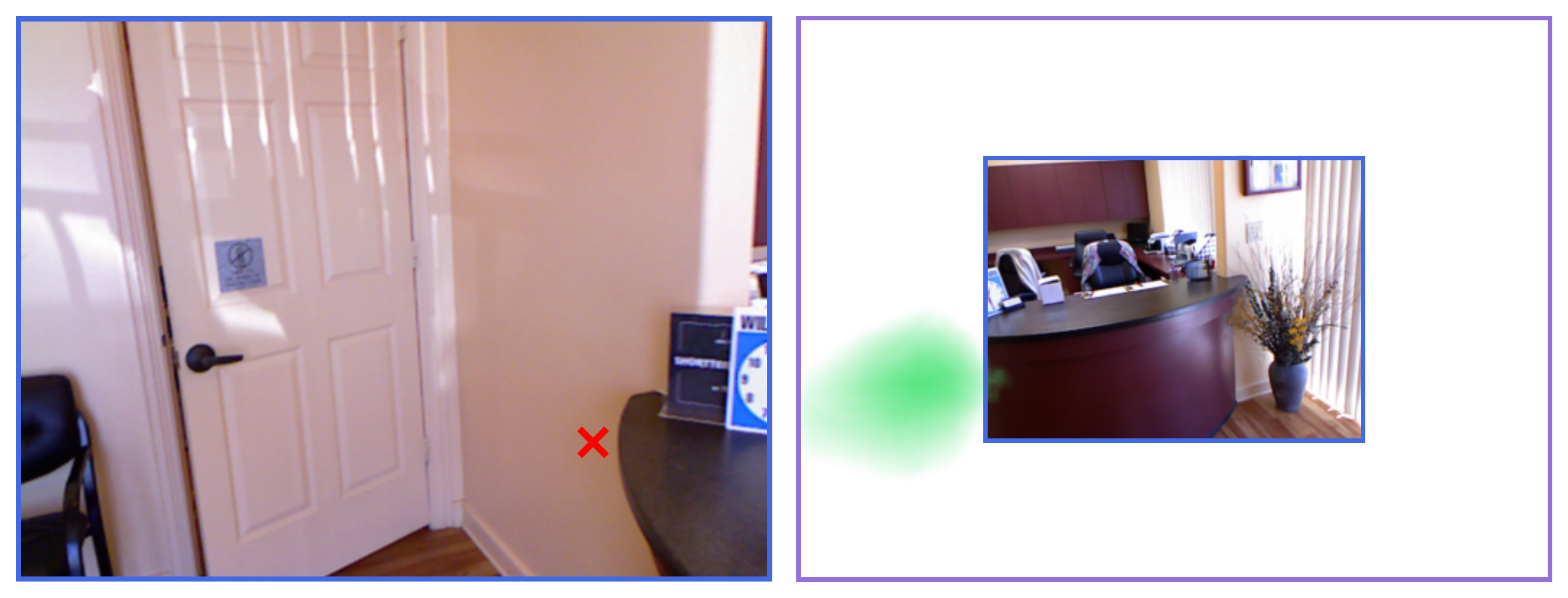}~
\includegraphics[width=0.47\linewidth]{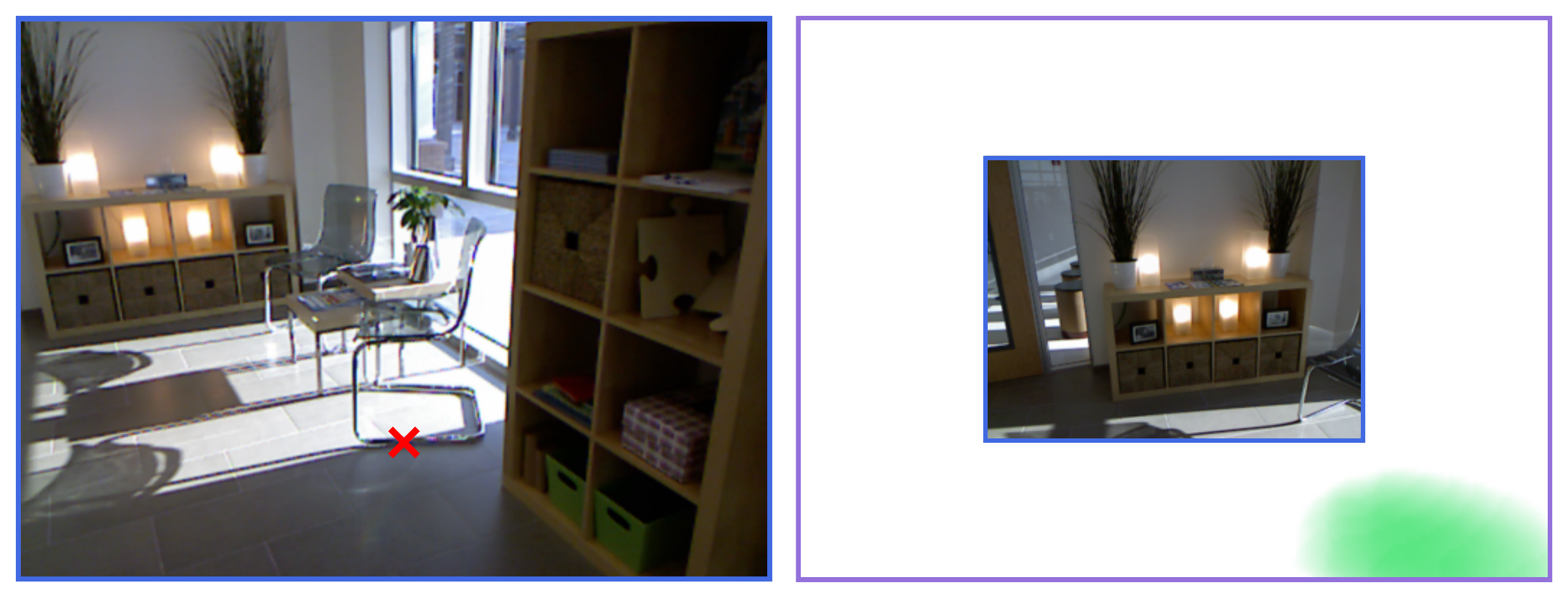}
\\
\includegraphics[width=0.47\linewidth]{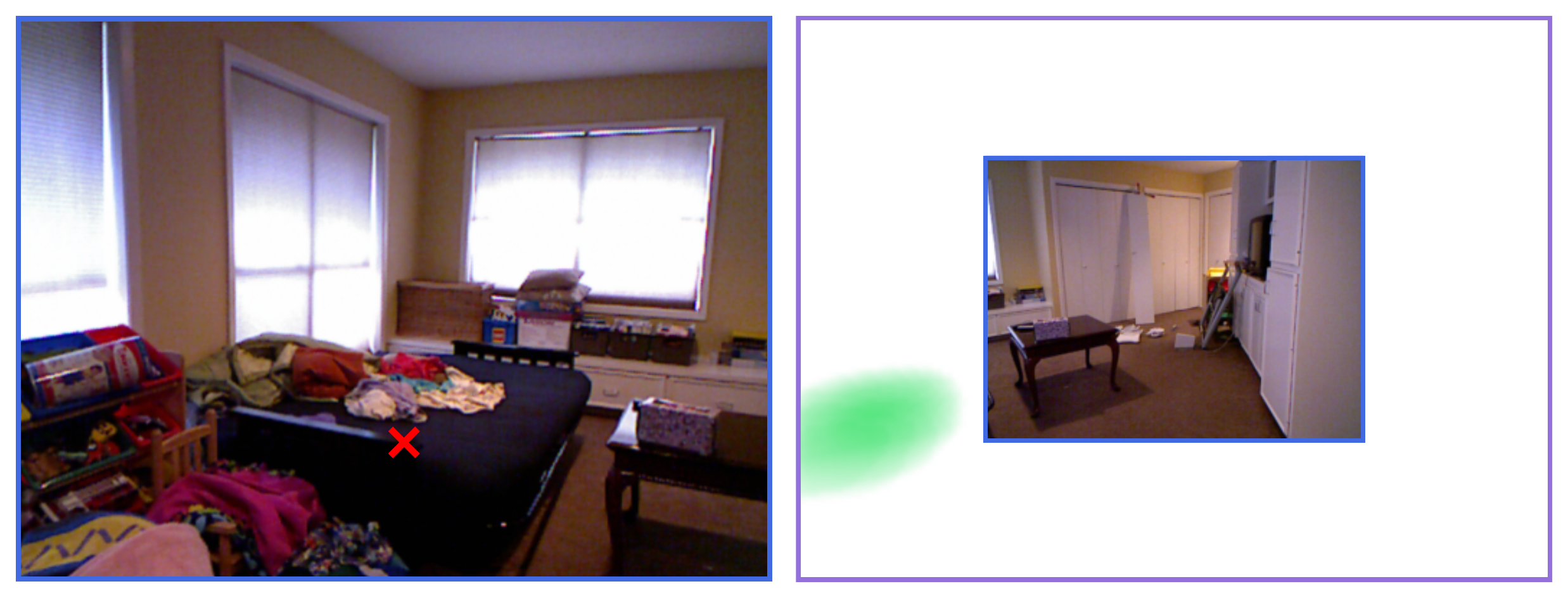}~
\includegraphics[width=0.47\linewidth]{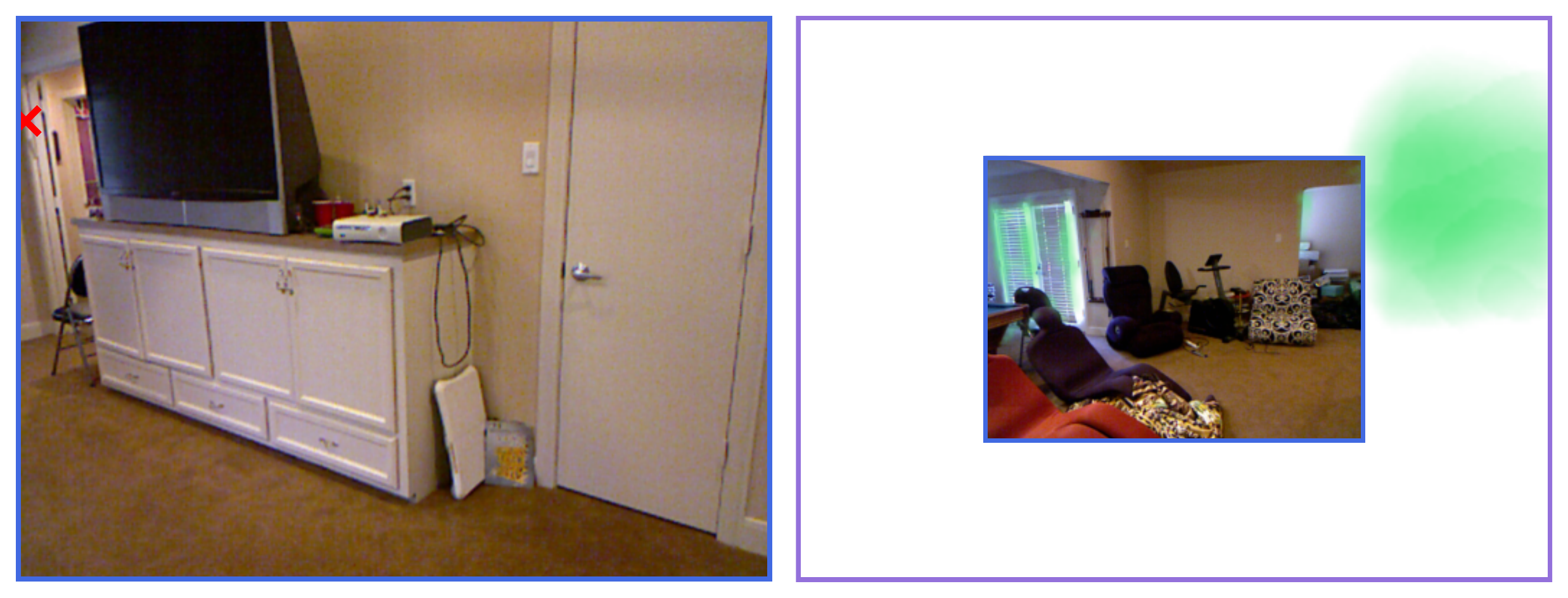}
\\
\includegraphics[width=0.47\linewidth]{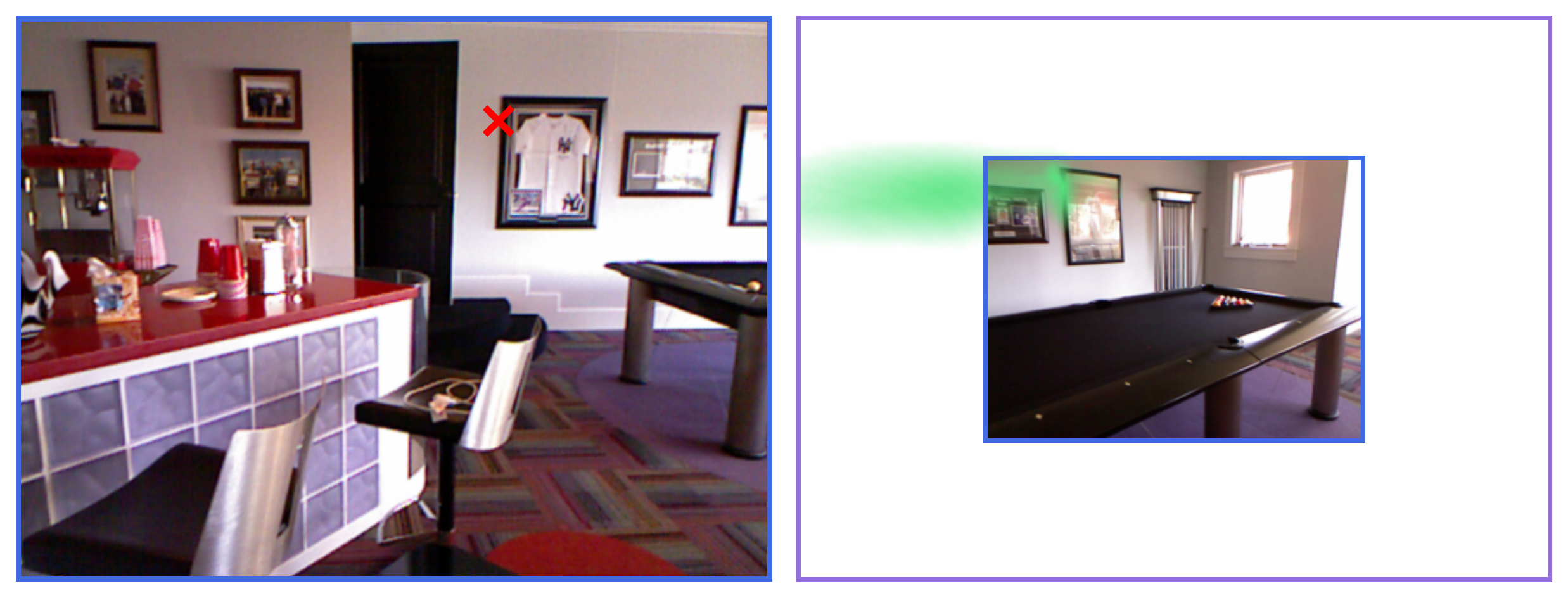}~
\includegraphics[width=0.47\linewidth]{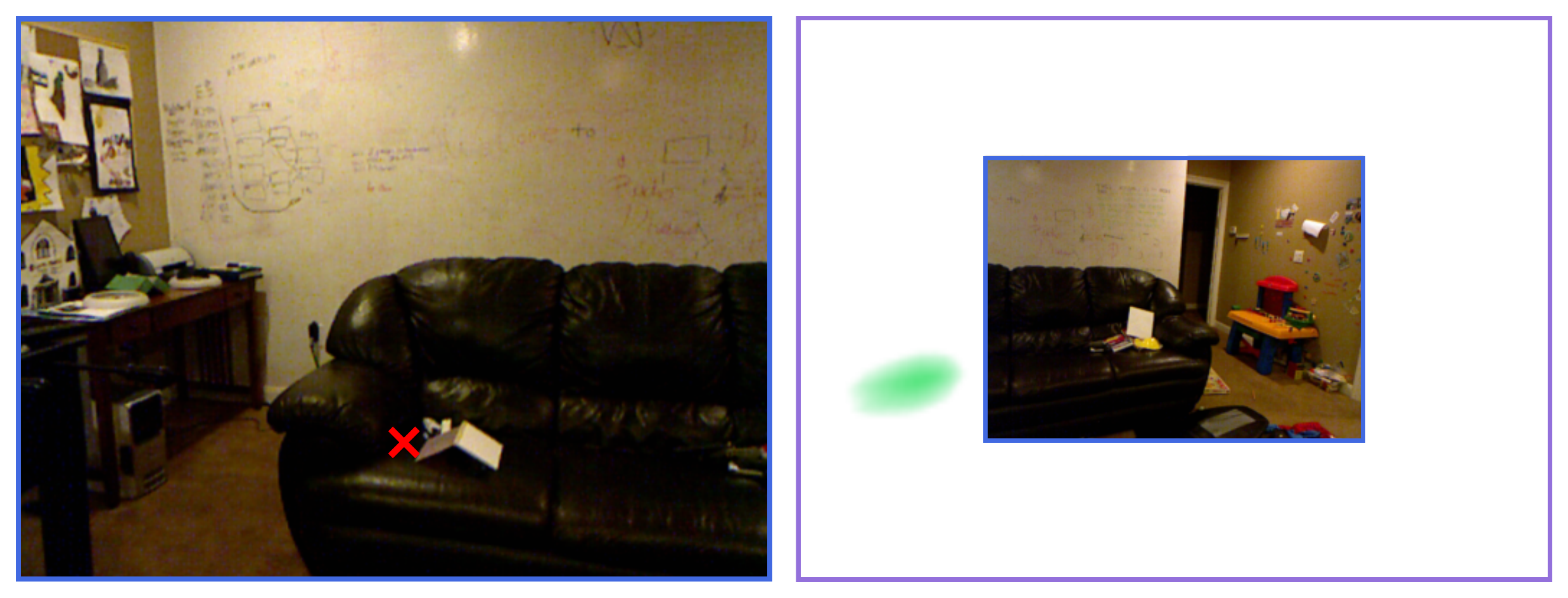}
\caption{\textbf{Qualitative results on the NYU dataset:} We evaluate \neurhal
on indoor images from the NYU~\cite{NYU} dataset and find it is able to outpaint
correspondences despite low visual overlaps. We report pairs of
source and target images and overlay the upsampled coarse loss map corresponding
to the source detection (in red) on the target image.}
\label{fig:nyu}
\end{figure}

So far we have demonstrated the ability of \neurhal to hallucinate
correspondences on unseen validation scenes from both ScanNet~\cite{ScanNet} and
Megadepth~\cite{Megadepth}. In order to further demonstrate the generalization
capacity of \neurhal, we report qualitative results obtained on the NYU Depth
Dataset~\cite{NYU} in Fig.~\ref{fig:nyu} and on the ETH-3D~\cite{ETH3D} dataset
in Fig.~\ref{fig:eth3D}. We use the set of indoor weights for NYU (\ie \neurhal
trained on ScanNet) and outdoor weights for ETH-3D (\ie \neurhal trained on
MegaDepth). We report the overlayed and upsampled coarse truncated loss map
computed following~\citet{NeuralReprojection} on low-overlap image pairs. We
find that \neurhal is able to robustly outpaint correspondences despite little
visual overlaps and strong relative camera motions. These visuals demonstrate
the strong generalization ability of \neurhal.

\subsection{Qualitative correspondence hallucination results and failure cases}

\begin{figure}
\captionsetup[subfigure]{position=b}
\centering
\raisebox{0.25in}{\rotatebox{90}{\small{\!\!\!$\text{-}\ln\C_\T$ ~~~~~~~~~~ Target ~~~~~~~ Source}}}
\subcaptionbox{Identification / Inpainting Examples}{
  \includegraphics[width=0.95\textwidth]{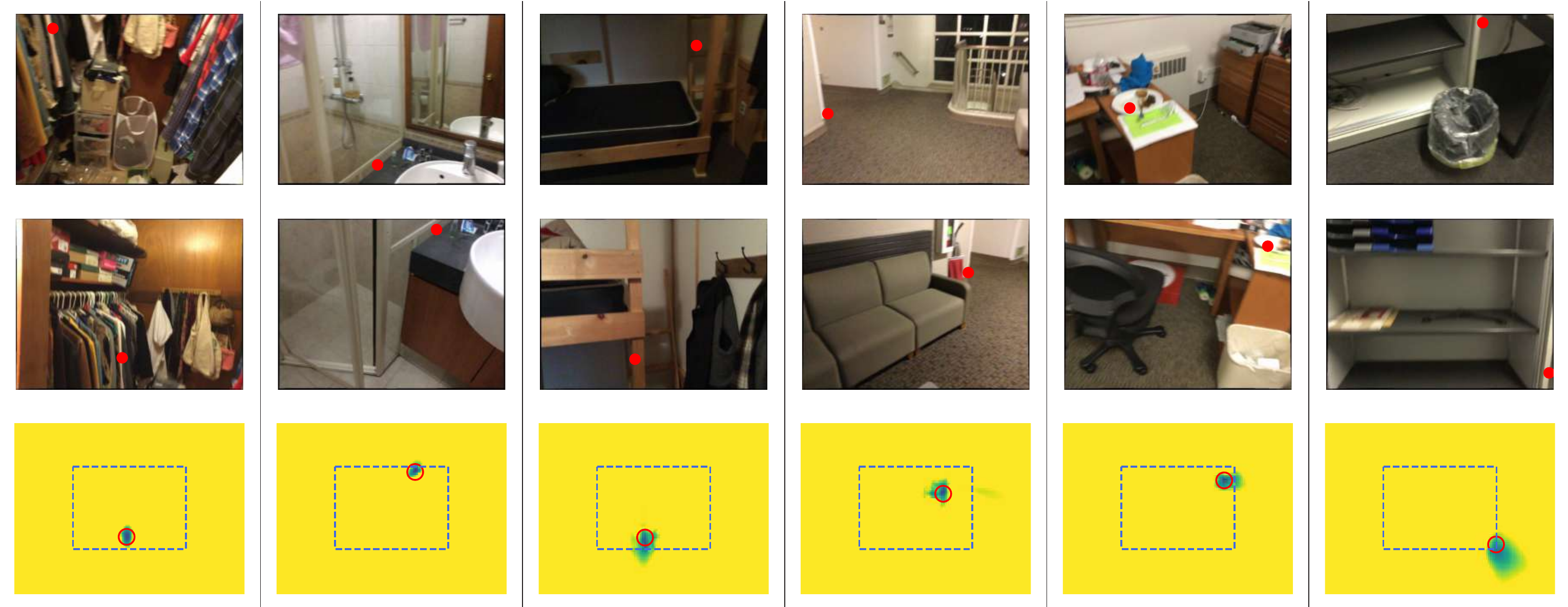}
}
\raisebox{0.25in}{\rotatebox{90}{\small{\!\!\!$\text{-}\ln\C_\T$ ~~~~~~~~~~ Target ~~~~~~~ Source}}}
\subcaptionbox{Outpainting Examples}{
  \includegraphics[width=0.95\textwidth]{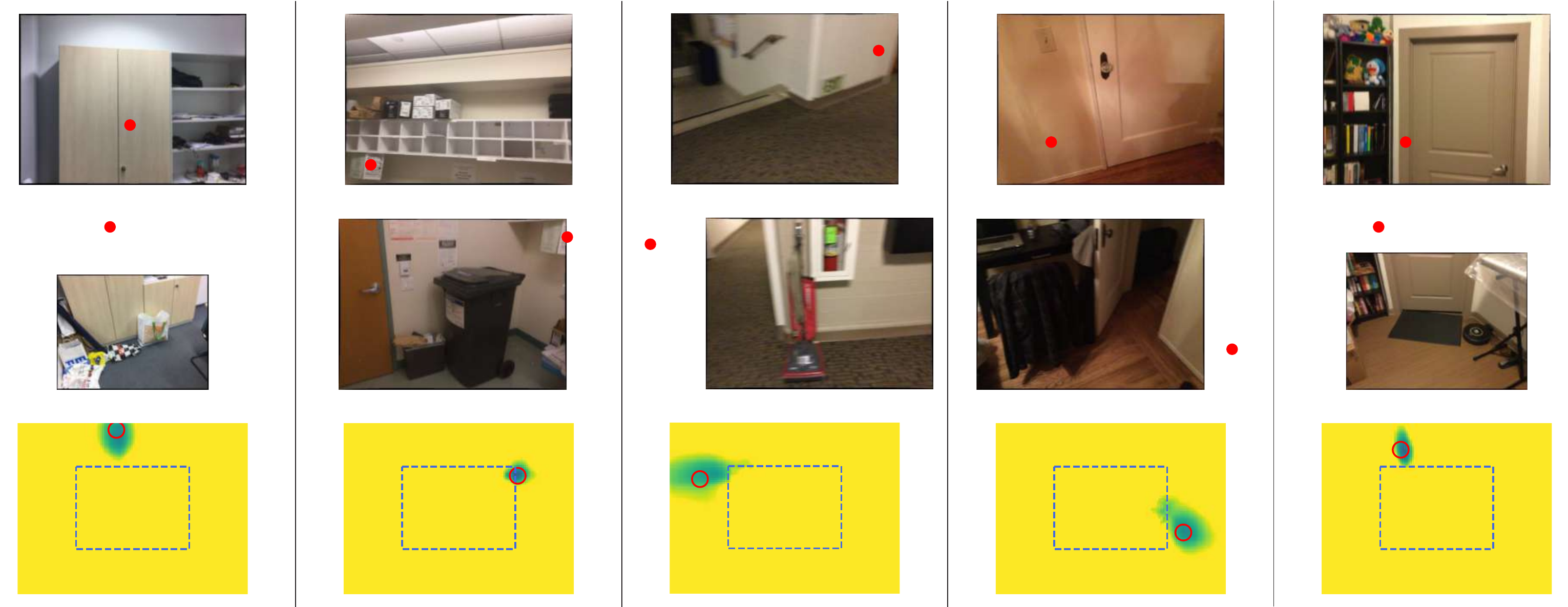}
}
\raisebox{0.25in}{\rotatebox{90}{\small{\!\!\!$\text{-}\ln\C_\T$ ~~~~~~~~~~ Target ~~~~~~~ Source}}}
\subcaptionbox{Challenging Examples}{
  \includegraphics[width=0.95\textwidth]{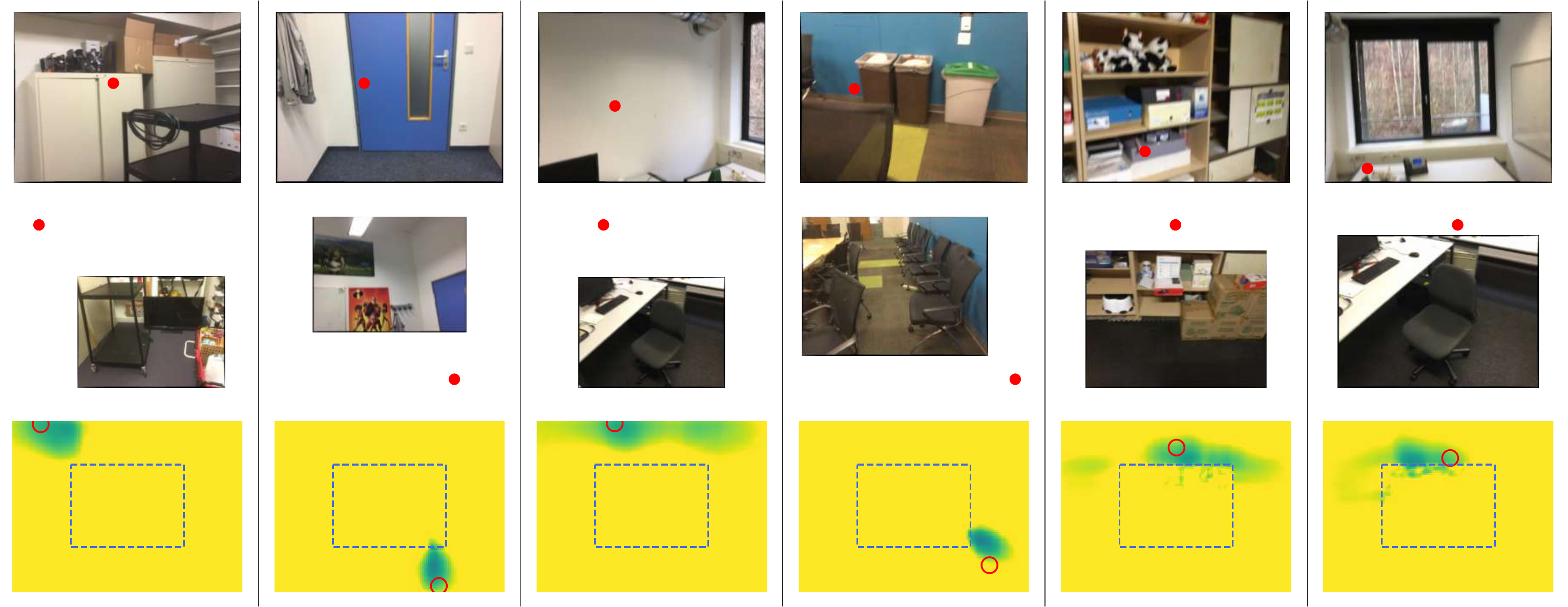}
}
\raisebox{0.25in}{\rotatebox{90}{\small{\!\!\!$\text{-}\ln\C_\T$ ~~~~~~~~~~ Target ~~~~~~~ Source}}}
\subcaptionbox{Borderline / Failure Cases}{
  \includegraphics[width=0.95\textwidth]{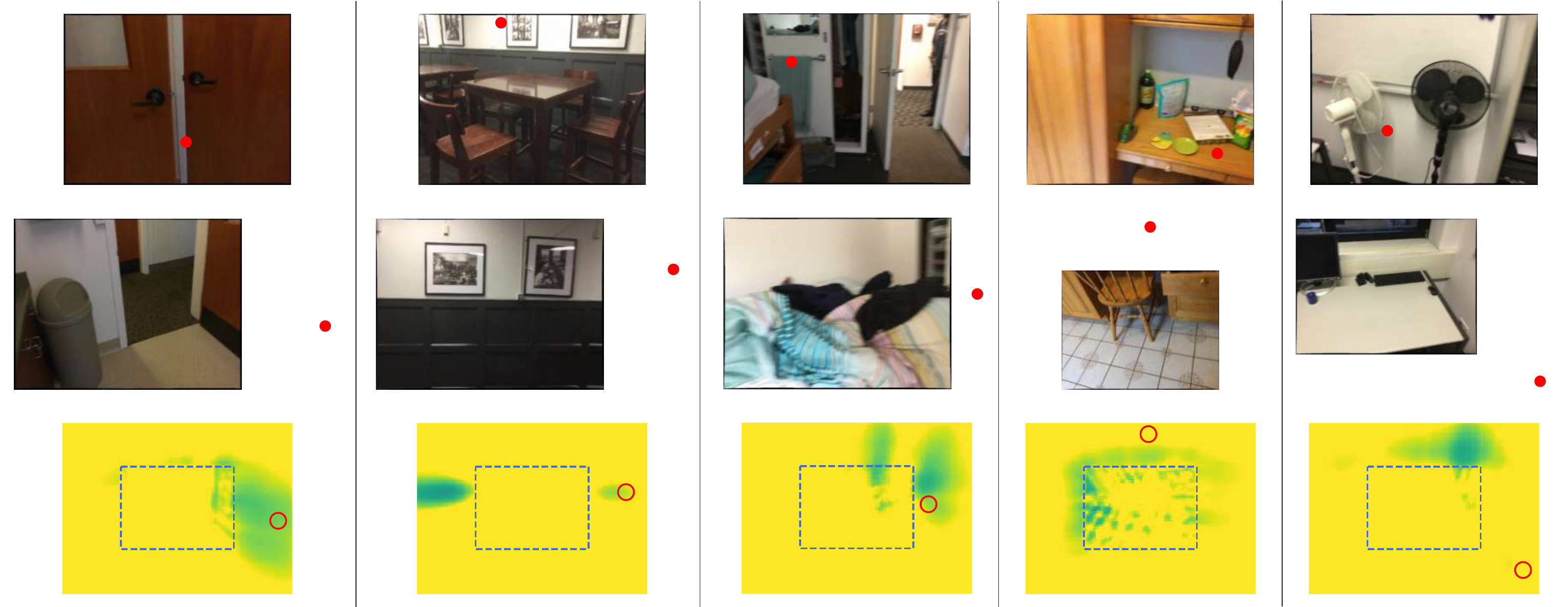}
}
\caption{\textbf{Additional qualitative ScanNet~\cite{ScanNet} examples.} See text for details.
}
\label{fig:scannet_qualitative}
\end{figure}

\begin{figure}
\captionsetup[subfigure]{position=b}
\centering
\raisebox{0.25in}{\rotatebox{90}{\small{\!\!\!$\text{-}\ln\C_\T$ ~~~~~~~~~~ Target ~~~~~~~ Source}}}
\subcaptionbox{Identification / Inpainting Examples}{
  \includegraphics[width=0.95\textwidth]{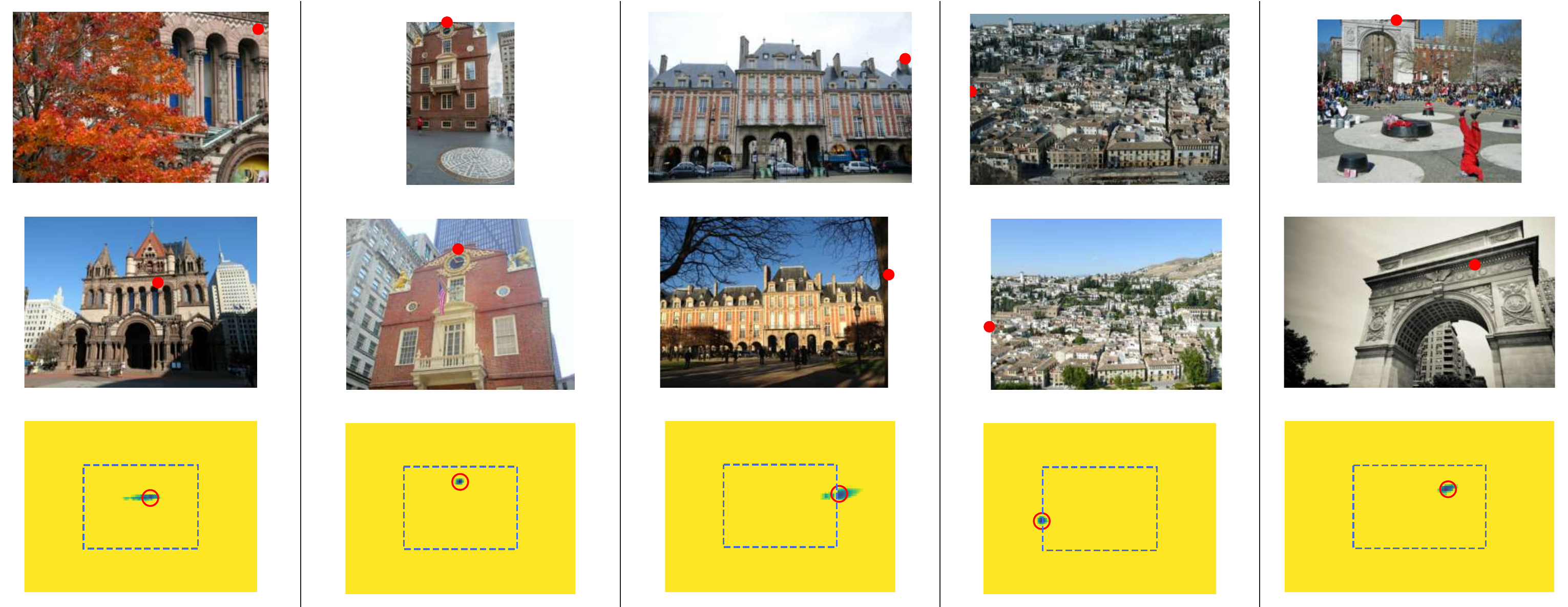}
}
\raisebox{0.25in}{\rotatebox{90}{\small{\!\!\!$\text{-}\ln\C_\T$ ~~~~~~~~~~ Target ~~~~~~~ Source}}}
\subcaptionbox{Outpainting Examples}{
  \includegraphics[width=0.95\textwidth]{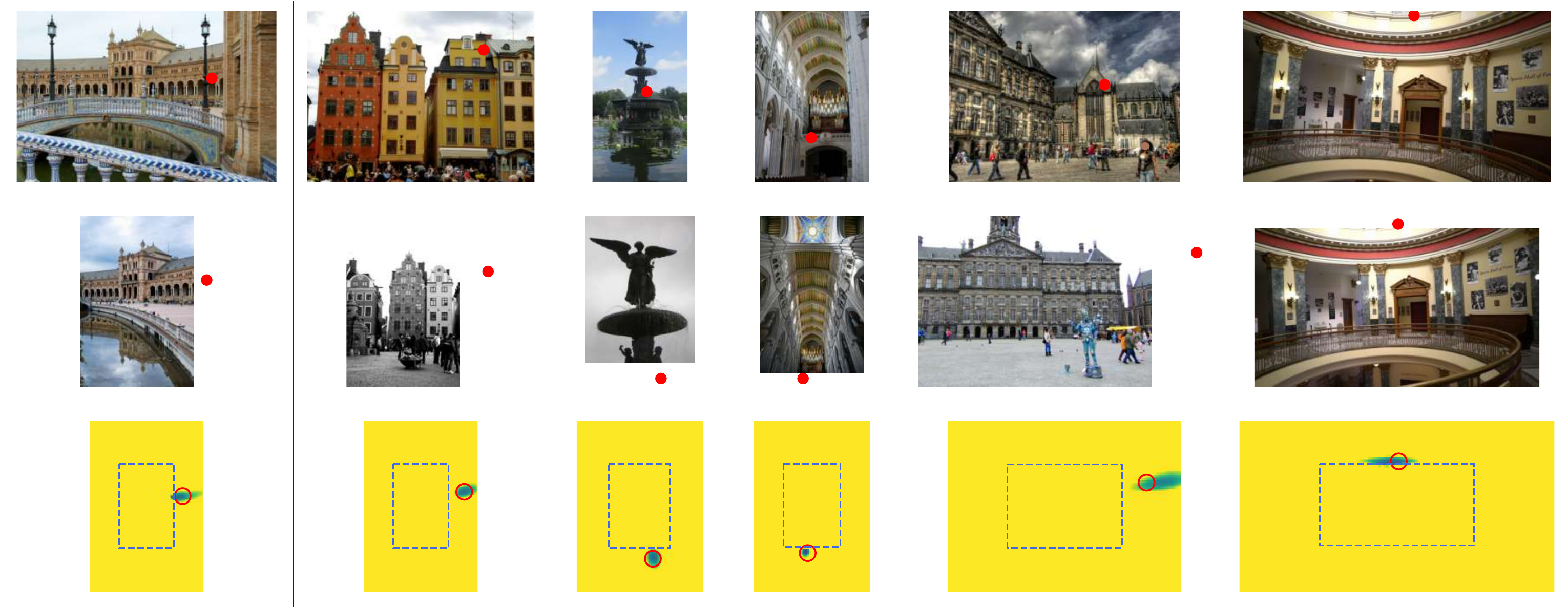}
}
\raisebox{0.25in}{\rotatebox{90}{\small{\!\!\!$\text{-}\ln\C_\T$ ~~~~~~~~~~ Target ~~~~~~~ Source}}}
\subcaptionbox{Challenging Examples}{
  \includegraphics[width=0.95\textwidth]{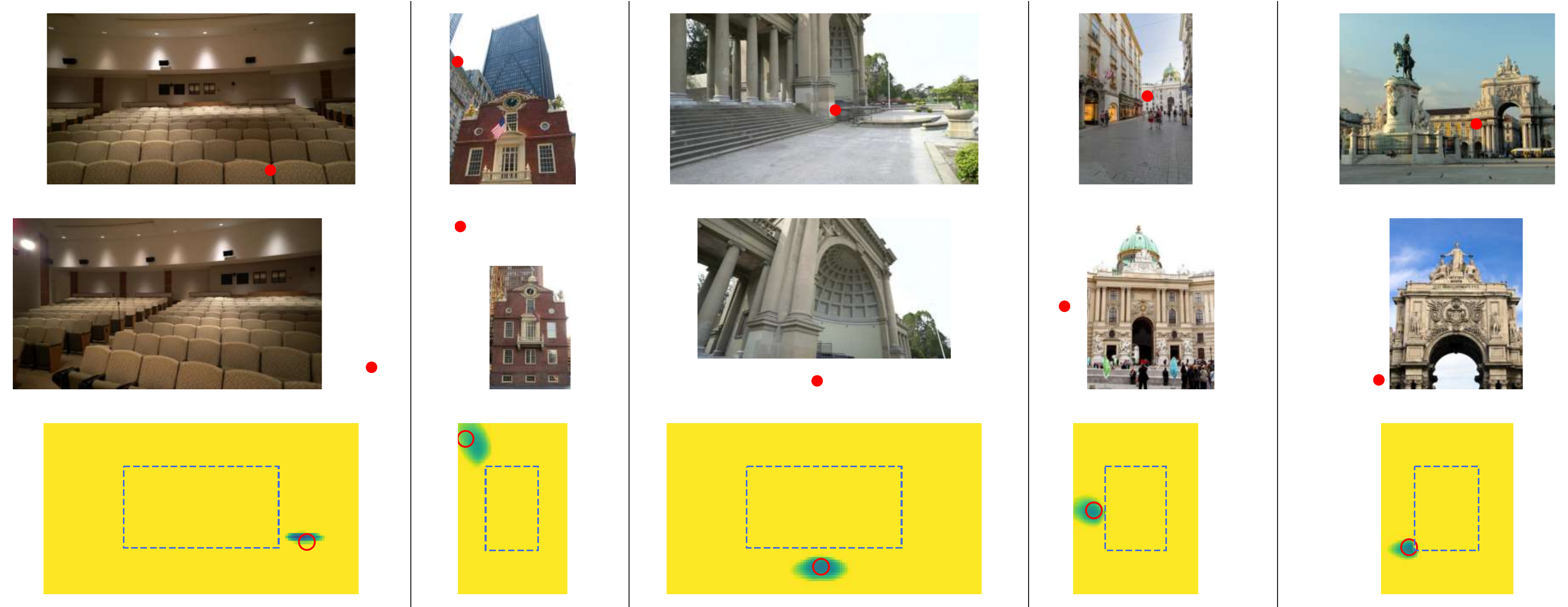}
}
\raisebox{0.25in}{\rotatebox{90}{\small{\!\!\!$\text{-}\ln\C_\T$ ~~~~~~~~~~ Target ~~~~~~~ Source}}}
\subcaptionbox{Borderline / Failure Cases}{
  \includegraphics[width=0.95\textwidth]{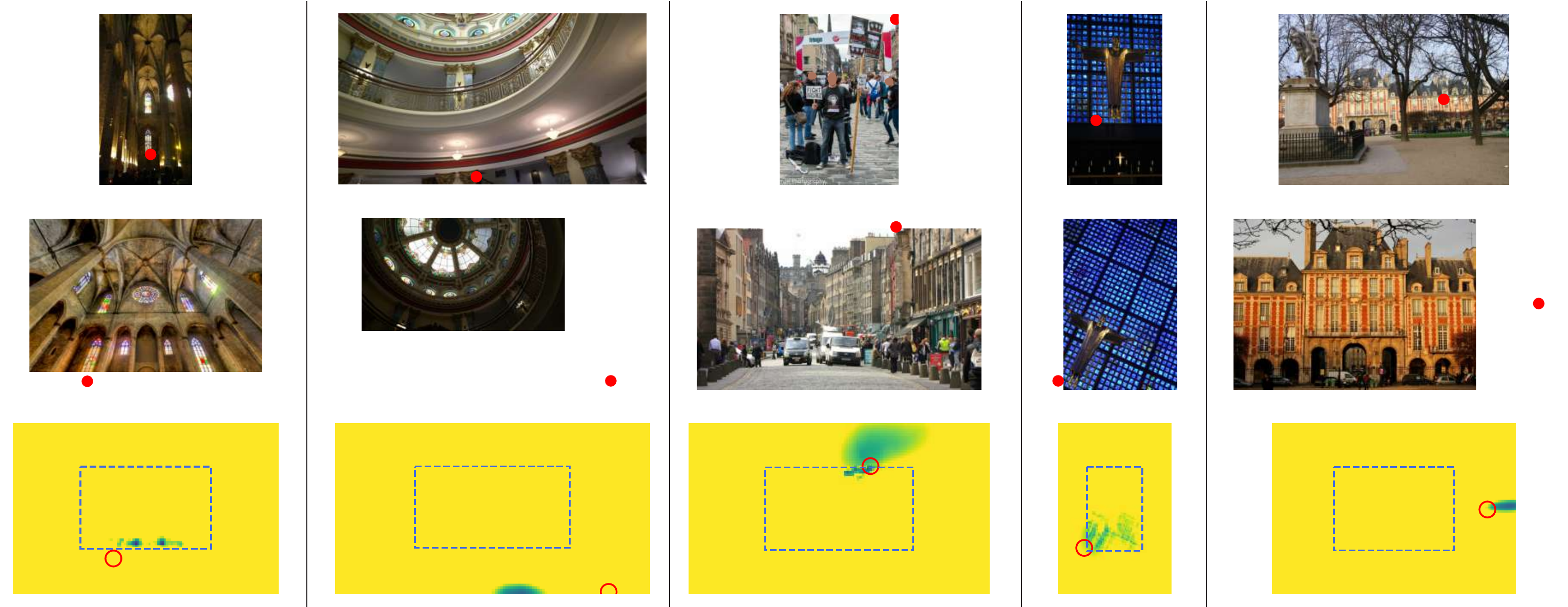}
}
\caption{\textbf{Additional qualitative Megadepth~\cite{Megadepth} examples.} See text for details.
}
\label{fig:megadepth_qualitative}
\end{figure}

To further demonstrate the ability of \neurhal to perform visual correspondence
hallucination, we report in Fig.~\ref{fig:scannet_qualitative} and
Fig.~\ref{fig:megadepth_qualitative} qualitative results on
ScanNet~\cite{ScanNet} and Megadepth~\cite{Megadepth} respectively on scenes
that were not seen at training-time.  In the target image and in the (negative
log) correspondence map, the red dot represents the ground truth keypoint's
correspondent. The dashed rectangles represent the borders of the target images.

Let us recall that \neurhal outputs probability distributions (\aka
correspondence maps) \emph{assuming the two input images are partially
overlapping}. It is essential to keep this assumption in mind when looking
at these qualitative results. For instance, concerning the example
Fig.~\ref{fig:scannet_qualitative}~(b)~(middle), it is very difficult for our
human visual system to be sure that the two images are actually overlapping, and
consequently the network prediction seems to good to be true. However, if we
\emph{assume} that there is an overlap, we realize that it is actually possible
to perform correspondence hallucination, by drawing out the two skirting
boards,  to correctly outpaint the correspondent. 

In fact, this overlapping assumption has a regularization effect in cases where
the covisible image areas show no distinctive regions, and one image could be at
an infinite translation of the other, \eg
Fig.~\ref{fig:scannet_qualitative}~(b)~(second to last).

In Fig.~\ref{fig:scannet_qualitative}~(d) and
Fig.~\ref{fig:megadepth_qualitative}~(d) we show failure cases where the
correspondence maps modes predicted by \neurhal are either partially or
completely off. We find that failure cases often correlate with strongly
ambiguous image pairs, or images that have extremely limited visual overlap.

\subsection{Qualitative camera pose estimation results}

\begin{figure}
\centering
~~~~~~~~~~~~~~~~
\small{SP+SG}
\hfill
\small{LoFTR}
\hfill
\small{DRCNet}
\hfill
\small{\neurhal}
~~~~~~~~~~~~~~~~
\\
\includegraphics[width=0.24\linewidth]{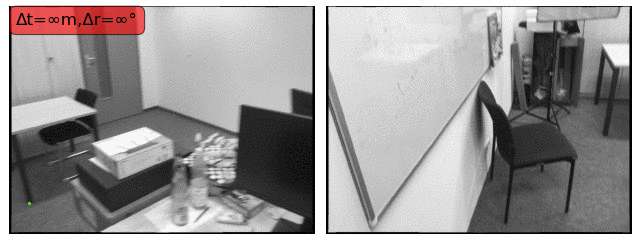}
\includegraphics[width=0.24\linewidth]{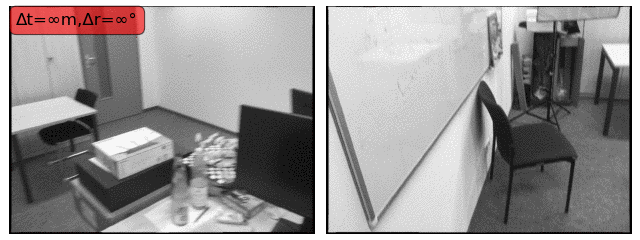}
\includegraphics[width=0.24\linewidth]{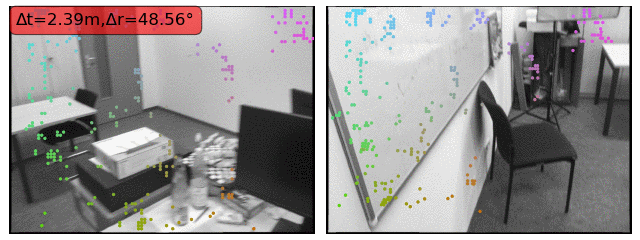}
\includegraphics[width=0.24\linewidth]{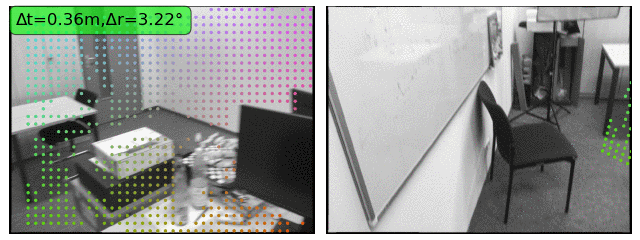}
\includegraphics[width=0.24\linewidth]{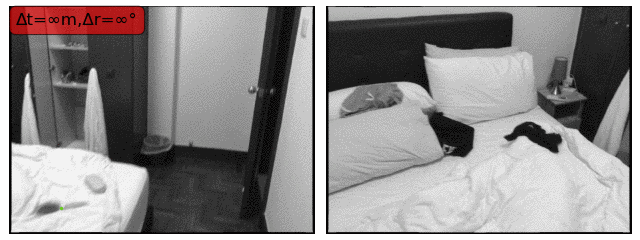}
\includegraphics[width=0.24\linewidth]{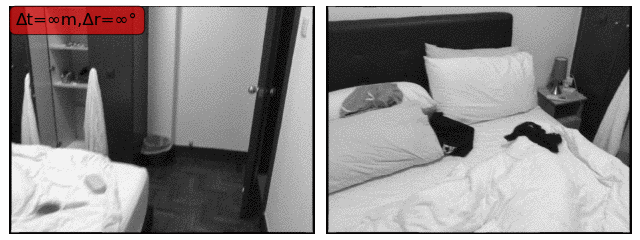}
\includegraphics[width=0.24\linewidth]{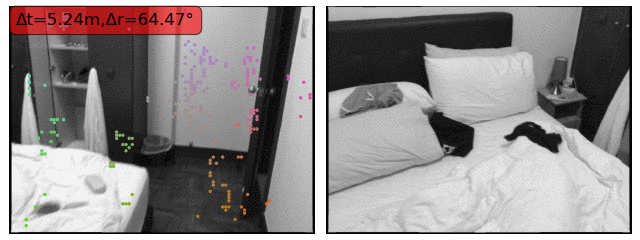}
\includegraphics[width=0.24\linewidth]{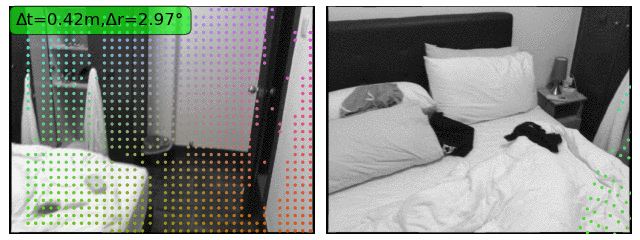}
\includegraphics[width=0.24\linewidth]{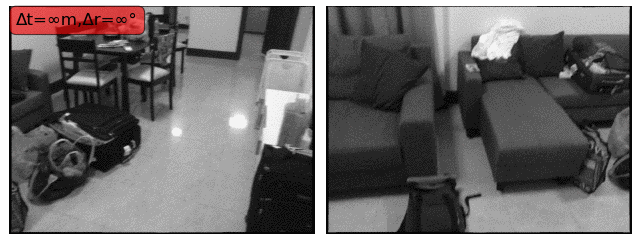}
\includegraphics[width=0.24\linewidth]{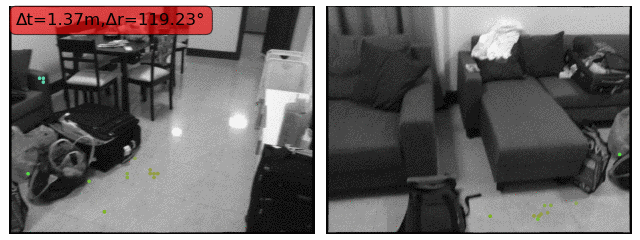}
\includegraphics[width=0.24\linewidth]{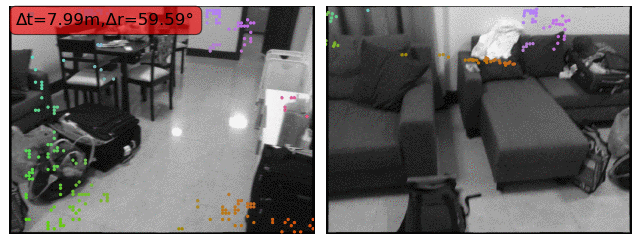}
\includegraphics[width=0.24\linewidth]{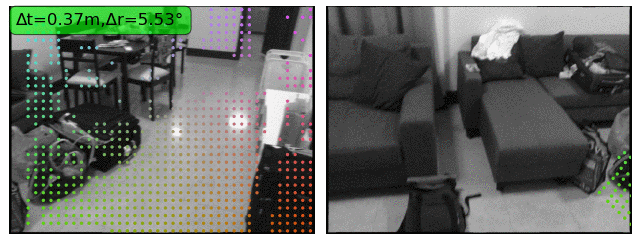}
\includegraphics[width=0.24\linewidth]{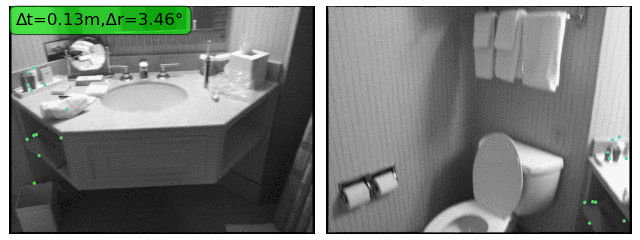}
\includegraphics[width=0.24\linewidth]{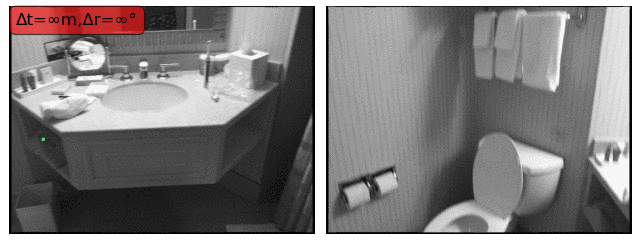}
\includegraphics[width=0.24\linewidth]{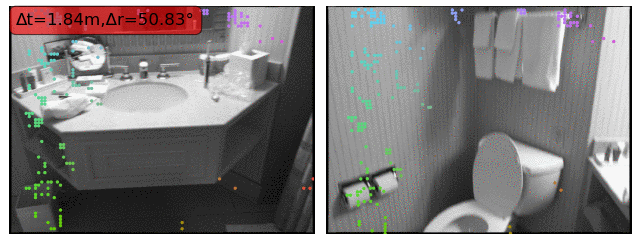}
\includegraphics[width=0.24\linewidth]{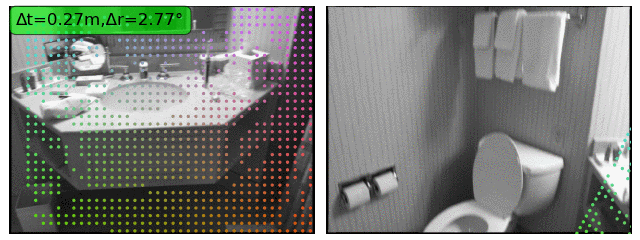}
\includegraphics[width=0.24\linewidth]{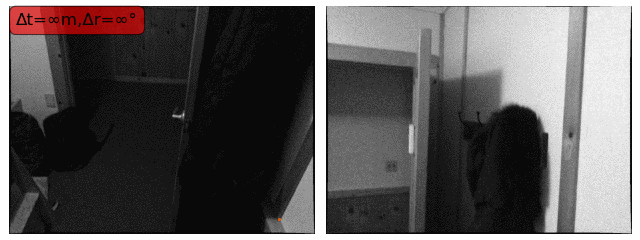}
\includegraphics[width=0.24\linewidth]{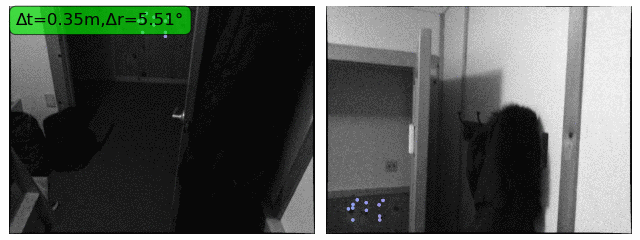}
\includegraphics[width=0.24\linewidth]{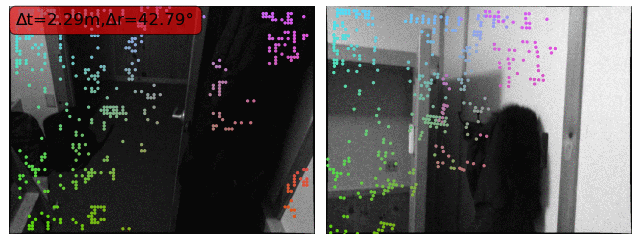}
\includegraphics[width=0.24\linewidth]{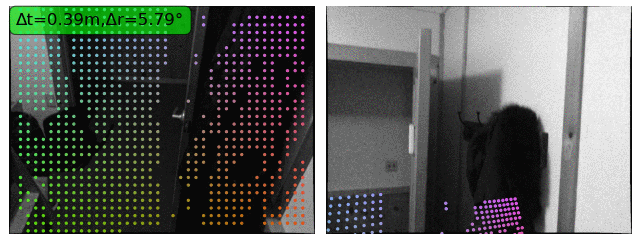}
\includegraphics[width=0.24\linewidth]{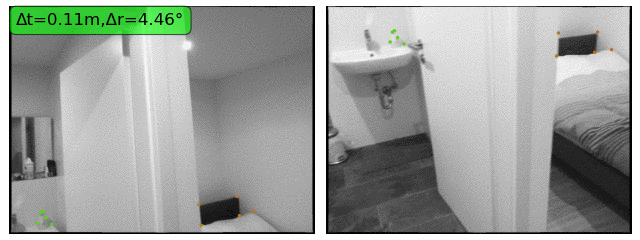}
\includegraphics[width=0.24\linewidth]{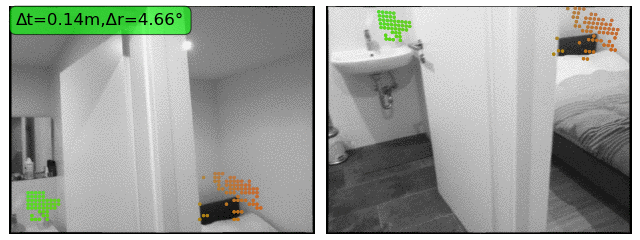}
\includegraphics[width=0.24\linewidth]{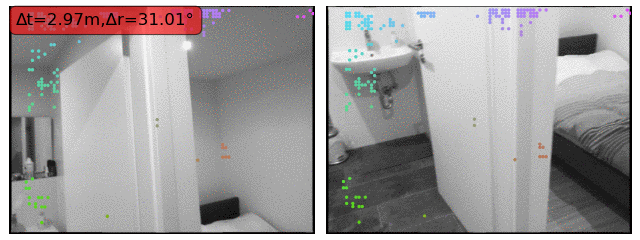}
\includegraphics[width=0.24\linewidth]{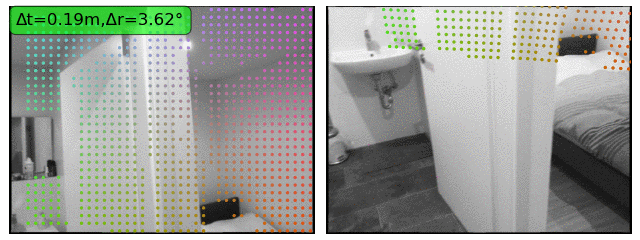}
\includegraphics[width=0.24\linewidth]{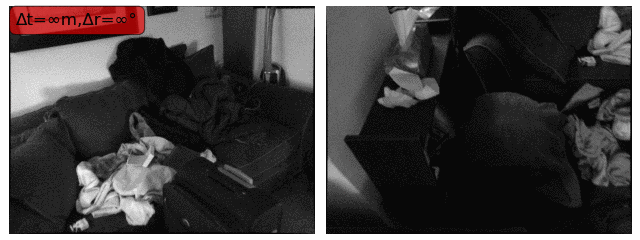}
\includegraphics[width=0.24\linewidth]{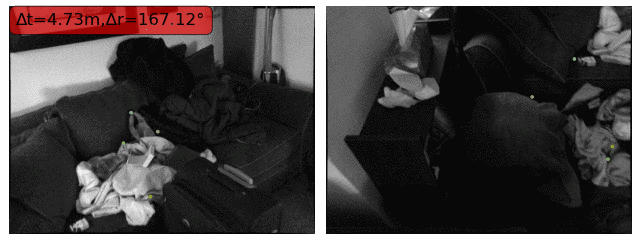}
\includegraphics[width=0.24\linewidth]{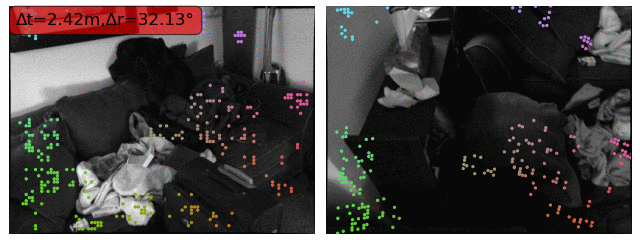}
\includegraphics[width=0.24\linewidth]{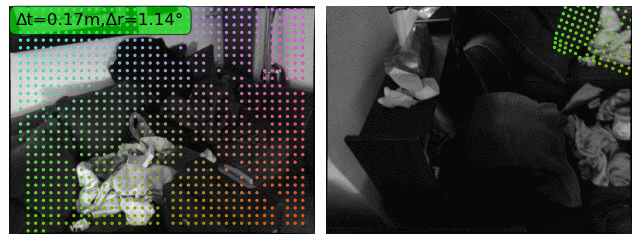}
\includegraphics[width=0.24\linewidth]{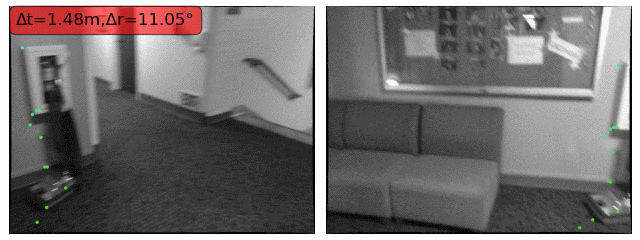}
\includegraphics[width=0.24\linewidth]{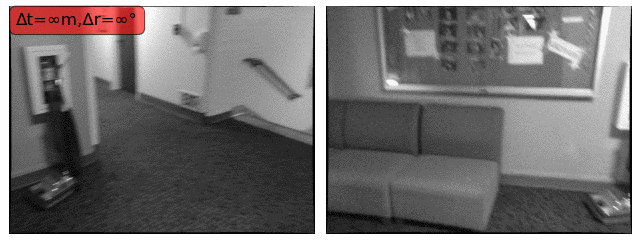}
\includegraphics[width=0.24\linewidth]{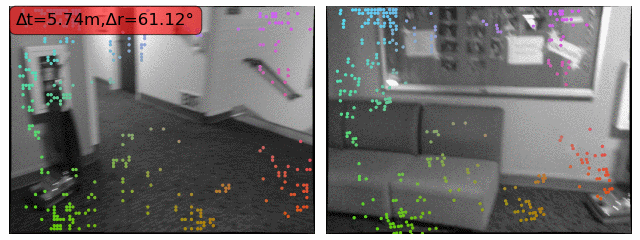}
\includegraphics[width=0.24\linewidth]{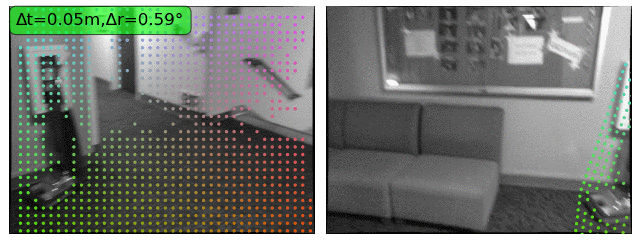}
\includegraphics[width=0.24\linewidth]{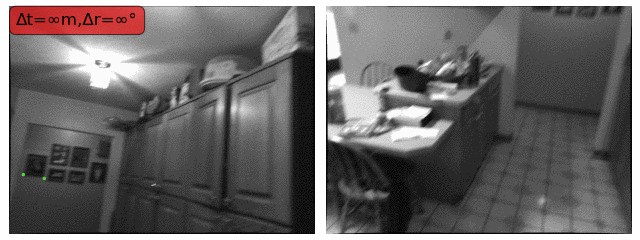}
\includegraphics[width=0.24\linewidth]{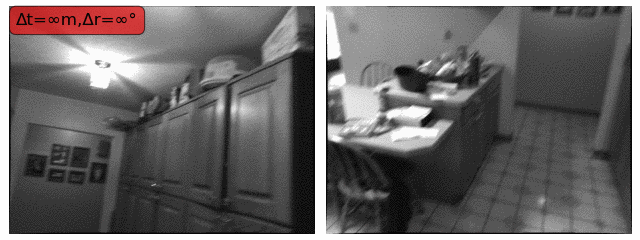}
\includegraphics[width=0.24\linewidth]{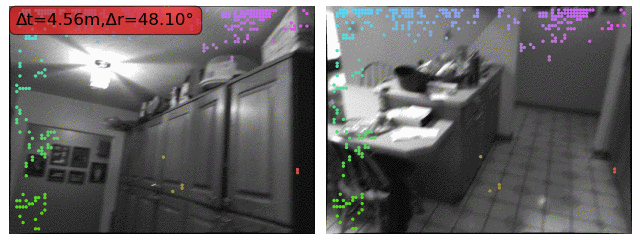}
\includegraphics[width=0.24\linewidth]{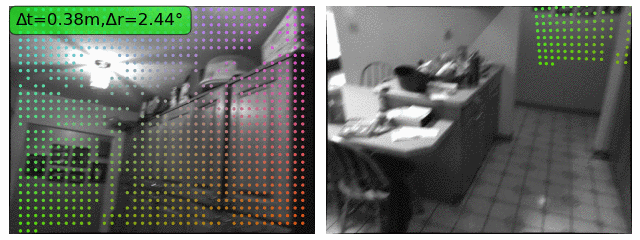}
\includegraphics[width=0.24\linewidth]{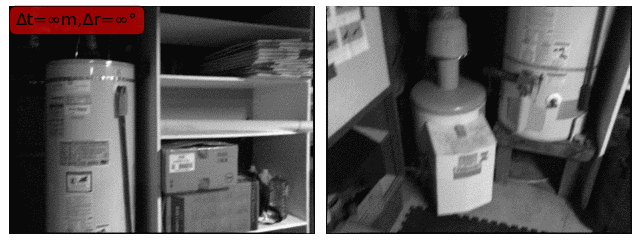}
\includegraphics[width=0.24\linewidth]{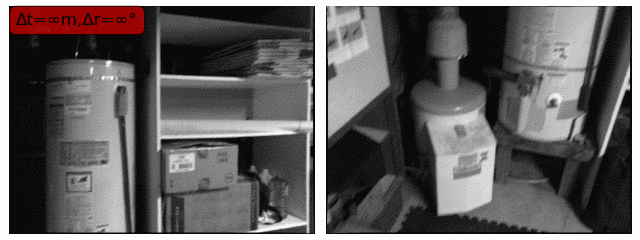}
\includegraphics[width=0.24\linewidth]{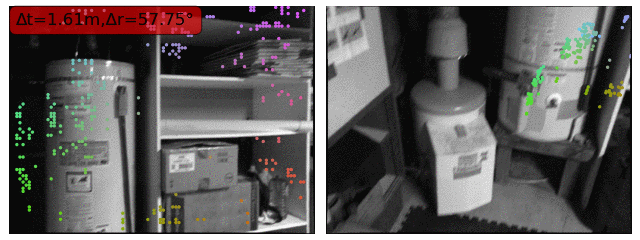}
\includegraphics[width=0.24\linewidth]{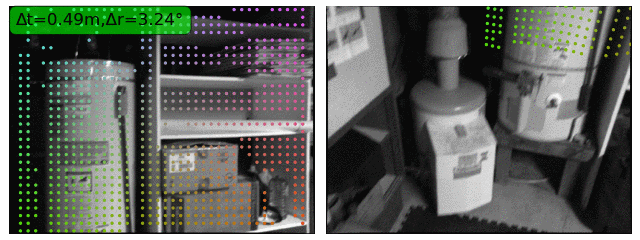}
\includegraphics[width=0.24\linewidth]{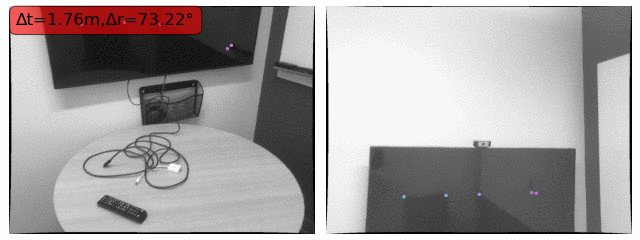}
\includegraphics[width=0.24\linewidth]{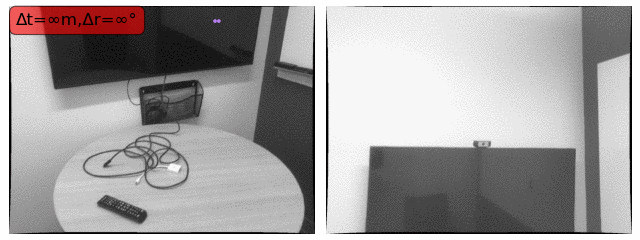}
\includegraphics[width=0.24\linewidth]{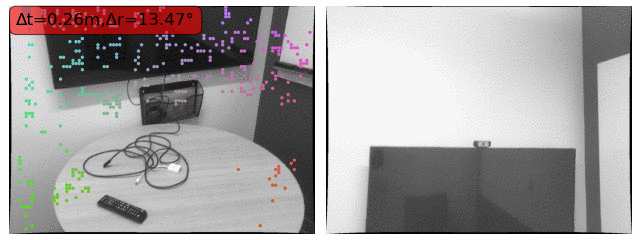}
\includegraphics[width=0.24\linewidth]{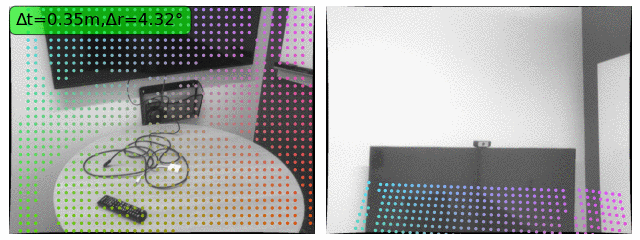}
\includegraphics[width=0.24\linewidth]{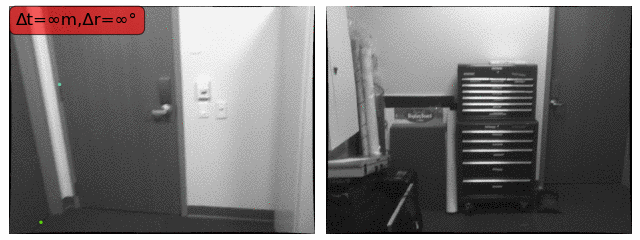}
\includegraphics[width=0.24\linewidth]{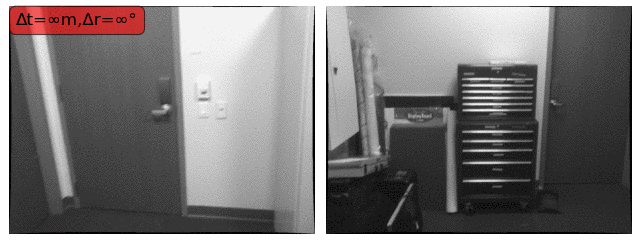}
\includegraphics[width=0.24\linewidth]{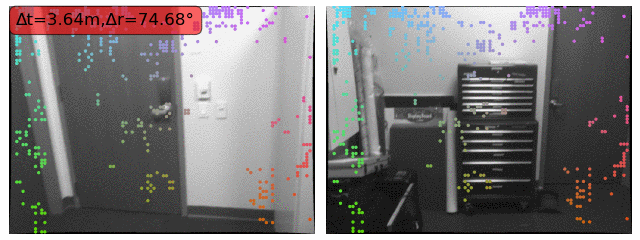}
\includegraphics[width=0.24\linewidth]{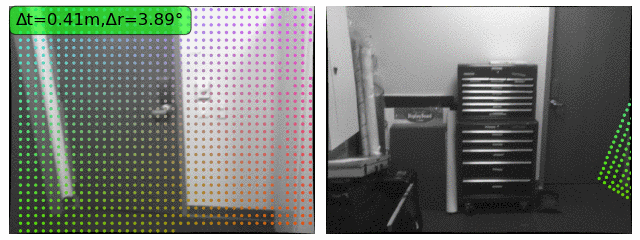}
\includegraphics[width=0.24\linewidth]{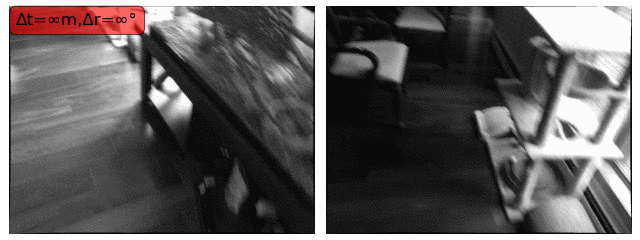}
\includegraphics[width=0.24\linewidth]{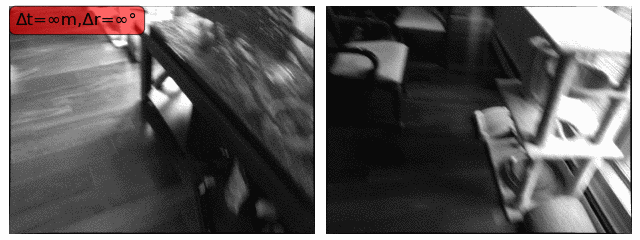}
\includegraphics[width=0.24\linewidth]{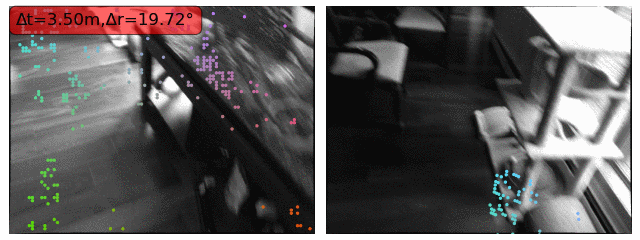}
\includegraphics[width=0.24\linewidth]{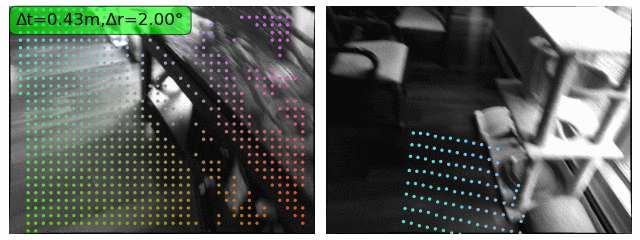}
\includegraphics[width=0.24\linewidth]{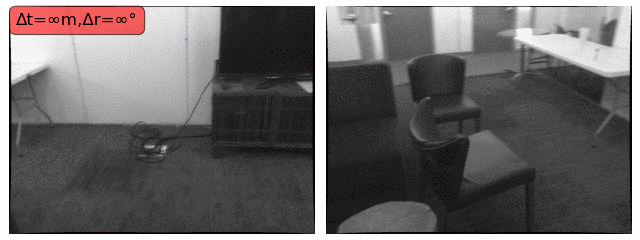}
\includegraphics[width=0.24\linewidth]{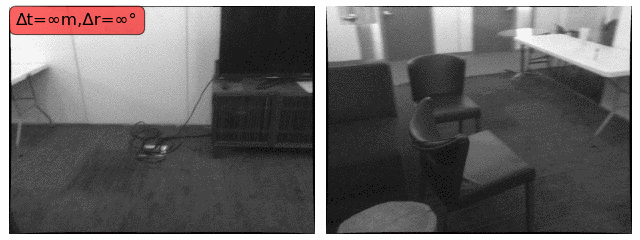}
\includegraphics[width=0.24\linewidth]{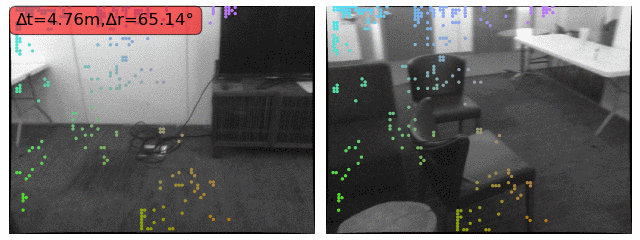}
\includegraphics[width=0.24\linewidth]{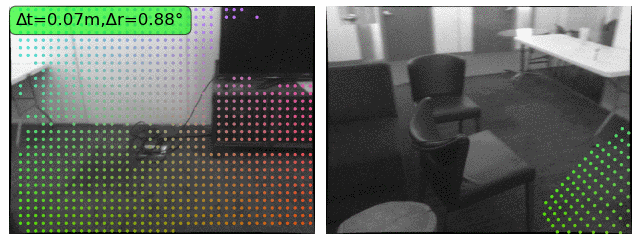}
\includegraphics[width=0.24\linewidth]{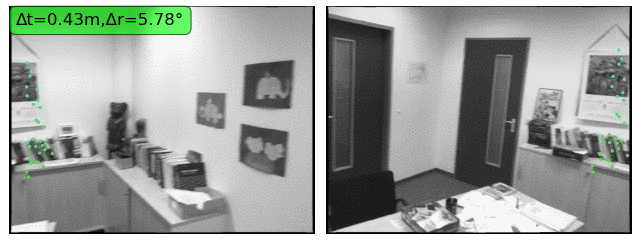}
\includegraphics[width=0.24\linewidth]{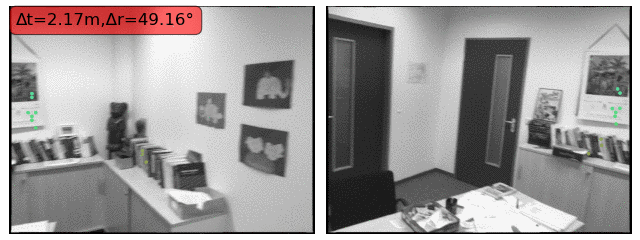}
\includegraphics[width=0.24\linewidth]{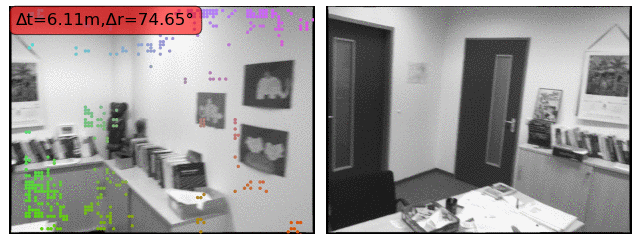}
\includegraphics[width=0.24\linewidth]{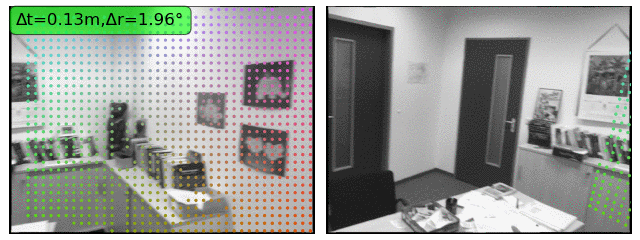}
\includegraphics[width=0.24\linewidth]{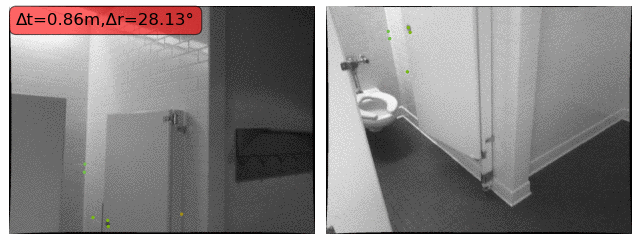}
\includegraphics[width=0.24\linewidth]{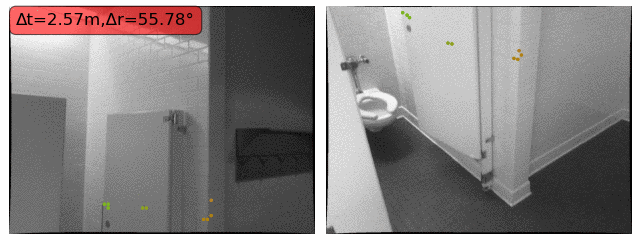}
\includegraphics[width=0.24\linewidth]{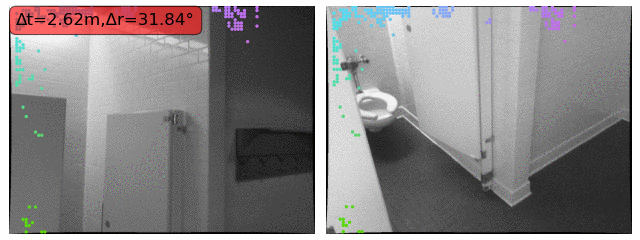}
\includegraphics[width=0.24\linewidth]{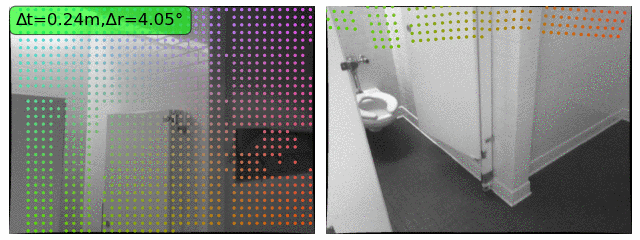}

\caption{\textbf{Qualitative camera pose estimation results on low-overlap
  images from ScanNet~\cite{ScanNet}:} We show for every method keypoints used
  as input for the camera pose estimator in the source image (left image), along
  with their predicted reprojection in the target image (right image). We
  color-code keypoints based 2D spatial position in the source image. We also
  report for every pair and every method the camera pose estimation error in
  translation and rotation, colored in \textcolor{green}{green} when the pose is
less than $\tau_t=0.5m$ and $\tau_r=10.0\degree$, and in
\textcolor{red}{red} otherwise.}

\label{fig:qualitative_localization} \end{figure}

\begin{figure}
\centering
~~~~~~~~~~~~~~~~
\small{SP+SG}
\hfill
\small{LoFTR}
\hfill
\small{DRCNet}
\hfill
\small{\neurhal}
~~~~~~~~~~~~~~~~
\\
\includegraphics[width=0.24\linewidth]{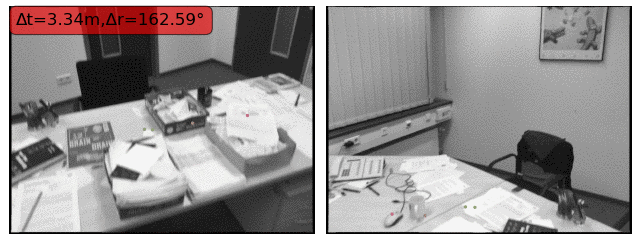}
\includegraphics[width=0.24\linewidth]{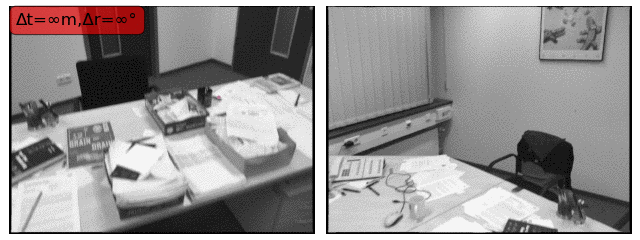}
\includegraphics[width=0.24\linewidth]{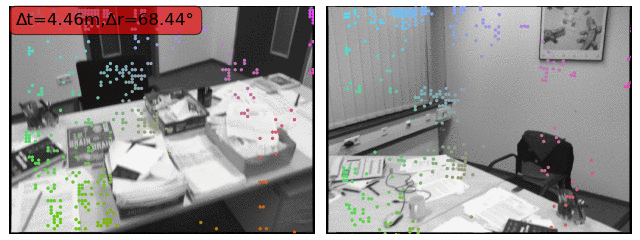}
\includegraphics[width=0.24\linewidth]{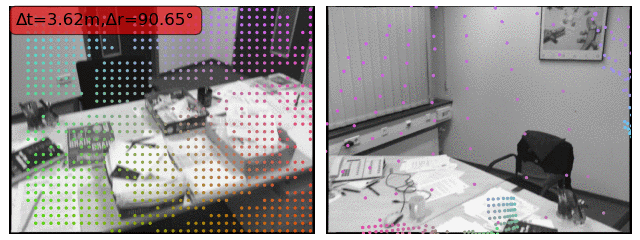}
\includegraphics[width=0.24\linewidth]{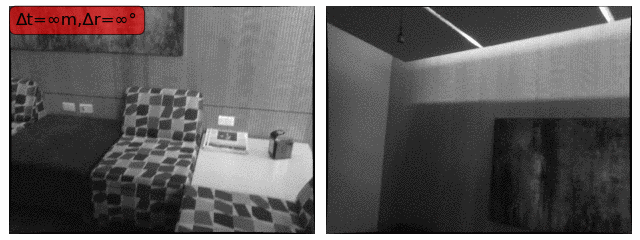}
\includegraphics[width=0.24\linewidth]{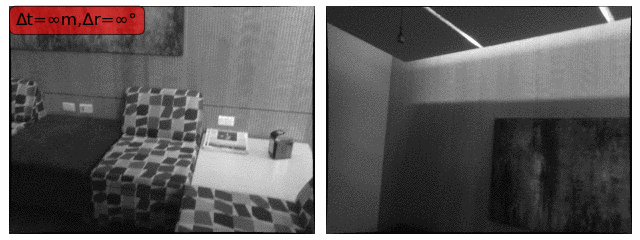}
\includegraphics[width=0.24\linewidth]{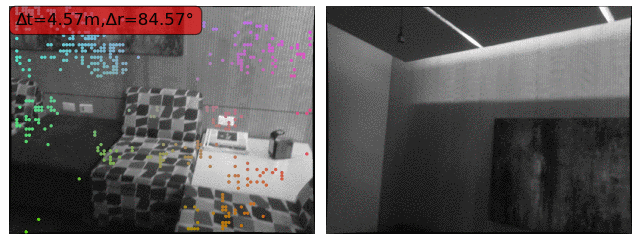}
\includegraphics[width=0.24\linewidth]{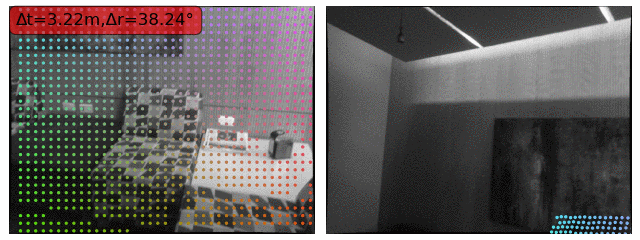}
\includegraphics[width=0.24\linewidth]{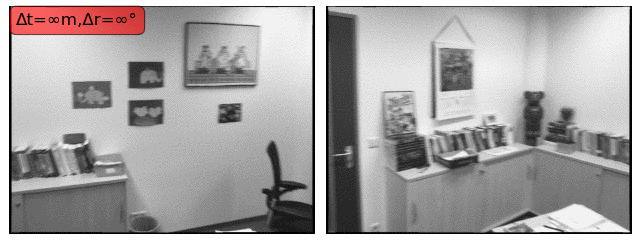}
\includegraphics[width=0.24\linewidth]{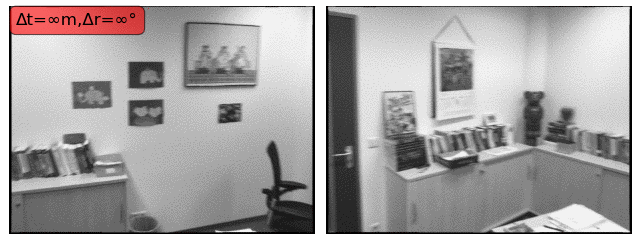}
\includegraphics[width=0.24\linewidth]{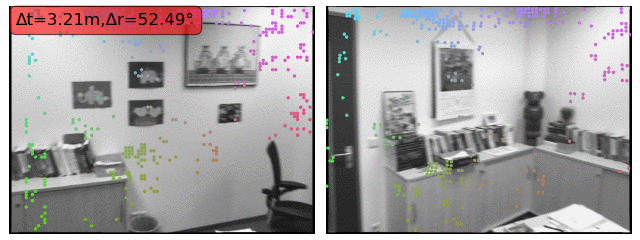}
\includegraphics[width=0.24\linewidth]{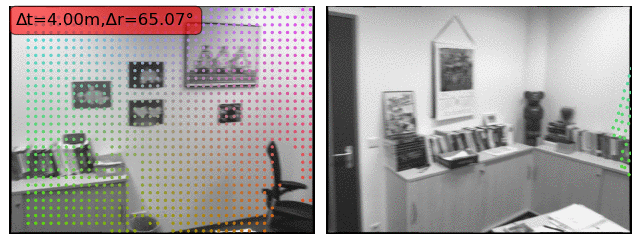}
\includegraphics[width=0.24\linewidth]{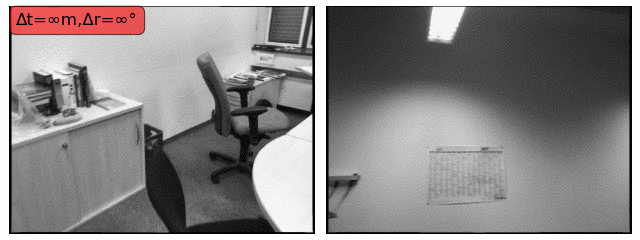}
\includegraphics[width=0.24\linewidth]{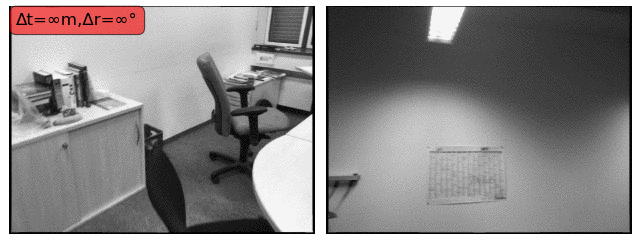}
\includegraphics[width=0.24\linewidth]{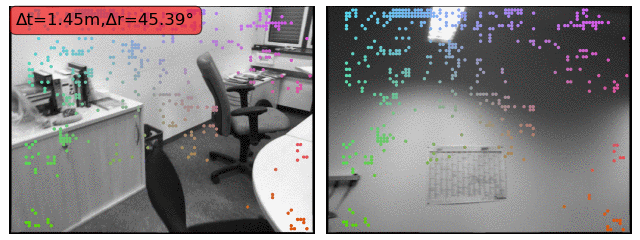}
\includegraphics[width=0.24\linewidth]{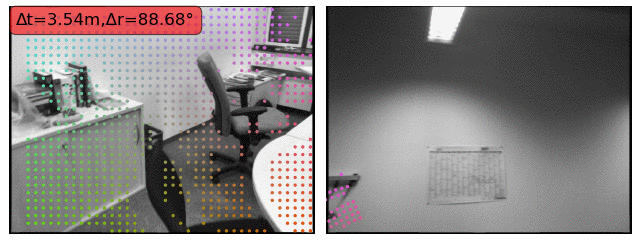}
\includegraphics[width=0.24\linewidth]{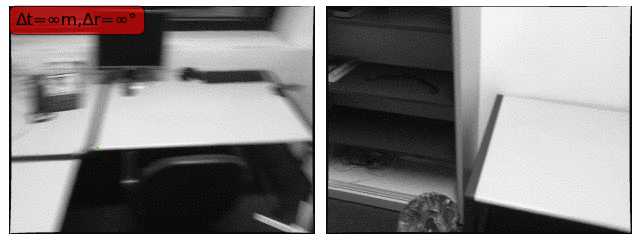}
\includegraphics[width=0.24\linewidth]{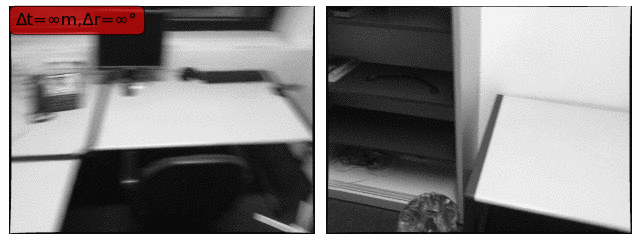}
\includegraphics[width=0.24\linewidth]{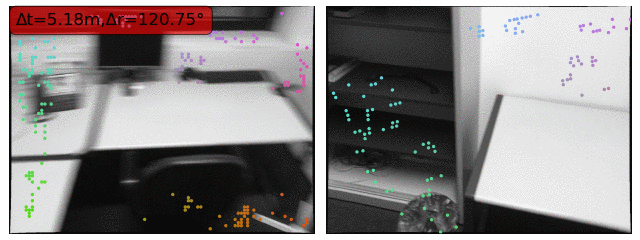}
\includegraphics[width=0.24\linewidth]{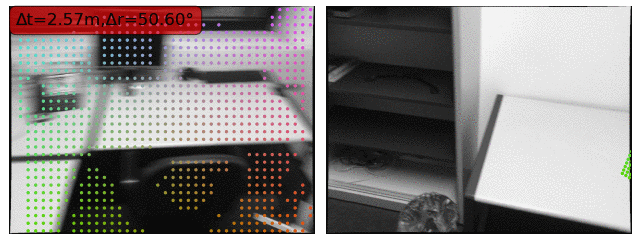}
\includegraphics[width=0.24\linewidth]{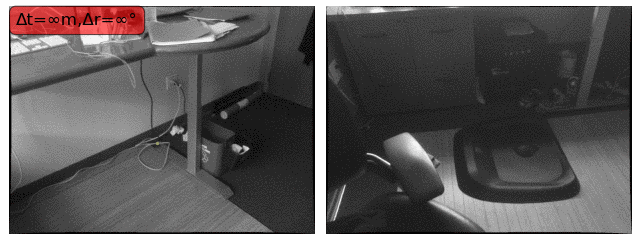}
\includegraphics[width=0.24\linewidth]{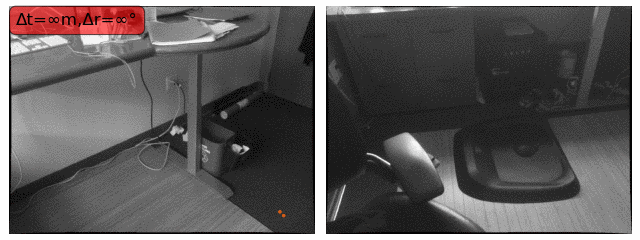}
\includegraphics[width=0.24\linewidth]{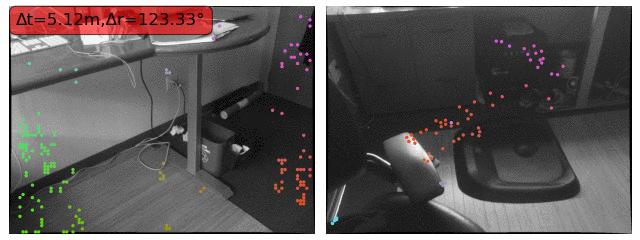}
\includegraphics[width=0.24\linewidth]{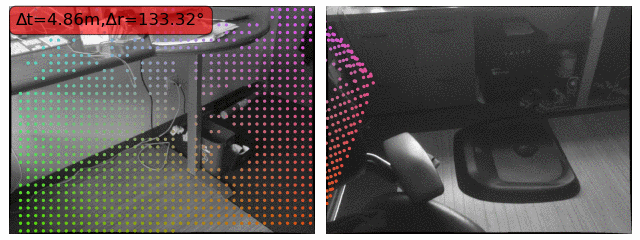}
\includegraphics[width=0.24\linewidth]{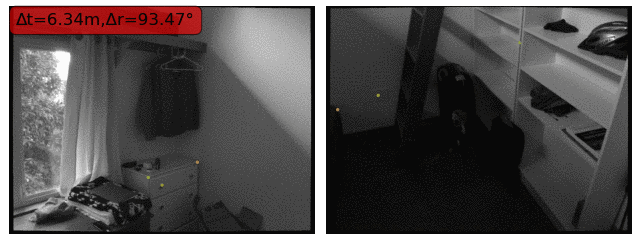}
\includegraphics[width=0.24\linewidth]{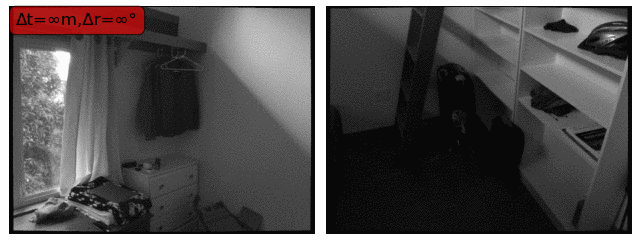}
\includegraphics[width=0.24\linewidth]{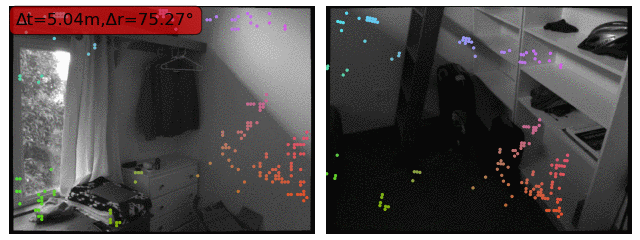}
\includegraphics[width=0.24\linewidth]{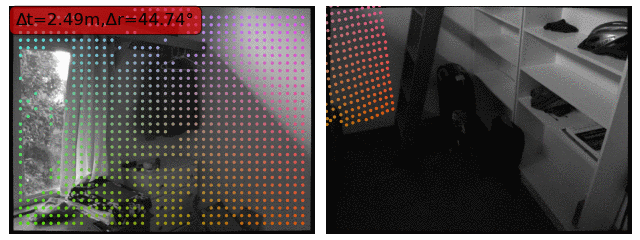}
\includegraphics[width=0.24\linewidth]{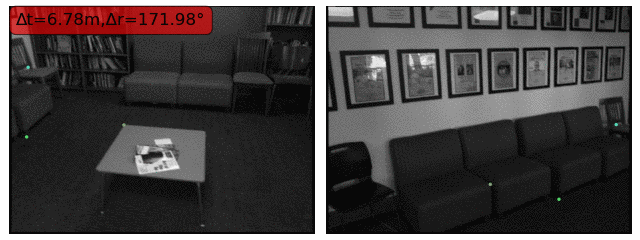}
\includegraphics[width=0.24\linewidth]{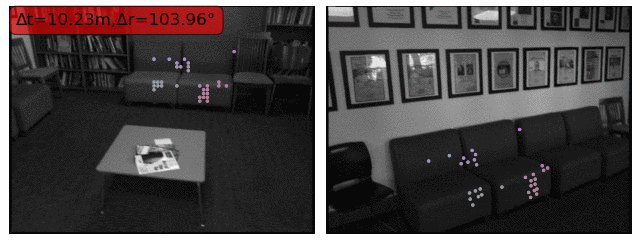}
\includegraphics[width=0.24\linewidth]{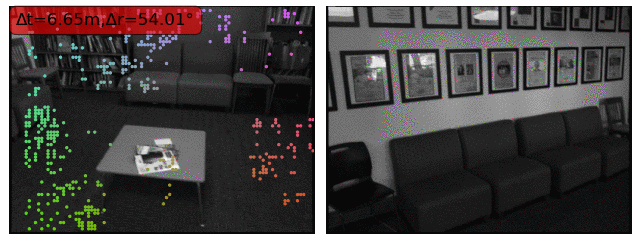}
\includegraphics[width=0.24\linewidth]{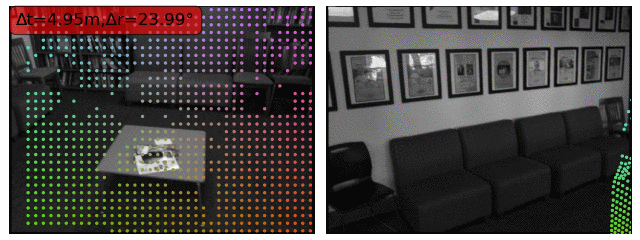}
\includegraphics[width=0.24\linewidth]{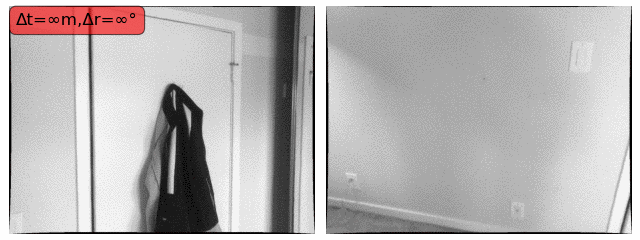}
\includegraphics[width=0.24\linewidth]{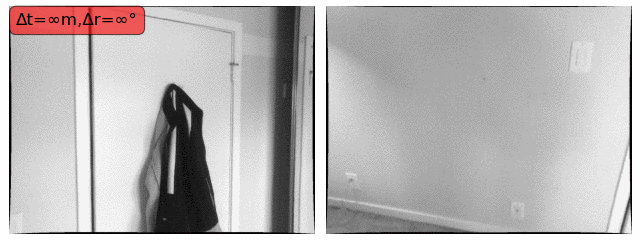}
\includegraphics[width=0.24\linewidth]{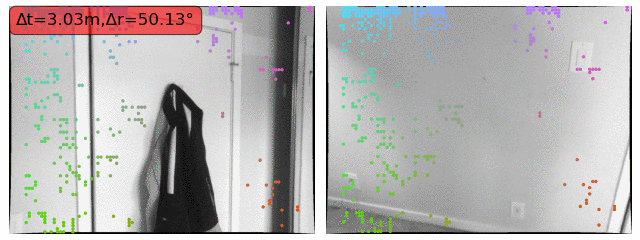}
\includegraphics[width=0.24\linewidth]{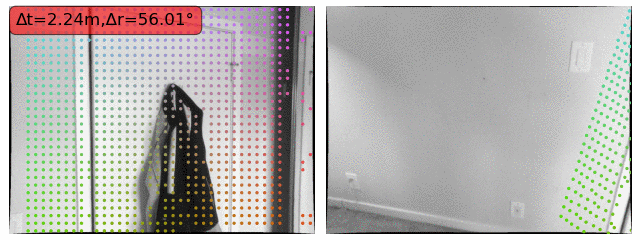}
\includegraphics[width=0.24\linewidth]{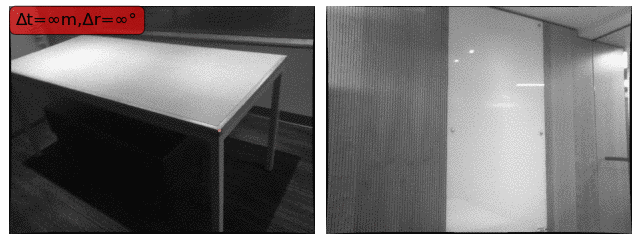}
\includegraphics[width=0.24\linewidth]{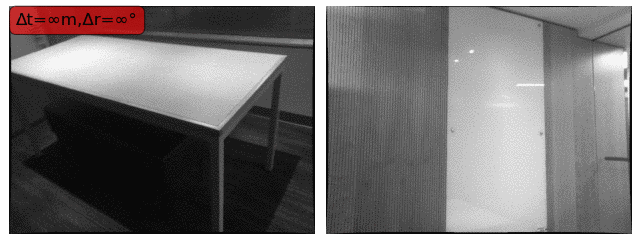}
\includegraphics[width=0.24\linewidth]{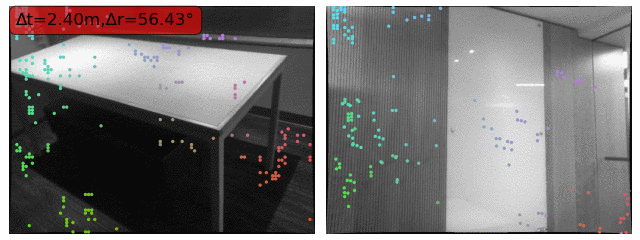}
\includegraphics[width=0.24\linewidth]{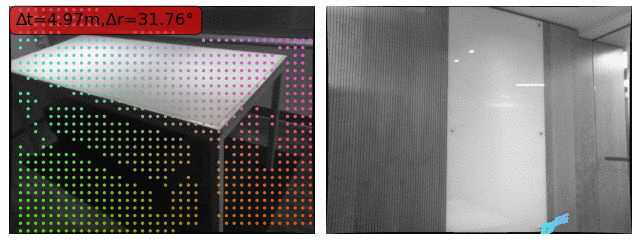}
\includegraphics[width=0.24\linewidth]{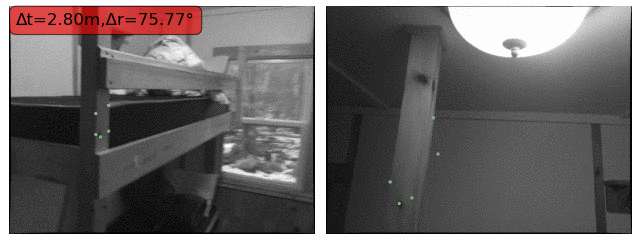}
\includegraphics[width=0.24\linewidth]{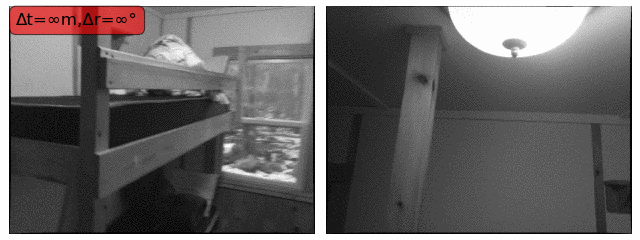}
\includegraphics[width=0.24\linewidth]{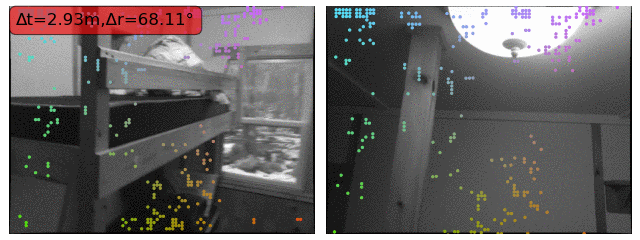}
\includegraphics[width=0.24\linewidth]{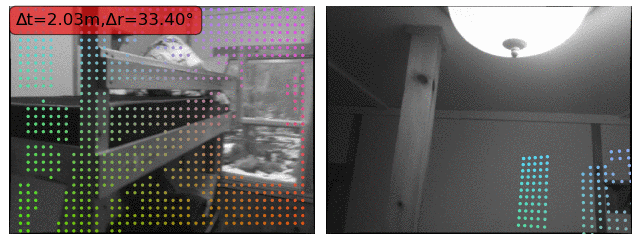}
\includegraphics[width=0.24\linewidth]{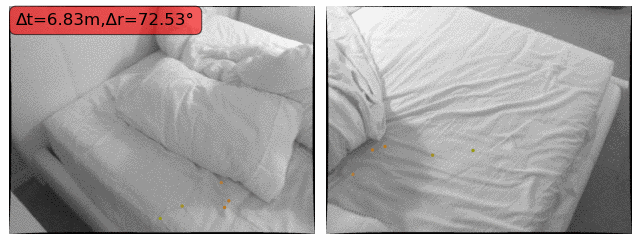}
\includegraphics[width=0.24\linewidth]{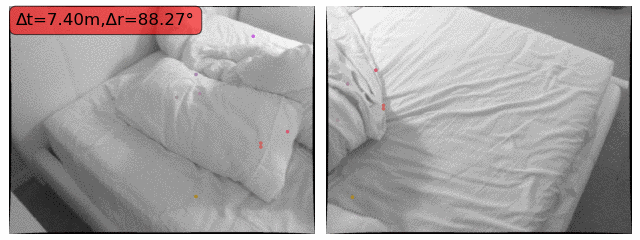}
\includegraphics[width=0.24\linewidth]{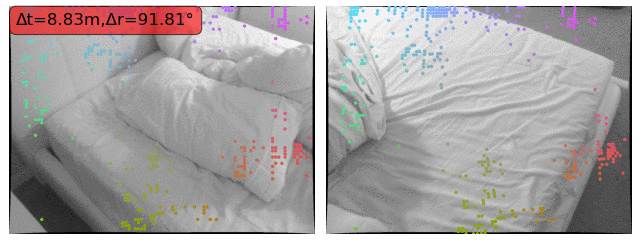}
\includegraphics[width=0.24\linewidth]{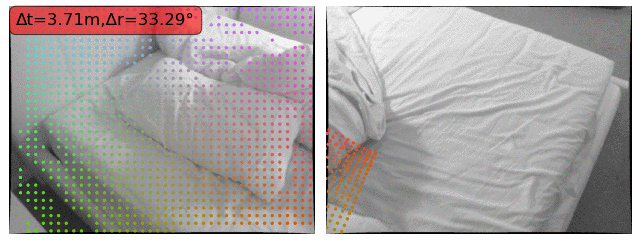}
\includegraphics[width=0.24\linewidth]{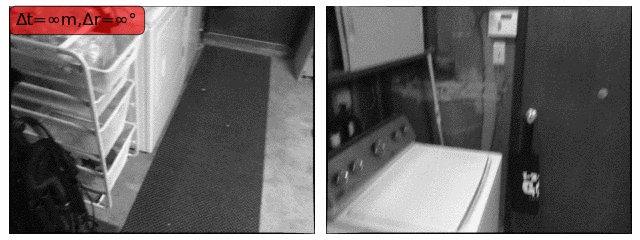}
\includegraphics[width=0.24\linewidth]{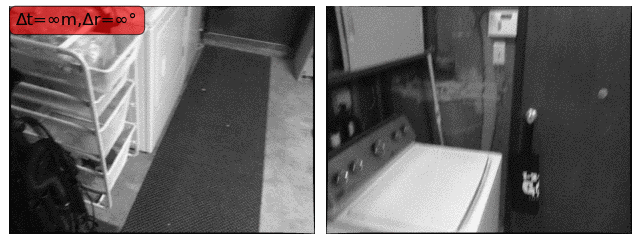}
\includegraphics[width=0.24\linewidth]{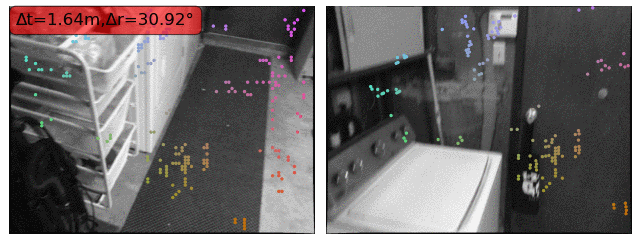}
\includegraphics[width=0.24\linewidth]{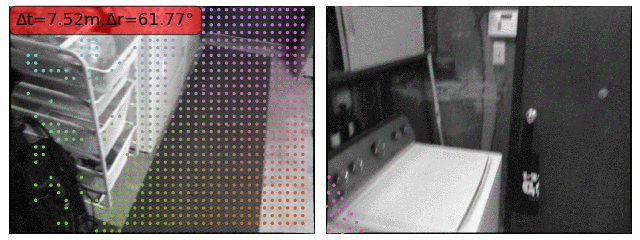}
\includegraphics[width=0.24\linewidth]{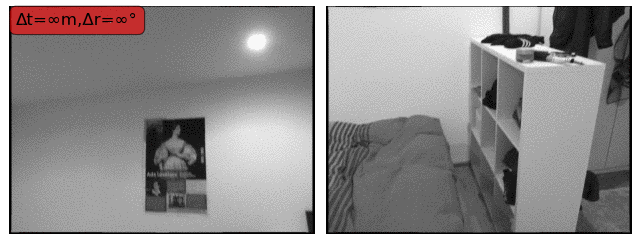}
\includegraphics[width=0.24\linewidth]{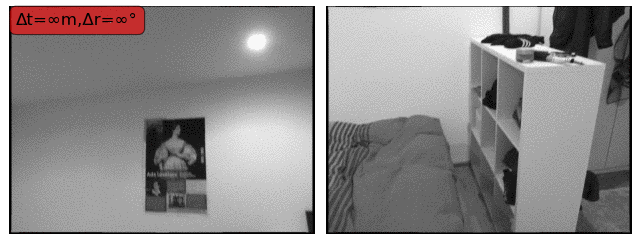}
\includegraphics[width=0.24\linewidth]{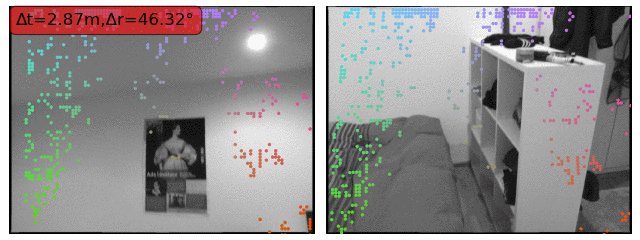}
\includegraphics[width=0.24\linewidth]{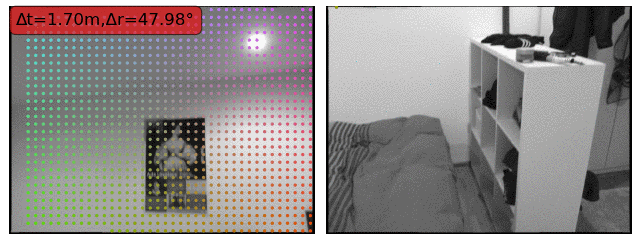}
\includegraphics[width=0.24\linewidth]{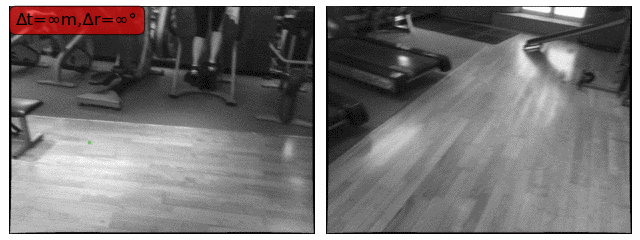}
\includegraphics[width=0.24\linewidth]{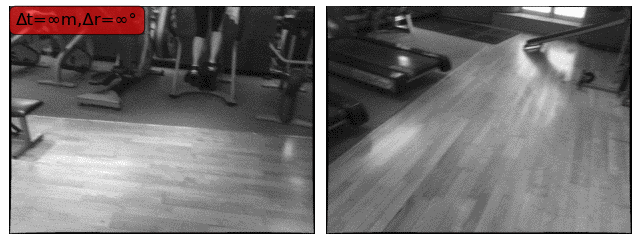}
\includegraphics[width=0.24\linewidth]{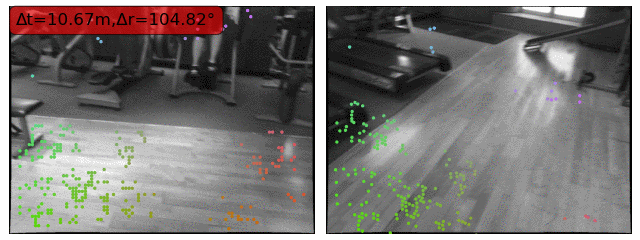}
\includegraphics[width=0.24\linewidth]{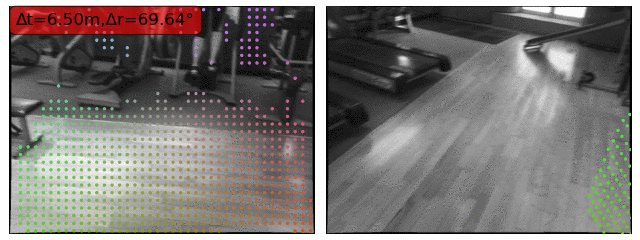}
\includegraphics[width=0.24\linewidth]{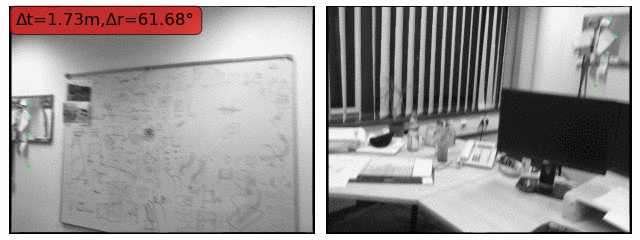}
\includegraphics[width=0.24\linewidth]{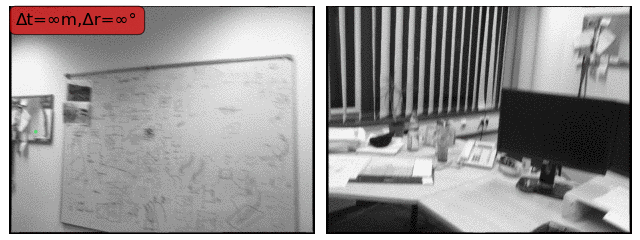}
\includegraphics[width=0.24\linewidth]{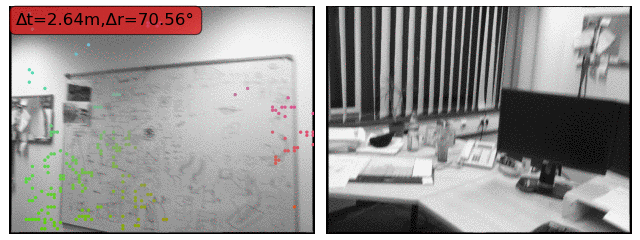}
\includegraphics[width=0.24\linewidth]{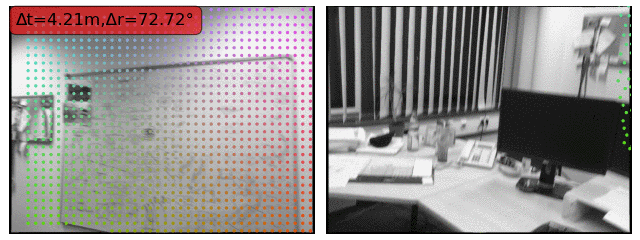}
\caption{\textbf{NeurHal failure cases on low-overlap images from ScanNet~\cite{ScanNet}:}
  We report cases where NeurHal fails to estimate a camera pose with an error
  less than $\tau_t=0.5m$ and $\tau_r=10.0\degree$. We find these cases often
  correlate with extremely low covisibility coupled with strong camera
  rotations.}
\label{fig:failure_qualitative_localization} \end{figure}

We show in Fig.~\ref{fig:qualitative_localization} qualitative results in camera
pose estimation on low-overlap images from ScanNet~\cite{ScanNet}, for \neurhal
and its three best-performing competitors.  For every method we display the
keypoints used as input to the camera pose estimator in the source image, along
with their reprojection at the estimated camera pose in the target image. For
methods using the pose estimator from~\cite{Chum2003LocallyOR}, the keypoints
are those that have been successfully matched. When using the pose estimator
of~\citet{NeuralReprojection}, the keypoints are those involved in the
prediction of the dense NRE maps.  We color in keypoints based on their spatial
2D position in the source image. We find that \neurhal strongly benefits from
its outpainting ability, in comparison with all other competitors which struggle
to find both sufficient and reliable correspondences.  We also report in
Fig.~\ref{fig:failure_qualitative_localization} failure cases for \neurhal. We
find that such cases correspond to image pairs exhibiting extremely limited
visual overlap, strong camera pose rotations and overall significant
ambiguities.

\end{document}